%% file: main.tex
\pdfoutput=1
\documentclass[]{mosi}
\usepackage{amsmath,amssymb,graphicx,booktabs,array,url}
\usepackage{hyperref}
\usepackage{float}
\usepackage{longtable}
\usepackage{enumitem}
\usepackage{makecell}
\title{RxnPFN: Towards Universal Training-free Prediction of Reaction Properties from Enantioselectivity to Yield}

\author{
Jiaqing Xie$^{1,2,4}$, Yuxin Wang$^{1,\dagger}$, Xiaopeng He$^{3}$, Ben Gao$^{5}$, Zihao Ye$^{2}$, Yuqiang Li$^{1,4}$, Tong Zhu$^{1,6}$, Zheng Ling$^{3}$, Xipeng Qiu$^{1,2,\dagger}$
\\[2mm]
{\normalfont \normalsize $^{1}$Shanghai Innovation Institute},
{\normalfont \normalsize $^{2}$Fudan University},
{\normalfont \normalsize $^{3}$Shanghai Jiao Tong University}\\
{\normalfont \normalsize $^{4}$Shanghai Artificial Intelligence Laboratory},
{\normalfont \normalsize $^{5}$Wuhan University},
{\normalfont \normalsize $^{6}$East China Normal University}\\
{\normalfont \normalsize $^{\dagger}$Corresponding author}

}

\newcommand{\figinclude}[3][\linewidth]{%
  \IfFileExists{#2}{\includegraphics[width=#1]{#2}}{%
    \IfFileExists{#3}{\includegraphics[width=#1]{#3}}{%
      \fbox{\parbox[c][0.25\textheight][c]{0.95\linewidth}{\centering\ttfamily Figure in preparation: \detokenize{#2}}}}}}

\newcommand{\rxnpfn}{RxnPFN}
\newcommand{\rxndirect}{RxnPFN w/o decomposition}

\abstract{\input{tex/1_abstract}}
\begin{document}
\maketitle
\section{Introduction}
\label{sec:intro}
\input{tex/2_introduction}

\section{Related work}
\label{sec:related}
\input{tex/3_related_work}

\section{Results}
\label{sec:results}
\input{tex/3_results}

\FloatBarrier
\section{Discussion}
\label{sec:discussion}
\input{tex/4_discussion}

\section{Methods}
\label{sec:methods}
\input{tex/5_methods}

\section{Data availability}
The benchmark reaction records are compiled from published experiments and public research datasets; sources, retained counts and processing steps are listed in Supplementary Note~\ref{app:data}. The Sadphos Heck collection compiles 703 published reactions from 13 papers on Pd/Sadphos catalysis (Sadphos ligand class; cited here via Singh et al.\ \citep{Li_Zhang_2024,Singh_2025}); the Co-AHO and out-of-sample hydrogenation sets are those released by Singh and Hern{\'a}ndez-Lobato \citep{Singh_2025}, and the Buchwald--Hartwig set is that of Ahneman et al.\ \citep{Ahneman_2018}. Processed benchmark splits, aggregate evidence files and all prediction summaries used for the figures will be released upon publication.

\section{Code availability}
Code for featurization, in-context inference, evaluation and figure generation will be released upon publication. Pretrained Kumo-Tabular weights are publicly available from NVIDIA under the model's released terms \citep{Kumo2026}.

\clearpage
\bibliographystyle{unsrtnat}
\bibliography{main}

\clearpage
\appendix
\setcounter{secnumdepth}{1}
\setcounter{section}{0}
\setcounter{figure}{0}
\setcounter{table}{0}
\renewcommand{\thesection}{\arabic{section}}
\renewcommand{\figurename}{Supplementary Figure}
\renewcommand{\tablename}{Supplementary Table}
\renewcommand{\theHfigure}{S\arabic{figure}}
\renewcommand{\theHtable}{S\arabic{table}}
\renewcommand{\theHsection}{S\arabic{section}}
\titleformat{\section}{\sectionfont\bfseries\mosiblue}{Supplementary Note~\thesection.}{0.5em}{}
{\noindent\sectionfont\bfseries Supplementary Information\par}
{\noindent\small Supplementary Information. The notes below cover: data sources; implementation details; quantitative details of the main results; further held-out reaction examples; complete baselines; information controls; additional benchmark analyses; representation, context and embedding exploration; uncertainty; yield and conversion; transfer to public collections; ScopeMap and SynAD; sparse-data study details; the Nature sparse-data benchmark; cobalt and nickel cases; and Co-AHO, the out-of-sample hydrogenation set and Buchwald--Hartwig.\par}
\vspace{2pt}
\input{tex/7_appendix}

\end{document}

%% file: tex/1_abstract.tex
Choosing a chiral catalyst for a new substrate is usually an exercise in analogy. Each measurement costs a synthesis and chiral analysis, so a screen rarely yields more than a few dozen enantioselectivities before a decision is needed, while the literature records thousands of related reactions that cannot be weighed quantitatively. We introduce RxnPFN to make that comparison explicit. Each reaction is encoded by component chemical roles and bonds broken and formed, and a frozen pretrained tabular network reads these descriptions alongside measured context outcomes, returning predictions in seconds with fixed weights. On 15,226 published asymmetric reactions, RxnPFN reaches pooled test $R^2=0.856$, against $0.814$ for the best tree ensemble on the same inputs and $0.816$ for the same network on a standard descriptor, assigning the favoured enantiomer correctly in 96.9\% of hydrogenations. In olefin hydrogenation, the recorded substrate-to-product stereochemical change encodes the observed enantiomer, whereas without product stereochemistry the chiral catalyst and substrate still predict the favoured enantiomer in 90\% of reactions, while ligand, metal and transformation govern selectivity magnitude. New campaigns enter by adding measurements rather than retraining: four reactions from a withheld study cut its prediction error by two thirds. The same frozen network predicts ketone-reduction conversion, oxidation yield, selectivity in cobalt hydrogenation and Sadphos Heck reactions, and Buchwald--Hartwig yield. Although accuracy tracks precedent proximity and declines when whole ligand families, scaffolds or publications are withheld, RxnPFN turns the growing literature record into a fast, training-free framework for quantitative reaction prediction and rapid catalyst adaptation.

%% file: tex/2_introduction.tex
Asymmetric catalysis is optimized one reaction system at a time. To find a catalyst for a new substrate class, a chemist typically varies the chiral ligand, the metal precursor, the solvent and the additives, and each data point costs a synthesis and, for selectivity, a chiral analysis \citep{Brown_2009,Zahrt_ChemRev_2019}. A campaign therefore rarely produces more than a few dozen measured enantioselectivities or yields before a decision has to be made about what to try next. The chemical literature, in contrast, contains thousands of related reactions, from substrate scopes and ligand screens of many research groups: published compilations alone record several thousand asymmetric hydrogenations \citep{Xu2021AHO,Cheng_2025}, ketone reductions, dihydroxylations and dearomatizations \citep{Ocampo2025,Gao2024Iodine}, and the distinct selectivity, yield and conversion outcomes of each \citep{OrgAIcat2026}. In practice chemists already use this record by analogy: a reaction of a similar substrate with a related ligand is the starting point for the next experiment. What they lack is a quantitative way to make that analogy, to weigh several precedents against each other and to know how far a prediction can be trusted.

Quantitative structure--selectivity relationships can make such comparisons explicit. Steric and electronic parameters of catalysts and substrates have revealed which substituents govern selectivity across a reaction series \citep{Zahrt_ChemRev_2019,Zahrt_2019,Reid_2019,Singh_2020,Betinol_Local_2025,Gao2024Iodine}. When the catalytic cycle is sufficiently well understood, descriptors of intermediates and transition states can extend these relationships to sparsely studied reactions \citep{Gallarati_2026,Hoque_2026}, including through three-dimensional models of the enantiodetermining transition state \citep{Hadler2026}. A model of one catalytic system can also provide a starting point for a related system \citep{Xu2025NiTransfer,Zhang2023ChiralAcid}. Extending either approach to a different mechanism requires a new choice of chemically relevant structures or descriptors.

Learning directly from reaction records offers another way to draw on published experiments. Molecular graphs, reaction fingerprints and language models have been used to predict selectivity and yield \citep{Yang_2019,Heid_2023,Schwaller_Yield_2021,Probst_2022,Ahneman_2018,Gao2026LoRAChem}. ChemAHNet incorporates the substrate--catalyst interactions characteristic of olefin hydrogenation \citep{Cheng_2025}, meta-learning has been used to identify selective hydrogenations from a few measurements on a new system \citep{Singh_2025}, and language models fine-tuned on reaction data have been paired with generative design of new ligands and checked experimentally \citep{Hoque2025CSP3H}. In reaction development, the set of relevant precedents changes as soon as the first results for a new ligand or substrate become available. We asked whether both the published precedents and those first measurements could be used directly, without retraining a model for each campaign.

A prior-fitted network predicts an outcome from a table of examples supplied to it at the time of prediction \citep{Hollmann2025,Kumo2026,TabICL2026,TabDPT2026,Xiaomi2026,Monroe2026}. Such a network is trained once, on synthetic tasks drawn from a prior over tabular regression, and is then applied without any further fitting: the table is the model's working memory rather than its training set. This sidesteps the cost that matters in reaction development, because a campaign no longer has to supply enough labelled examples for a fit. It does not remove the chemistry, however. For asymmetric catalysis, two features of the experimental record matter. The first is the role of each molecule. Changing a ligand and changing a substrate have different chemical consequences, even if the same substituent is exchanged. Reaction fingerprints capture structural changes \citep{Rogers_2010,Schneider_2016,Probst_2022}, and role-aware descriptions have been used for yield prediction and data curation \citep{Mirji2026,OrgAIcat2026}. The second is what the measurement establishes. Some studies report which enantiomer is favoured, whereas others record only the magnitude of the enantiomeric excess (ee). An off-the-shelf prior-fitted network reads such a table column by column, with no notion of which molecule plays which part, and returns one regression whether the label is a signed free energy or a magnitude. Supplying the roles explicitly, and matching the readout to what the label establishes, is what this work adds to the backbone itself.

We introduce RxnPFN, which supplies a frozen pretrained tabular network with the components, the bonds changed and the measured outcomes of related reactions (Fig.~\ref{fig:method}) \citep{Kumo2026}. No weight of the network is updated, and the only quantities fitted to a campaign are label-free projections of the input. On held-out reactions from three families, olefin hydrogenation, ketone reduction and asymmetric oxidation, RxnPFN reaches $R^2=0.856$, against $0.814$ for the strongest fitted tree ensemble with the same chemical information and $0.816$ for the same frozen network on a standard descriptor. Predicting 1,501 outcomes from a prepared table of 13,181 reactions takes about twelve seconds. Fine-tuning a reaction language model takes 153~s, and training a reaction graph network for 80 epochs takes about 2,660~s (Supplementary Table~\ref{tab:cost}).

The gain comes from describing the record in two ways that the backbone cannot supply by itself. Reactions are described by the chemical role of each component and by the bonds broken and formed. The stored measurement is treated as what it establishes, a signed free energy where the favoured enantiomer was recorded and a magnitude where only the enantiomeric excess was. Separating these two clarifies what the model reads from a reaction record. When product stereochemistry is supplied in olefin hydrogenation, the substrate-to-product change carries the recorded configurational assignment; when product stereochemistry is withheld, the chiral catalyst and substrate still recover 90\% of the favoured enantiomers, while the ligand, metal centre and transformation govern the magnitude of selectivity. In ketone reduction, the catalyst complex is the strongest contributor to magnitude prediction.

To define both the reach and the boundaries of this approach, we map how accuracy varies with precedent proximity across grouped splits that hold out ligand families, substrate scaffolds or publications, extend the same frozen model to additional reaction classes, metals and yield or conversion endpoints, and show how a few measurements from a new campaign rapidly adapt predictions to its remaining reactions.

%% file: tex/3_related_work.tex
Descriptor-based regression has guided catalyst selection and enantioselectivity prediction across many reaction classes \citep{Zahrt_ChemRev_2019,Zahrt_2019,Reid_2019,Singh_2020}. When the catalytic cycle is sufficiently well understood, descriptors of intermediates and transition states extend these relationships to sparsely studied reactions \citep{Gallarati_2026,Hoque_2026}, and a model of one catalytic system can seed a related one, as in the transfer of selectivity models from palladium to nickel \citep{Xu2025NiTransfer} or the design of chiral carboxylic acids for cobalt catalysis \citep{Zhang2023ChiralAcid}. A different line learns a reaction representation from the records themselves: circular and reaction fingerprints \citep{Rogers_2010,Schwaller_ReactionSpace_2021,Probst_2022}, molecular and reaction graphs \citep{Heid_2023,Xu2025RXNGraphormer,AguilarBejarano2025}, and reaction language models \citep{Schwaller_Yield_2021,Sagawa_2025,Burns_2025}. ChemAHNet is the closest of these to the present task, a chemistry-informed graph network released with 9,478 hydrogenations \citep{Cheng_2025}, and prototypical meta-learning has been used to identify selective hydrogenations from a small support set \citep{Singh_2025}. All of these methods fit parameters to the reaction class at hand; while task-specific models remain strong on narrowly defined benchmarks, each new campaign requires retraining or dedicated descriptors.

A prior-fitted network removes that fitting step: TabPFN showed that pretraining on a prior over synthetic tasks yields a reusable in-context predictor for small tables \citep{Hollmann2025}, and TabICL and further releases extend the approach to larger contexts \citep{TabICL2026,Kumo2026,TabDPT2026,Xiaomi2026}. Monroe brings in-context probabilistic inference to molecular-property prediction \citep{Monroe2026}; it conditions on a pretrained molecular encoder, whereas here the table itself carries the chemistry and every weight stays fixed. Two properties of a reaction table are then ours to supply. The first is the role of each molecule: reaction role assignment has a literature of its own \citep{Schneider_2016}, and role-aware descriptions have been used for yield prediction and data curation \citep{Mirji2026,OrgAIcat2026}. The second is what the measurement establishes: some studies identify the favoured enantiomer, while others report only the magnitude of enantiomeric excess. Our readout handles both signed free-energy differences and magnitude-only labels. Similarity-based retrieval has been used for general tabular data \citep{Thomas2024LoCalPFN}. We compare retrieval with local regression on the same inputs. For uncertainty, conformalized quantile regression calibrates quantile intervals on held-out data \citep{Romano2019,Angelopoulos2021}, and TabCUQ separates context-induced uncertainty in tabular in-context models \citep{TabCUQ2026}; we use the former to recalibrate Kumo's native quantiles and report coverage per collection. Reaction-scope exploration and applicability domains are addressed by ScopeMap and SynAD \citep{Li2026ScopeMap,Tan2026SynAD}, whose released data we re-evaluate in Supplementary Note~\ref{app:external}. RxnPFN's own contribution is the combination that a frozen network can exploit: chemical roles supplied explicitly, the transformation described as a change of environments, a readout matched to what the label records, and no network weight updated.

%% file: tex/3_results.tex
\begin{figure*}[p]
\centering
\figinclude[0.95\linewidth]{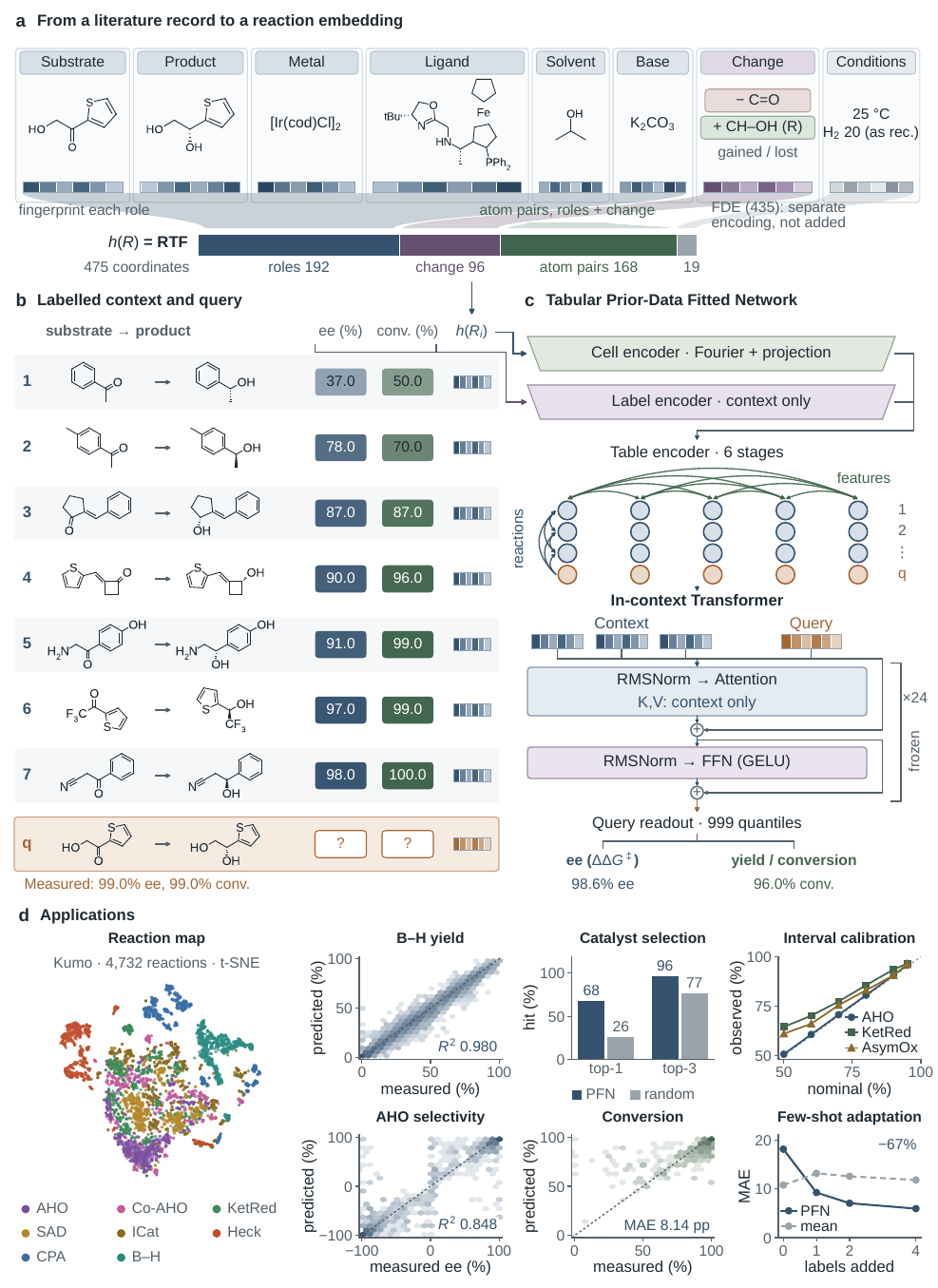}{figures/fig1_method.png}
\caption{\textbf{RxnPFN: encoding literature reactions for in-context prediction and applications.}}
\label{fig:method}
\end{figure*}
\begin{figure*}[!t]
\ContinuedFloat
\captionsetup{font=footnotesize,singlelinecheck=false}
\caption[]{ \textbf{a}, Chemical roles (with the additive role shown here as Base, K$_2$CO$_3$), substrate-to-product changes, atom-pair counts and conditions form a role--transformation fingerprint (RTF). The fingerprint--descriptor encoding (FDE) is a separate 435-dimensional encoding built from the same roles, which adds a PCA of the reaction-level R-SPOC descriptor to compact role blocks. It is used in place of the RTF for the magnitude head, not added to it. Feature bars are schematic.
\textbf{b}, Seven recorded training examples selected to illustrate the observed range of enantiomeric excess (ee) and conversion, and one query. The prediction uses the full 3,430-reaction KetRed context. The measured outcome of the query is shown for comparison and was not supplied to the model.
\textbf{c}, Frozen Kumo-Tabular combines a table encoder with in-context Transformer layers. Attention uses context keys and values; residual connections are shown. The readout returns 999 quantiles, with separate passes for each target.
\textbf{d}, Left, t-SNE of cross-fitted Kumo reaction embeddings (up to 600 per class), coloured by class. AHO, asymmetric olefin hydrogenation; Co-AHO, cobalt-catalysed hydrogenation; KetRed, ketone reduction; SAD, Sharpless dihydroxylation; ICat, iodine-catalysed dearomatization; CPA, chiral phosphoric acid; Heck, Pd/Sadphos Heck reactions; B--H, Buchwald--Hartwig amination. Right, top row: Buchwald--Hartwig (B--H) yield (11,880 held-out predictions; $R^2$ averaged over ten splits), catalyst selection (82 matched groups; PFN versus random), and native interval calibration (AHO selectivity, KetRed conversion and AsymOx yield; separate proper-fit evaluation). Bottom row: AHO selectivity (940 test reactions; $R^2$ computed on the ee (\%) scale, which is not directly comparable with the $\Delta\Delta G^\ddagger$ scale used in Table~\ref{tab:main}), KetRed conversion (402 test reactions), and Sadphos Heck adaptation (up to four labels per paper, twelve papers, ten label orders). Hexagons show log counts; dashed diagonals denote equality. MAE, mean absolute error; pp, percentage points. Selection and evaluation details are in Methods.}
\end{figure*}
\subsection{In-context prediction from chemical roles and precedents}
\label{sec:overview}

A published substrate scope or ligand screen provides both the structures of related reactions and the outcomes that distinguish them. RxnPFN reads these records as a table, with one row per reaction (Fig.~\ref{fig:method}). Building a row takes three steps. First, each component keeps its chemical role: substrate, product, metal-containing catalyst, ligand, solvent and additive are separate blocks, so that the same substituent in a ligand and in a substrate enters the model as different information. Second, the change between substrate and product is described as the bond environments lost and gained, which makes the transformation itself an input rather than a label. Third, the two encodings are formed from these blocks. The role--transformation fingerprint (RTF) concatenates chirality-aware counts for the six roles, the gained and lost environments and the parsed conditions into 475 coordinates. The fingerprint--descriptor encoding (FDE) adds 192 coordinates of physicochemical descriptors aggregated over catalyst, conditions and transformation, and is used for the magnitude head. The frozen Kumo-Tabular network then reads the table and returns, for a query row, a distribution over the outcome with its weights fixed throughout. Two readouts act on that distribution. For enantioselectivity, the magnitude and the sense of induction are estimated separately wherever both are recorded, and a reference reaction can be supplied when the published series shares a common ligand or substrate (Methods and Supplementary Note~\ref{app:details}).

\subsection{Enantioselectivity across hydrogenations, reductions and oxidations}
\label{sec:main}

We assembled 15,226 enantioselectivity measurements from three bodies of published work (Fig.~\ref{fig:main}a and Methods). The asymmetric olefin hydrogenation collection (AHO) covers di-, tri- and tetrasubstituted olefins and includes the identity of the favoured enantiomer in its signed $\Delta\Delta G^\ddagger$ values \citep{Xu2021AHO,Cheng_2025}. The ketone reduction collection (KetRed) follows prochiral ketones to secondary alcohols. The asymmetric oxidation collection (AsymOx) comprises Sharpless dihydroxylations and hypervalent-iodine-catalysed dearomatizations of phenols and naphthols. The reduction and oxidation records give the magnitude of enantioselection.

\begin{table}[t]
\centering\footnotesize
\caption{\textbf{Test results on the 1,501-reaction benchmark under two training settings.} \emph{Pooled}: one model (or one in-context table) for all three datasets, using a 13,181-reaction final context for the frozen models (training protocols for fitted baselines are specified below); pooled $R^2$, MAE and RMSE (kcal\,mol$^{-1}$) over all test reactions. \emph{Per dataset}: one model per dataset, trained (or given as context) only that dataset's training reactions, with every label-free projection refitted on them; test $R^2$ per dataset and their macro average. \rxnpfn{} is the validation-selected pipeline (FDE magnitude head, RTF direction head); the variant with an RTF magnitude head is shown for reference. For the magnitude-only datasets KetRed and AsymOx, \rxnpfn{} reduces to its magnitude head, so its per-dataset values there agree with those of \rxndirect{} on the same input to within the seed spread. Best displayed values per column in bold; where two rows tie within the seed spread, the validation-selected model is bold. The \rxndirect{} row on the FDE input is a single-seed reference and carries no spread. Values for the three RxnPFN rows are means over model seeds that change the estimator ensemble and the feature order: five seeds for the pooled columns, three for the per-dataset columns, and one for the per-dataset columns of the RTF-magnitude row. Seed s.d. is at most $0.007$ throughout.}
\label{tab:main}
\setlength{\tabcolsep}{3.5pt}
\begin{tabular}{llrrrrrrr}
\toprule
& & \multicolumn{3}{c}{Pooled} & \multicolumn{4}{c}{Per dataset ($R^2$)}\\
\cmidrule(lr){3-5}\cmidrule(lr){6-9}
Model & Input & $R^2$ & MAE & RMSE & AHO & KetRed & AsymOx & Macro\\
\midrule
ExtraTrees & R-SPOC & 0.807 & 0.471 & 0.868 & 0.729 & 0.756 & 0.706 & 0.730\\
RandomForest & R-SPOC & 0.793 & 0.534 & 0.898 & 0.719 & 0.685 & 0.698 & 0.701\\
XGBoost & R-SPOC & 0.746 & 0.670 & 0.995 & 0.665 & 0.670 & 0.725 & 0.687\\
ExtraTrees & RTF & 0.814 & 0.508 & 0.852 & 0.751 & 0.701 & 0.698 & 0.717\\
\midrule
TabPFN 3.5 & R-SPOC PCA & 0.794 & 0.519 & 0.897 & 0.724 & 0.730 & 0.707 & 0.721\\
TabICLv2 & R-SPOC PCA & 0.776 & 0.534 & 0.934 & 0.706 & 0.712 & 0.688 & 0.702\\
Xiaomi-TabLDM & R-SPOC PCA & 0.767 & 0.562 & 0.953 & 0.688 & 0.693 & 0.699 & 0.693\\
TabDPT 1.3 & R-SPOC PCA & 0.765 & 0.563 & 0.957 & 0.685 & 0.697 & 0.676 & 0.686\\
Kumo & R-SPOC & 0.816 & 0.488 & 0.848 & 0.741 & 0.739 & 0.704 & 0.728\\
\midrule
DRFP + XGBoost$^\dagger$ & reaction fingerprint & 0.754 & 0.622 & 0.979 & 0.648 & 0.690 & 0.684 & 0.674\\
Chemprop (CGR)$^\dagger$ & reaction graph & 0.795 & 0.562 & 0.894 & 0.726 & 0.612 & \textbf{0.734} & 0.691\\
rxnfp-BERT, fine-tuned$^\dagger$ & reaction SMILES & 0.719 & 0.654 & 1.047 & 0.591 & 0.580 & 0.588 & 0.586\\
rxnfp + MLP$^\dagger$ & reaction SMILES & 0.585 & 0.867 & 1.272 & 0.453 & 0.512 & 0.571 & 0.512\\
ReactionT5v2-yield$^\ddagger$ & reaction SMILES (T5) & 0.678 & 0.745 & 1.120 & 0.557 & 0.508 & 0.568 & 0.544\\
CheMeleon + Chemprop v2$^\ddagger$ & molecule graphs & 0.767 & 0.594 & 0.953 & 0.682 & 0.655 & 0.698 & 0.678\\
RXNGraphormer$^\S$ & reaction graphs & 0.721 & 0.643 & 1.042 & 0.645 & 0.646 & -0.222 & 0.356\\
ChemAHNet, our retraining$^\S$ & reaction SMILES & 0.609 & 0.798 & 1.234 & 0.492 & 0.620 & 0.556 & 0.556\\
ChemAHNet, as reported$^\P$ & reaction SMILES & -- & -- & -- & 0.591 & -- & -- & --\\
HCat-GNet$^\S$ & molecular graphs & 0.556 & 0.938 & 1.316 & 0.283 & 0.589 & 0.604 & 0.492\\
\midrule
\rxndirect & FDE (435) & 0.831 & 0.448 & 0.812 & 0.786 & 0.764 & 0.716 & 0.756\\
\rxndirect & RTF (475) & 0.841 & 0.445 & 0.787 & 0.787 & 0.760 & 0.723 & 0.757\\
\rxnpfn{} (RTF magnitude) & RTF & 0.856 & 0.431 & \textbf{0.748} & 0.806 & 0.759 & 0.722 & 0.762\\
\rxnpfn{} (selected) & FDE + RTF & \textbf{0.856} & \textbf{0.430} & 0.750 & \textbf{0.805} & \textbf{0.767} & 0.712 & \textbf{0.761}\\
\bottomrule
\end{tabular}

\smallskip
\footnotesize $^\dagger$DRFP: 2,048-bit differential reaction fingerprint (radius 3) with XGBoost in the settings of ref.~\citealp{Probst_2022}; 48 reactions cannot be fingerprinted and are encoded as zeros. Chemprop: condensed-graph reaction input with validation-selected checkpoints (pooled, 12,165 fitting reactions; per dataset, that dataset's). rxnfp-BERT: fine-tuned end to end, learning rate chosen once on pooled validation. rxnfp + MLP: early stopping on 10\% of the training reactions. Chemprop, rxnfp-BERT and rxnfp + MLP: mean of three seeds. $^\ddagger$ReactionT5v2 and CheMeleon: mean of three validation-selected fits, recomputed from saved test predictions. $^\S$RXNGraphormer, ChemAHNet and HCat-GNet \citep{Xu2025RXNGraphormer,Cheng_2025,AguilarBejarano2025}: three seeds on the original held-out memberships, with checkpoints selected on validation reactions; RXNGraphormer and HCat-GNet omit the same two unparsable KetRed fitting structures and retain every held-out reaction, and all values are means of per-seed metrics rather than an across-seed ensemble. $^\P$Published value on the identical 940-reaction test set \citep{Cheng_2025}; the $^\S$ row is our retraining. RXNGraphormer's AsymOx arm ($-0.222$) is a failed transfer; over AHO and KetRed alone its macro average is $0.646$.
\end{table}

Under the per-dataset setting of Table~\ref{tab:main}, with the recorded product structures supplied, RxnPFN predicted selectivity across all three collections (Fig.~\ref{fig:main}b--e). Its macro-averaged test $R^2$ was $0.761$, against $0.730$ for the strongest tree ensemble and $0.728$ for the same frozen backbone on a standard descriptor. Given the three collections as a single table, the same recipe reaches a pooled $R^2=0.856$ with MAE $0.430$~kcal\,mol$^{-1}$ (Table~\ref{tab:main}). The margin was widest on olefin hydrogenation, $0.805$ against $0.751$ for the best tree model on the same input, the only collection with signed labels and therefore the only one in which the direction of induction has to be predicted, and smallest on the oxidations, where Chemprop was nominally higher at $0.734$ against $0.712$. For a signed label the additional failure mode is the sense of induction: a single reversal turns a small error in free energy into a large error in ee. Magnitude-only collections cannot incur that error and are easier on the same scale, and the advantage of separating magnitude from direction disappears there. Because a pooled score combines the two kinds of measurement and rewards correct assignment of the favoured enantiomer, we report the collections separately and treat magnitude as its own quantity (Supplementary Table~\ref{tab:components}).

Two features of the ketone-reduction labels limit what that collection can show: at the top of the ee scale a rounding step moves the free-energy label by more than a typical measurement error, and most of its records carry no reported temperature. Supplementary Note~\ref{app:benchmark} quantifies both.

The hydrogenation collection allows a direct comparison with ChemAHNet, a network designed for this collection \citep{Cheng_2025}. On the 940 test reactions released with that study, RxnPFN assigns the favoured enantiomer correctly for 96.9\% of the signed reactions when the recorded product is supplied, and 89.7\% when its stereochemistry is removed (Supplementary Table~\ref{tab:components}); ChemAHNet was trained on reactant, solvent and catalyst strings only, and its own published measure is $R^2=0.591$ on the same 940 reactions. Product stereochemistry is not available before an experiment, so the product-free setting is the one to compare. There RxnPFN reaches $R^2=0.729$, and with the product constitution but no stereochemistry, the setting in which the intended connectivity is fixed while the major enantiomer is not, it reaches $0.716$; the two differ by $0.013$, an interval their bootstrap does not resolve. No network weight is fitted to hydrogenation, and the only quantities fitted are label-free projections of the input, so the gap indicates how much of what a network designed for this collection achieves can be recovered from measured reactions alone. Our own retraining of ChemAHNet, with checkpoint selection on validation reactions, reaches $0.492$ on the same test set; we compare against the reported $0.591$ because the released checkpoint and its selection procedure are not available (Table~\ref{tab:main}, footnote). How far the assignment depends on the product structure, rather than on the substrate and catalyst, is examined next.

\input{tex/figs/fig2.tex}

What a collection borrows from another depends on what its labels record. In the single-seed pooling comparison, combining the three collections in one context helped hydrogenation slightly ($0.811$ against $0.805$) and hurt the two magnitude-only collections ($0.738$ against $0.764$ for ketone reduction, $0.709$ against $0.716$ for the oxidations), which are therefore predicted better from their own reactions (Supplementary Fig.~\ref{fig:settings}). The gain from mixing signed measurements with unsigned ones is confined to the decomposition, which acts only on signed labels. Describing the chemical role of each component and the substrate-to-product change was what mattered, and that advantage held across new partitions of the data and across several pretrained tabular backbones (Fig.~\ref{fig:main}f and Supplementary Fig.~\ref{fig:resplits}), against tree ensembles, reaction graphs and language models trained on the same splits \citep{Xu2025RXNGraphormer,Sagawa_2025,Burns_2025}.

A selectivity value matters less to a chemist than the order in which to try catalysts. Grouping reactions in which substrate and conditions were held constant and only the catalyst varied, the catalyst ranked first by RxnPFN was the most selective measured one in most groups, and far more often than chance (Fig.~\ref{fig:main}g). Because nothing is trained, such a ranking for a new substrate is obtained by inference over the precedents alone (Fig.~\ref{fig:main}h). The cost of a new task is one table read: 13,181 context reactions applied to 1,501 queries take about twelve seconds on one GPU with every parameter frozen, against 153~s to fine-tune a reaction language model and roughly 2,660~s to train a graph network for 80 epochs (Supplementary Table~\ref{tab:cost}).

\subsection{Product configuration and the sense of asymmetric induction}
\label{sec:ablation}

Before a synthesis, the intended product connectivity is often known while the configuration of the major enantiomer remains to be determined. We first asked which structural features accounted for the accuracy on complete reaction records. Describing the atom pairs in each component and the environments gained or lost in the reaction improved selectivity predictions over a description of the molecules alone (Fig.~\ref{fig:ablation}a). Separating the magnitude of selectivity from its sense gave the largest improvement on the validation reactions, allowing all three collections to inform the magnitude even when the major enantiomer was unspecified.

\input{tex/figs/fig3.tex}

For olefin hydrogenation, the signed measurement distinguishes the two orientations of the olefin relative to the catalyst \citep{Cheng_2025}. The configuration of the recorded product therefore gives direct information about the sense of induction. Even with that information available, the strength of discrimination between the two faces was harder to predict (Fig.~\ref{fig:ablation}b).

In the single-run product-information control, withholding product stereochemistry reduced hydrogenation $R^2$ from 0.805 to 0.716. Removing the product altogether did not lower it further (0.729; the difference between the two ablations is 0.013, 95\% bootstrap interval $-0.001$ to $0.028$ on the 940 test reactions, so their ordering is not resolved), whereas removing the product stereochemistry for ExtraTrees lowered $R^2$ from 0.751 to 0.570. RxnPFN therefore outperformed ExtraTrees both when given the expected product connectivity and when given only substrates, catalysts and conditions. Removing product stereochemistry had little effect on the magnitude-only reduction and oxidation predictions (Fig.~\ref{fig:ablation}c and Supplementary Tables~\ref{tab:controls} and~\ref{tab:components}). Measurements near the upper limit of ee also contributed to the predictions, particularly for ketone reduction, where such records are frequent (Fig.~\ref{fig:ablation}d).

An exacting test of stereochemical consistency is to reflect the whole reaction. In an achiral environment, inverting every stereocentre should reverse the sense of induction and preserve its magnitude. For eligible hydrogenations, the measured precedents alone often gave the expected reversal, although the model predicted the reflected reactions less accurately than the originals. Adding the mirror images of the eligible training reactions with their signs reversed brought the predictions for the two series into close agreement. The predicted sense then reversed in 99\% of the mirrored hydrogenations (Fig.~\ref{fig:ablation}e--g).

Uncertainty about the major enantiomer also changes the shape of a prediction. A hydrogenation may be strongly selective even when either enantiomer remains plausible. We therefore combined the predicted magnitude distribution with its mirror image, weighting each by the probability of the corresponding sense of induction. The resulting intervals attained close to their nominal coverage on hydrogenation. Across the three collections, \rxndirect{} also gave smaller errors for reactions with narrower intervals (Fig.~\ref{fig:ablation}h--k).

\subsection{Adapting literature precedents to a new catalyst screen}
\label{sec:generalization}

The reach of reasoning by analogy is set by the precedents. The released split shares substrates between partitions, so it measures the choice of a catalyst for a substrate that has precedents, which is the task the predictor is built for. Grouping by substrate--product pair lowers pooled $R^2$ to $0.765$ on average across three folds (Fig.~\ref{fig:generalization}b). Predictions deteriorate further when an entire ligand family, substrate scaffold or hydrogenation paper is absent from the precedents: pooled $R^2$ falls to $0.611$ when ligand families are held out, $0.498$ when publications are held out and $0.460$ when substrate scaffolds are held out. The gain over ExtraTrees (0.558, 0.430 and 0.405) and over TabPFN (0.596, 0.441 and 0.403) persists in every grouped split, including the hardest: even where the nearest precedents are far away, the model that reads precedents stays ahead of the models fitted to the reaction class. Absolute accuracy nevertheless degrades as the precedents recede, which is the operating characteristic of any method that predicts from a table of examples rather than from parameters fitted to the class. On the chiral phosphoric acid benchmark of Zahrt and co-workers \citep{Zahrt_2019}, RxnPFN approached published methods for a new substrate or, with a full catalyst fingerprint, a new catalyst. Changing both partners together remained more difficult (Fig.~\ref{fig:external_gallery}f and Supplementary Fig.~\ref{fig:extrapolation}).

With all 13,181 labels, Table~\ref{tab:main} gives five-seed pooled $R^2$ of $0.856$ for RxnPFN and $0.841$ for \rxndirect{}. To isolate the contribution of the 1,016 validation labels, we compared both context sizes over the same three seeds (Supplementary Note~\ref{app:benchmark}). Using only the 12,165 fitting labels gave $0.839$ and $0.825$, respectively, drops of $0.016$ and $0.017$ in the paired comparison. A single run of ExtraTrees on RTF (800 trees) reached $0.794$ with those same fitting labels and 1,501 test reactions. With the label budget matched, RxnPFN still leads ExtraTrees, the strongest fitted baseline, by $0.045$ in pooled $R^2$.

On the OrgAIcat aldol and Michael addition collections, RxnPFN gave $R^2$ values of 0.737 and 0.802 and mean absolute errors of 0.280 and 0.249~kcal\,mol$^{-1}$, respectively, in random five-fold cross-validation. OrgAIcat reported $R^2$ values of 0.716 and 0.774 and errors of 0.282 and 0.265~kcal\,mol$^{-1}$ using ten-fold cross-validation \citep{OrgAIcat2026}. RxnPFN's errors increased on both collections when the test catalysts were absent from the context (Fig.~\ref{fig:external_gallery}a--c and Supplementary Table~\ref{tab:pilot-transfer}).

The encoding also carries beyond the benchmark, and two further collections separate what transfers from how much context it takes to profit from what transfers. On 220 enantioselective $\beta$-C(sp$^3$)--H activation reactions, a bond-forming class absent from the benchmark \citep{Hoque2025CSP3H}, the frozen model reached $R^2=0.570$ with a mean absolute error of $0.229$~kcal\,mol$^{-1}$ under random five-fold cross-validation. An ExtraTrees model trained on the same RTF coordinates and the same splits reached $0.566$ and $0.226$~kcal\,mol$^{-1}$, giving similar scores. Both were far above the context mean ($R^2=-0.013$). Each fold left about 170 precedents. On 1,224 reactions from a high-throughput screen of rhodium-catalysed hydrogenation \citep{Kalikadien2024}, the chemistry of the benchmark's largest collection but with a much denser table of precedents, the frozen model reached $0.811$ against $0.732$ for ExtraTrees, a margin of $0.079$ that held in every fold (Fig.~\ref{fig:external_gallery}d,e and Supplementary Table~\ref{tab:transfer-new}).

Together, these two collections clarify how context density shapes performance: the encoding transfers to an unseen bond-forming class even with roughly 170 precedents, and once a denser library of measured precedents is available, in-context inference outperforms fitting a tree model to the same records by a clear margin.

\input{tex/figs/fig4.tex}

The published Pd/Sadphos Heck studies illustrate what changes between campaigns \citep{Li_Zhang_2024,Singh_2025}. A ligand associated with high selectivity in the study in which it was most often used gave much lower selectivity under the conditions reported elsewhere (Supplementary Note~\ref{app:details}). With an entire paper held out, the historical measurements alone gave poorer predictions on average than the historical mean (Supplementary Fig.~\ref{fig:heck}). Adding four measurements from the held-out study reduced the mean absolute error from 18.1 to 5.9 on the reported selectivity scale. The remaining reactions were also better ranked on average. Correcting the historical predictions by the mean bias on those same four reactions recovered only part of the improvement (Fig.~\ref{fig:generalization}d,e,g and Supplementary Table~\ref{tab:matched-transfer}). In retrospective screens within these studies, because several reactions per paper lie close to the maximum selectivity, all policies reached the single highest-selectivity entry in a similar number of tests; within the screening budget, however, RxnPFN consistently recovered a larger fraction of the three most selective reactions (Fig.~\ref{fig:generalization}f,h).

\subsection{A common ligand or substrate as a reference}
\label{sec:smalldata}

Published ligand screens and substrate scopes often share a reference reaction. In the first sparse-data study of Gallarati et al.\ \citep{Gallarati_2026}, one series held the substrate fixed and the other held the ligand fixed. The authors' models, built from descriptors of catalytic intermediates and transition states, predicted absolute selectivities more accurately than a direct application of RxnPFN. Inspection of these errors prompted us to predict the change from a measured reaction sharing the ligand or substrate. This reduced the error below that of the best published model on the nine held-out reactions, which also agreed closely with a simple additive estimate (Fig.~\ref{fig:generalization}c and Supplementary Fig.~\ref{fig:smalldata}). The correction gave no improvement on the nickel cross-coupling case. Transfer also depended on which partner changed in other studies. A model accurate within a series of cobalt chiral acids was less accurate than the authors' model for new acids, while for palladium-to-nickel transfer the two models performed best in combination (Fig.~\ref{fig:external_gallery}g--i and Supplementary Note~\ref{app:campaigns}).

\input{tex/figs/fig_external_gallery.tex}

\subsection{Conversion and yield across catalyst series}
\label{sec:yield}

A ligand that favours one enantiomer may still give poor conversion. We therefore predicted conversion for ketone reductions and yield for asymmetric oxidations as separate outcomes, using the corresponding measured precedents. Near-complete conversion was common in KetRed, while the oxidation yields spanned a broader range (Fig.~\ref{fig:yield_main}a). Within each collection, RxnPFN predicted the held-out outcomes more accurately on average than the tree ensembles, graph networks and reaction language models tested (Fig.~\ref{fig:yield_main}b,c and Supplementary Table~\ref{tab:yield_modern}).

\input{tex/figs/fig5.tex}

The transfer hydrogenation of a 1-naphthyl imidazolyl ketone to alcohol \textbf{19} illustrates how changing the catalyst affects conversion \citep{Zheng_2021_ATH}. Under otherwise identical conditions, the three-carbon-tethered Ru catalyst \textbf{2} gave 90\% conversion, whereas catalyst \textbf{5} gave only 26\%. RxnPFN correctly placed catalyst \textbf{4} between these extremes, although it underestimated the loss of conversion with \textbf{5} (Fig.~\ref{fig:yield_main}d). Across both collections, the prediction intervals contained the measured outcomes at close to their nominal frequency (Fig.~\ref{fig:yield_main}e), while signed errors pulled predictions towards the bulk of each distribution: low-conversion or low-yielding reactions were overpredicted on average and yields near 100\% were slightly underpredicted (Fig.~\ref{fig:yield_main}h).

As with selectivity, adding a few yields from a new Heck study improved predictions for the remaining reactions. In retrospective searches of these studies, choosing the next reaction by predicted yield typically found the highest-yielding combination sooner than random selection (Fig.~\ref{fig:yield_main}f,g).

\subsection{Transfer to cobalt hydrogenation and Buchwald--Hartwig amination}
\label{sec:newdata}

Hydrogenations with cobalt allowed us to ask what precedents from rhodium, iridium and ruthenium contribute when both the metal and the ligand families change. For a substrate held out of the cobalt collection \citep{Singh_2025}, the magnitude of enantioselection was best predicted by RxnPFN among the models compared. When few cobalt measurements were available, including the other metals' hydrogenations reduced the error relative to using the cobalt reactions alone or correcting the older predictions by a constant (Fig.~\ref{fig:newdata}f and Supplementary Table~\ref{tab:matched-transfer}). Holding out a whole cobalt publication was harder. Even with the other metals in the context, RxnPFN had a negative $R^2$, while ExtraTrees obtained a positive score.

We also examined iridium and rhodium hydrogenations from papers outside our benchmark. On the task defined by Singh and Hern{\'a}ndez-Lobato \citep{Singh_2025}, adding measurements from the new collection improved the identification of reactions reaching at least 80\% ee. Most reactions in this set already exceed that threshold. Performance was comparable to the authors' meta-learned model, although the two evaluations used different training data and product inputs (Fig.~\ref{fig:newdata}h and Methods).

In palladium-catalysed Buchwald--Hartwig amination, the outcome of interest is yield. The screen of Ahneman et al.\ systematically varies aryl halides, ligands, bases and isoxazole additives \citep{Ahneman_2018}. When the precedents sampled this reagent space, the frozen model reached $R^2=0.980$ with a root-mean-square error of 3.8 percentage points, ahead of the published tree, graph and language-model baselines on the same splits. Withholding entire groups of additives reduced accuracy, where task-trained models such as CheMeleon and Yield-BERT achieved slightly higher mean scores (Fig.~\ref{fig:newdata}g and Supplementary Table~\ref{tab:newdata-bh}) \citep{Probst_2022,Zhao2025BHRGNN}.

Across these reaction classes, neighbours in the structural fingerprint map tended to belong to the same class. In the representation produced by the pretrained network, the classes overlapped more, yet nearby reactions within a class had more similar measured outcomes. Thus structural resemblance and resemblance in reactivity gave different groupings of the same reactions (Fig.~\ref{fig:newdata}a--d).

\input{tex/figs/fig_newdata.tex}

\subsection{Chemical features and precedents that govern a prediction}
\label{sec:insight}

We examined how predicted selectivity responds to a change of functional group, alkene geometry or ligand, and which measured reactions contribute to an individual prediction (Fig.~\ref{fig:insight}). The expected directions of the structural effects were specified before evaluating the edited molecules.

On complete reaction records, the two halves of the problem draw on different parts of the input. For hydrogenation with recorded product stereochemistry, scrambling the substrate-to-product change largely removed the direction head's ability to read the favoured enantiomer, because that block carries the recorded product configuration relative to the olefin, while leaving selectivity magnitude predictable from the catalyst complex, ligand, transformation and solvent together; because the solvent also changes with the experimental protocol, its own contribution could not be separated. When product stereochemistry is withheld before featurization, as shown in the information controls above, the chiral catalyst and substrate still recover about 90\% of the induction sense (Supplementary Table~\ref{tab:components}). Ketone reduction behaves differently again: there the catalyst complex dominates magnitude prediction (Fig.~\ref{fig:insight}a). In aryl methyl ketones, the carbonyl carbon and the adjacent aryl and methyl carbons were particularly influential, placing the signal around the prochiral centre. In dehydroamino-acid esters it was concentrated at the prochiral $\alpha$-carbon, while the amide and ester groups showed little enrichment (Fig.~\ref{fig:insight}b,c). Such maps locate the parts of a structure on which the prediction depends \citep{Cheng_2025}.

Editing structures in silico tests those readings against mechanism (Fig.~\ref{fig:insight}d). Replacing the N-acetyl carbonyl of an enamide by methylene removes a site of rhodium coordination in the accepted hydrogenation mechanism \citep{Halpern1982}, and the predicted ee falls sharply. The control matters as much as the effect: homologation with the carbonyl retained also lowers the prediction, by about half as much, so the model responds to the acyl group in general rather than specifically to loss of a coordinating atom. Replacing aryl by cyclohexyl in aryl ketones likewise lowered predicted ee. Changing an $E$-olefin to its $Z$ isomer in a Sharpless dihydroxylation lowered predicted ee in the direction expected for the cinchona-derived ligands \citep{Kolb1994}. By comparison, exchanging rhodium and iridium at fixed ligand had little effect on the predictions. The literature records usually pair each metal with its own ligand families and provide few examples that isolate the effect of the metal.

\input{tex/figs/fig_insight.tex}

For a hydrogenation with a ferrocenyl P,N ligand, the most influential precedents differed only in the P-aryl substituents. The 3-fluorophenyl analogue raised the predicted selectivity, while difluoro- and dichlorophenyl analogues lowered it, in each case in line with the measurement. More generally, a prediction rests on a narrow and identifiable set of precedents. Removing the reactions of the same paper or ligand family roughly doubles the error, whereas removing the same number of arbitrary reactions changes almost nothing (Fig.~\ref{fig:insight}e,f). The model is therefore not averaging over the whole table; it is reading a few entries that a chemist would also have reached for. Errors were largest for rhodium hydrogenations of tetrasubstituted olefins and rose for substrates with few close precedents. Hydrogenation intervals widened as substrate similarity decreased, with broadly similar observed coverage on the original test split (Fig.~\ref{fig:insight}g--j and Supplementary Table~\ref{tab:novelty-coverage}). A separate limit arises at the most selective end of the scale. Many literature ee values are rounded to 99\%, where a small change in ee makes a substantial difference to $\Delta\Delta G^\ddagger$ (Fig.~\ref{fig:insight}k).

\subsection{How many precedents a prediction needs}
\label{sec:retrieval}

The published record of asymmetric catalysis grows without bound, whereas the cost of an in-context prediction grows with the length of the table the model reads. A library that keeps every measured reaction therefore becomes slower to query exactly as it becomes more informative. If a prediction is governed by its nearest precedents, the library should be searched rather than supplied whole, and retrieval would then earn its place twice: a shorter table is cheaper to read, and rows unlike the query no longer dilute its prediction. The benchmark's 13,181-reaction library is small enough to read in full, so what this setting measures is the second effect, on accuracy; the first follows whenever the length of the context is capped rather than fixed by the size of the library. For every test reaction we replaced the full context with the $K$ most similar measured reactions, ranked by cosine similarity in the RTF input, an arrangement used for general tabular data by \citet{Thomas2024LoCalPFN}, and compared it with $K$ reactions drawn at random from the same context. Both arms score the same queries at the same context size, so the difference measures which rows were chosen rather than how many (Supplementary Table~\ref{tab:retrieval}). The benchmark retrieval and component-ablation comparisons use scores from a single run per setting. For the selected RxnPFN model, Table~\ref{tab:main} averages five seeds for the pooled results and three for the per-dataset results.

On four collections, retrieval improved $R^2$ over the corresponding full-context run by at most $0.012$. On hydrogenation, 2,000 retrieved reactions, under a quarter of the 8,538 available, reached $R^2=0.817$ against $0.805$ for the whole collection, and 1,000 matched it at $0.807$. The oxidations matched at 1,000 ($0.719$ against $0.716$), and on the sampled query sets of Supplementary Table~\ref{tab:retrieval} the OrgAIcat Michael additions matched at 2,000 ($0.818$ against $0.816$) and the Buchwald--Hartwig yields at 2,000 ($0.984$ against $0.983$). Random selection of the same size stayed far behind: 1,000 random hydrogenations gave $0.566$ and even 4,000 gave $0.744$. Ketone reduction and the OrgAIcat aldol reactions were the exceptions, and narrowly so. The whole table remained best ($0.764$ against $0.760$ at 2,000, and $0.754$ against $0.752$ at 4,000 on the aldol query sample), but retrieved rows still beat random ones of the same size by $0.089$ and $0.033$ respectively. The same held when the three benchmark classes shared one library. With all 13,181 training reactions pooled, a query could retrieve hydrogenations, ketone reductions and oxidations alike, yet its nearest precedents came largely from its own class: 95\%, 89\% and 74\% of the 1,000 nearest, although those classes make up 65\%, 26\% and 9\% of the library. The similarity therefore recovers the reaction class without being told it. In the single-run pooled comparison, 4,000 retrieved reactions, 30\% of the library, gave $R^2=0.855$ against $0.855$ for all 13,181, with hydrogenation rising to $0.812$ from $0.811$ and ketone reduction to $0.742$ from $0.738$, and 1,000 retrieved reactions, 7.6\% of it, gave $0.848$. A random 1,000 gave $0.660$ and a random 4,000 gave $0.781$.

Similarity-based retrieval therefore decouples inference cost from total library growth whenever a query's precedents form a compact, well-separated neighbourhood---matching or exceeding full-table accuracy on four of the six collections while capping context length. Where the library holds little that a query does not already resemble, as for ketone reduction and the OrgAIcat aldol reactions here, the discarded rows carry information that their nearest neighbours do not replace, and the full table remains better. The overlap between a query's neighbourhood and the rest of the library separates the two cases, and it can be estimated before any prediction is made.

We compared RxnPFN with local regressors using the same RTF and FDE inputs and test splits. The regressors included cosine $k$-nearest neighbours with uniform or clipped cosine-similarity weights ($k=10$ to $1{,}000$), linear ridge and ridge on radial basis features. For each query set we report the highest observed test $R^2$ among the configurations evaluated (Supplementary Table~\ref{tab:local}). The best local scores were $0.574$ for hydrogenation, $0.612$ for ketone reduction and $0.687$ pooled, against $0.805$, $0.767$ and $0.856$ for RxnPFN. The gap was smallest for the oxidations ($0.675$ against $0.712$). What retrieval supplies is a conditioning set. The frozen network reads the retrieved reactions alongside the query and can recover full-table accuracy even where weighted averages of nearby labels fall well short.

\subsection{Selectivity in individual reactions}
\label{sec:cases}

Figure~\ref{fig:cases} places measured and predicted selectivities alongside the reaction structures, with examples selected across each class's error distribution. The largest discrepancies clustered in tetrasubstituted alkene hydrogenation, where assigning the wrong enantiomer can reverse the sign of a substantial ee. Ketone reductions and Heck reactions with measurements from the same study had the fewest large errors. For the Heck examples we retain the authors' reported selectivity scale. Individual conversion and yield predictions are shown in Supplementary Fig.~\ref{fig:cases-more-y}.

\input{tex/figs/fig6.tex}

%% file: tex/figs/fig2.tex
\begin{figure*}[p]
\centering
\figinclude[0.975\linewidth]{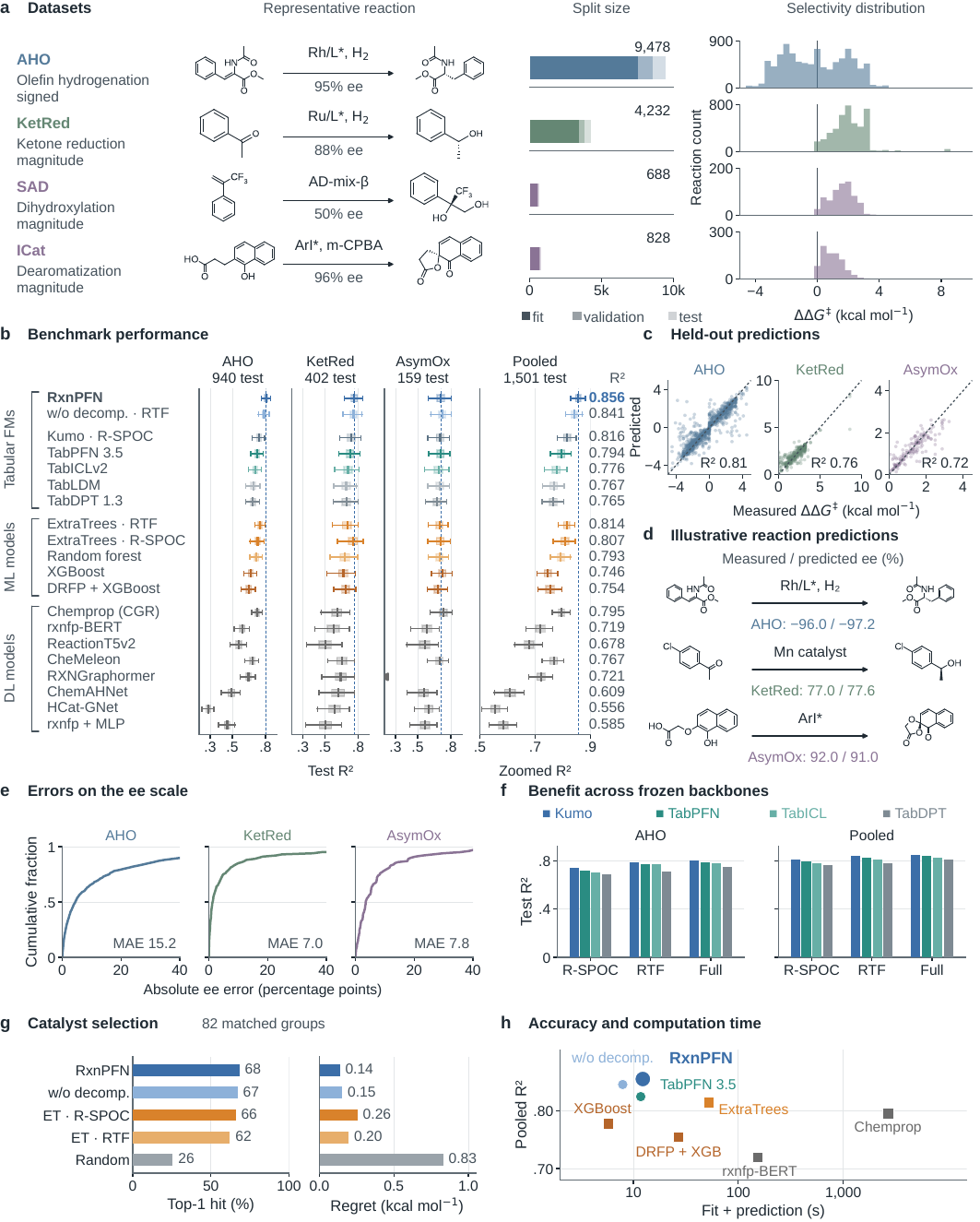}{figures/fig2_benchmark.png}
\caption{\textbf{RxnPFN predicts literature enantioselectivity and supports catalyst selection.}}
\label{fig:main}
\end{figure*}
\begin{figure*}[!t]
\ContinuedFloat
\captionsetup{font=footnotesize,singlelinecheck=false}
\caption[]{
\textbf{a}, Training reactions, fitting, validation and test allocations, and label histograms (0.4~kcal\,mol$^{-1}$ bins; vertical axes, reaction counts). AHO, olefin hydrogenation; KetRed, ketone reduction; SAD, Sharpless dihydroxylation; ICat, I(III) dearomatization. \textbf{b}, Test $R^2$ under per-dataset and pooled training, for tabular foundation models (FMs), conventional machine-learning (ML) models and deep-learning (DL) models. All four columns use 2,000 substrate--product splits. \textbf{c}, All held-out RxnPFN predictions; dashed lines, equality. \textbf{d}, Illustrative held-out reactions from AHO, KetRed and AsymOx; numbers are measured/predicted ee (\%). The AHO and KetRed examples were chosen from measured quantiles, the AsymOx example for its error. \textbf{e}, Cumulative absolute ee errors and mean absolute error (MAE). \textbf{f}, Test $R^2$ with R-SPOC, RTF and the full recipe across four frozen backbones. \textbf{g}, Catalyst ranking within 82 matched substrate/condition groups: top-1 hit rate and regret, the measured selectivity lost relative to the best catalyst. \textbf{h}, Pooled accuracy against measured fit-plus-prediction time (logarithmic axis) for 1,501 queries; circles, in-context models; squares, trained models.}
\end{figure*}

%% file: tex/figs/fig3.tex
\clearpage
\begin{figure}[H]
\centering
\includegraphics[width=0.99\linewidth,height=0.97\textheight,keepaspectratio]{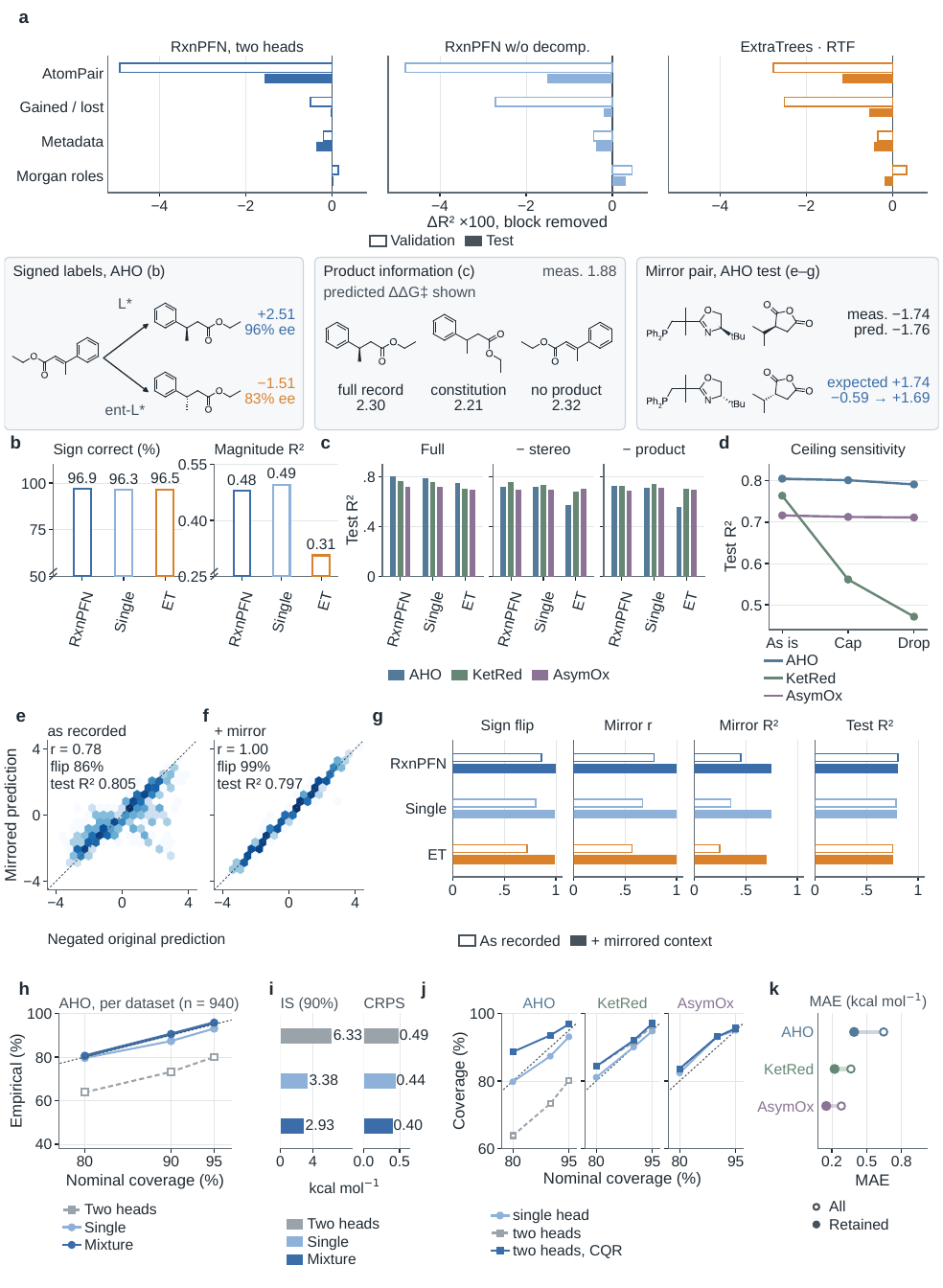}
\phantomcaption
\label{fig:ablation}
\end{figure}
\clearpage
\begin{figure}[H]
\ContinuedFloat
\captionsetup{font=footnotesize,singlelinecheck=false}
\caption[]{\input{tex/figs/fig3_caption.tex}}
\end{figure}
\clearpage

%% file: tex/figs/fig3_caption.tex
\textbf{Input information, mirror symmetry and predictive uncertainty.} Per-dataset results use only that dataset's training reactions or context; ``Single'' is RxnPFN without magnitude--direction decomposition.
\textbf{a}, Change in pooled $R^2$ ($\times100$) when one RTF block is removed; hollow and filled bars denote validation and test, the zero line the full input. Removing the atom-pair block costs $4.9$ and $4.8$ points on validation for the two-head and single-head models and $1.5$ points on test in both models, so the block contributes on both splits, more strongly on validation reactions than on this test set. The strip links the probes below to recorded reactions (selection rules in Methods): an AHO substrate with opposite signs for a ligand and its mirror image; one test reaction with the full, constitution-only and product-free inputs and their predicted $\Delta\Delta G^\ddagger$; and a strict mirror pair before and after mirror augmentation.
\textbf{b}, Sign accuracy on the 904 AHO test reactions with non-zero labels and $R^2$ of $|\Delta\Delta G^\ddagger|$; the axes are broken.
\textbf{c}, Test $R^2$ with the full record, product constitution without stereochemistry ($-$ stereo) and without product ($-$ product); magnitude-only datasets use RxnPFN's magnitude head.
\textbf{d}, Test $R^2$ of RxnPFN with context labels as recorded, capped at 99\% ee, or with records at $|ee|\ge99.5\%$ dropped; test membership is unchanged.
\textbf{e,f}, Predictions for mirrored reactions against the negated original predictions, for the 501 eligible AHO test reactions, before (e) and after (f) adding 4,463 mirrored training reactions with negated labels to the context. Hexagon intensity is log count.
\textbf{g}, The same comparison for all three models, before (hollow) and after (filled): fraction of signs that flip, correlation with the negated original prediction, $R^2$ against the negated label, and test $R^2$ on the originals.
\textbf{h}, Empirical against nominal interval coverage on AHO (1,458 calibration, 940 test reactions) for the clipped two-head readout, the single head and the sign mixture $p(y)=\pi M(y)+(1-\pi)M(-y)$, with $\pi=(1+\widehat{s})/2$.
\textbf{i}, Interval score at 90\% and CRPS for those readouts; lower is better.
\textbf{j}, Per-dataset coverage of the single head, the native two-head readout and the two-head readout after CQR.
\textbf{k}, MAE over all test reactions (hollow) and over those with a single-head 90\% interval narrower than the dataset's calibration median (filled). Energies are in kcal\,mol$^{-1}$.

%% file: tex/figs/fig4.tex
\clearpage
\begin{figure}[H]
\centering
\includegraphics[width=0.99\linewidth,height=0.97\textheight,keepaspectratio]{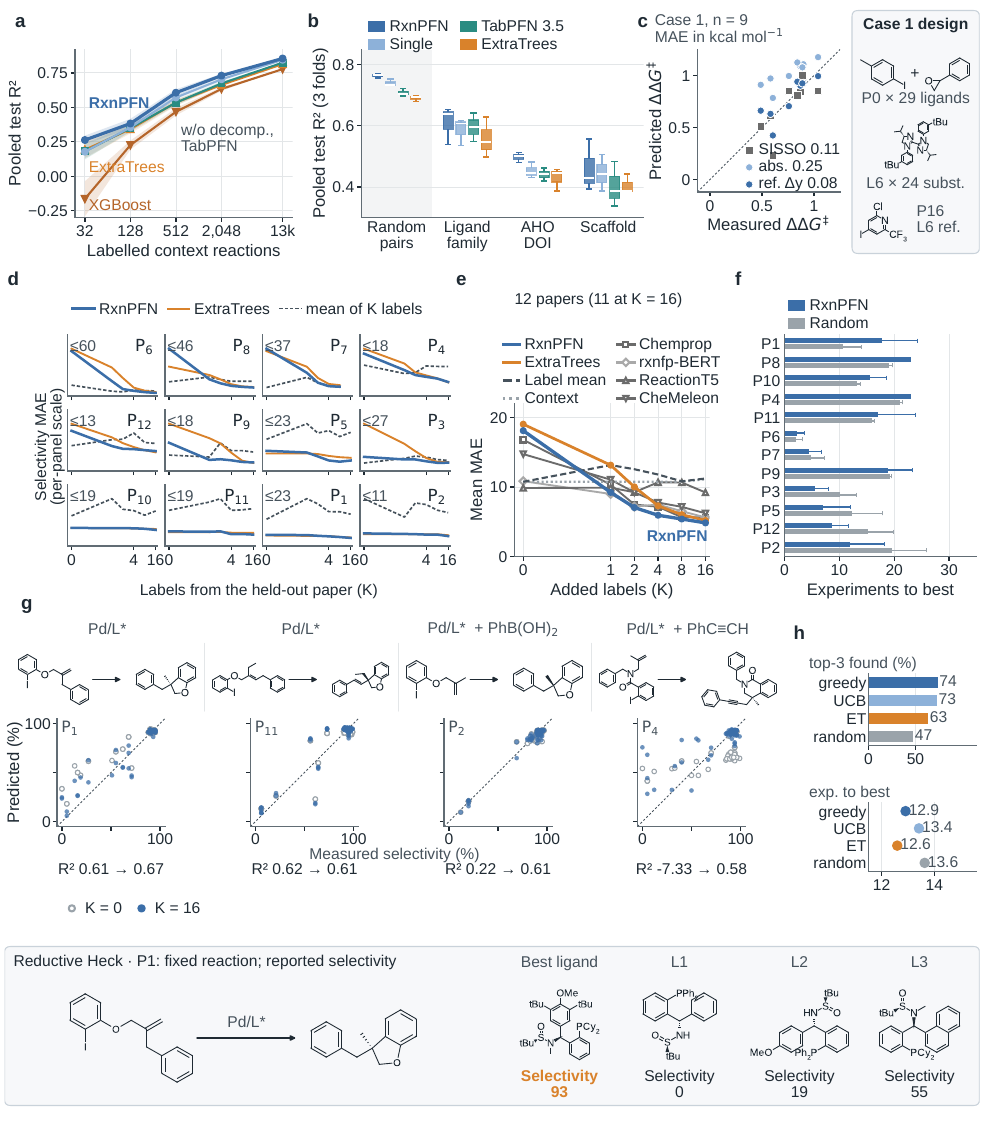}
\phantomcaption
\label{fig:generalization}
\end{figure}
\clearpage
\begin{figure}[H]
\ContinuedFloat
\captionsetup{font=footnotesize,singlelinecheck=false}
\caption[]{\input{tex/figs/fig4_caption.tex}}
\end{figure}
\clearpage

%% file: tex/figs/fig4_caption.tex
\textbf{Transfer to new chemistry and adaptation within reaction campaigns.}
\textbf{a}, Pooled test $R^2$ against labelled context size, using the RTF input; means and s.d. over five source-stratified random subsets, with the full 13,181-reaction context evaluated once.
\textbf{b}, Pooled test $R^2$ for random substrate--product-pair splits and splits holding out ligand families, AHO publications or substrate scaffolds; boxes summarize three folds. Single, RxnPFN without magnitude--direction decomposition; the shaded group is the random-pair reference.
\textbf{c}, Case~1 of Gallarati et al.\ \citep{Gallarati_2026}: predicted versus measured $\Delta\Delta G^\ddagger$ for the nine official out-of-sample reactions, comparing the published SISSO predictions with frozen-model predictions of the absolute target and of the reference-relative response $\Delta y$; legend values are MAE in kcal\,mol$^{-1}$.
\textbf{d}, Sadphos Heck adaptation, one panel per held-out paper: selectivity MAE on twelve fixed queries (six for paper 217) as $K$ measured reactions from that paper are added to an otherwise cross-paper context; ten random orders, with the upper bound of each panel's own scale printed as $\leq N$. Papers are ordered by MAE at $K=0$ and labelled P1--P12 in the order 164, 180, 181, 200, 210, 213, 217, 237, 250, 251, 258 and 264.
\textbf{e}, Mean selectivity MAE across twelve papers through $K=8$ and eleven at $K=16$; the deep-learning baselines use three of the ten label orders.
\textbf{f}, Simulated screening within each paper: experiments needed to find its best measured reaction under greedy RxnPFN and under random selection, as means and s.d.\ over ten random starts, papers ordered from the largest to the smallest advantage of random selection. When the best reaction was not reached within the screening budget, the bar is drawn one test beyond it; this affects RxnPFN for P4 and P8, and no random run. Per-paper differences lie within the spread across starts, and the pooled comparison is in \textbf{h}.
\textbf{g}, Parity before ($K=0$, hollow) and after ($K=16$, filled) adaptation for P1, P11, P2 and P4, the papers with mean $K=16$ $R^2\ge0.5$; each panel reports the mean $R^2$ at the two context sizes. The gain is confined to the two papers (P2 and P4) whose historical predictions were worse than the mean of the context labels; for P1 and P11, which transferred reasonably well from the start, the added labels change little.
\textbf{h}, The same campaigns summarized across papers: the fraction of each paper's three most selective reactions recovered within budget, and the mean experiments to the best, for greedy and UCB RxnPFN, ExtraTrees and random selection.
The strip below illustrates ligand screening within P1: the fixed reductive-Heck reaction with the best measured ligand and L1--L3, ordered from lowest to highest selectivity (1st, 5th, 8th and 12th of the 12 ligands tested, spanning 0 to 93 on the source-reported scale).

%% file: tex/figs/fig_external_gallery.tex
\begin{figure*}[p]
\centering
\figinclude[0.975\linewidth]{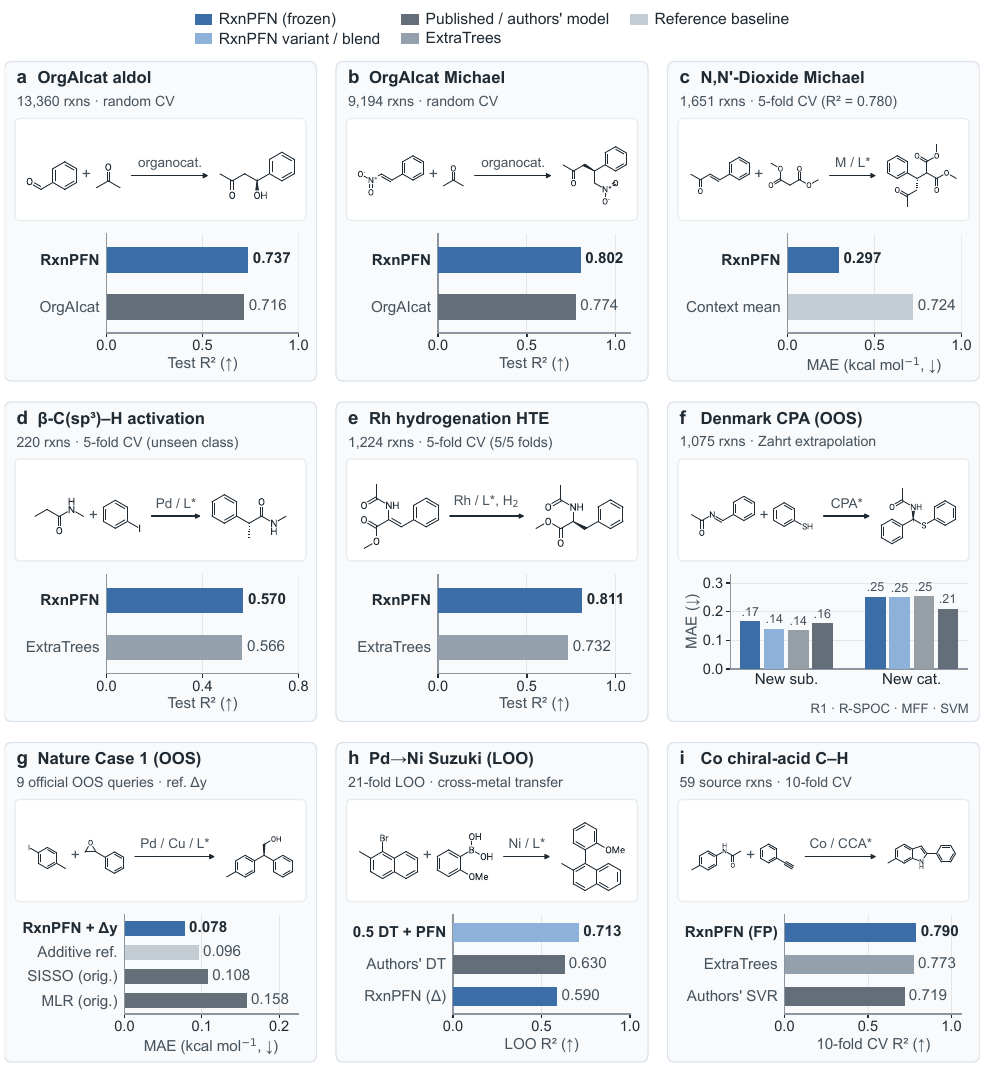}{figures/fig_external_gallery.png}
\caption{\textbf{Cross-reaction transfer, extrapolation and sparse-data benchmarks outside the main collections.} Legend on the next page.}
\label{fig:external_gallery}
\end{figure*}
\begin{figure*}[!t]
\ContinuedFloat
\captionsetup{font=footnotesize,singlelinecheck=false}
\caption[]{Representative reaction schemes and direct comparisons between frozen RxnPFN and published or fitted baselines across nine external catalytic systems.
\textbf{a}, OrgAIcat organocatalytic aldol collection (13,360 reactions): random five-fold cross-validation $R^2$ for RxnPFN versus the published ten-fold cross-validation score of OrgAIcat \citep{OrgAIcat2026}.
\textbf{b}, OrgAIcat organocatalytic Michael addition collection (9,194 reactions), compared on the same basis \citep{OrgAIcat2026}.
\textbf{c}, $N,N'$-dioxide/metal-catalysed Michael additions (1,651 reactions, random five-fold cross-validation; $R^2=0.780$): MAE in kcal\,mol$^{-1}$ against the context-mean baseline \citep{TangMichael2026}.
\textbf{d}, Enantioselective $\beta$-C(sp$^3$)--H activation (220 reactions, random five-fold cross-validation): an unseen bond-forming class comparing RxnPFN and ExtraTrees on the same RTF coordinates \citep{Hoque2025CSP3H}.
\textbf{e}, High-throughput rhodium-catalysed asymmetric hydrogenation screen (1,224 reactions, random five-fold cross-validation): RxnPFN leads ExtraTrees in all five folds \citep{Kalikadien2024}.
\textbf{f}, Denmark chiral phosphoric acid (CPA) thiol-addition extrapolation benchmark (1,075 reactions, Zahrt splits \citep{Zahrt_2019}): MAE (kcal\,mol$^{-1}$) on new substrates and new catalysts for RxnPFN with uncompressed role counts (R1), frozen Kumo on R-SPOC, and published MFF \citep{Sandfort_2020} and ASO/AEIF-SVM \citep{Zahrt_2019} models (full curves in Supplementary Fig.~\ref{fig:extrapolation}).
\textbf{g}, Nature sparse-data benchmark Case~1 (nine official out-of-sample reactions \citep{Gallarati_2026}): MAE (kcal\,mol$^{-1}$) for RxnPFN with the reference-relative response $\Delta y$, the training-only additive reference formula, and the published SISSO and MLR models.
\textbf{h}, Pd-to-Ni transfer in atroposelective Suzuki--Miyaura cross-coupling (21-fold leave-one-out \citep{Xu2025NiTransfer}): $R^2$ for the authors' decision-tree delta model, frozen RxnPFN with author RDKit descriptors, and their fixed 50/50 ensemble.
\textbf{i}, Cobalt chiral-acid catalysis (59 source reactions, ten-fold cross-validation \citep{Zhang2023ChiralAcid}): $R^2$ for frozen RxnPFN and ExtraTrees on raw role fingerprints against the authors' 27-feature SVR; on the ten unseen target acids, the authors' delta SVR remains stronger ($R^2=0.775$ versus $0.267$, Supplementary Table~\ref{tab:smalldata}). Full metrics and catalyst-held-out transfer controls are in Supplementary Tables~\ref{tab:pilot-transfer}, \ref{tab:transfer-new} and~\ref{tab:smalldata}.}
\end{figure*}

%% file: tex/figs/fig5.tex
\begin{figure*}[p]
\centering
\includegraphics[width=0.99\linewidth,height=0.98\textheight,keepaspectratio]{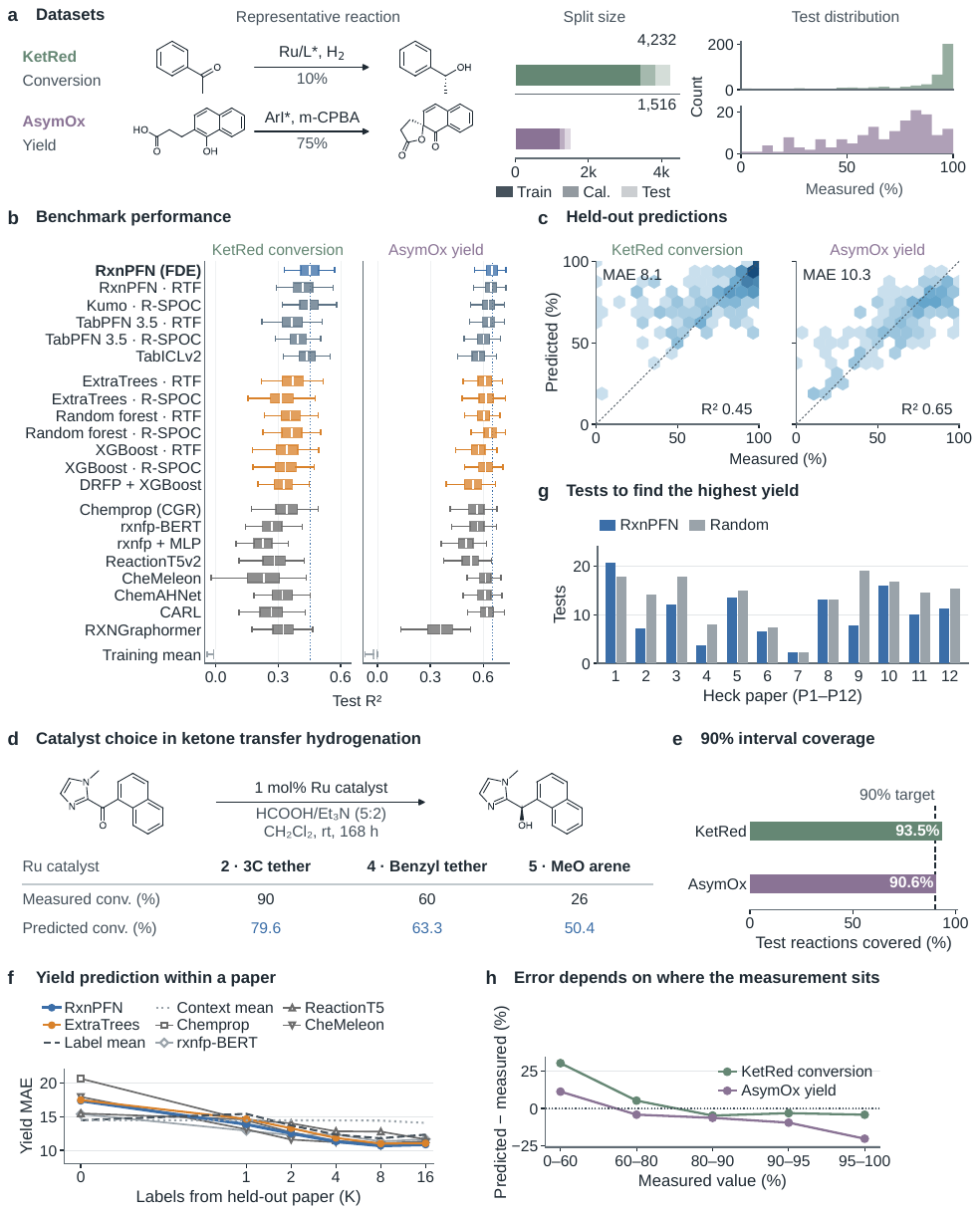}
\phantomcaption
\label{fig:yield_main}
\end{figure*}
\begin{figure*}[!t]
\ContinuedFloat
\captionsetup{font=footnotesize,singlelinecheck=false}
\caption[]{\textbf{Predicting conversion and yield from published reactions.}
\textbf{a}, KetRed conversion and AsymOx yield. Each training example uses the most frequent substrate, its most frequent catalyst and the record nearest their median outcome; histograms, test reactions in five-percentage-point bins.
\textbf{b}, Test $R^2$ across 2,000 bootstrap samples of substrate--product pairs shared by all models within a task. Boxes, interquartile range; white lines, median; whiskers, 2.5th and 97.5th percentiles; dotted lines, the observed RxnPFN scores.
\textbf{c}, RxnPFN predictions against measured test outcomes; darker hexagons hold more reactions on a shared logarithmic scale. MAE is in percentage points.
\textbf{d}, Asymmetric transfer hydrogenation of a 1-naphthyl imidazolyl ketone to alcohol \textbf{19} in Zheng et al.\ \citep{Zheng_2021_ATH}: the three $(R,R)$ Ru catalysts for this substrate and product in the KetRed test split, at 1~mol\% in formic acid/triethylamine (5:2) and dichloromethane for 168~h. Catalyst numbers follow the source; measured conversions sit beside the frozen predictions.
\textbf{e}, Native 90\% intervals contain 376 of 402 KetRed conversions and 144 of 159 AsymOx yields; dashed line, nominal coverage.
\textbf{f}, Yield MAE over held-out Sadphos Heck papers given $K$ measured yields from each paper (twelve papers through $K=8$, eleven at $K=16$).
\textbf{g}, Simulated Heck screens: mean tests to reach each paper's highest-yielding reaction under greedy RxnPFN or random selection, from the same two random reactions and up to twenty more, over ten starts per paper. Unsuccessful runs are scored one test beyond the budget.
\textbf{h}, Mean signed error (predicted minus measured, in percentage points) across bins of the measured outcome on the KetRed conversion and AsymOx yield test sets.
}
\end{figure*}

%% file: tex/figs/fig_newdata.tex
\begin{figure*}[p]
\centering
\figinclude[0.975\linewidth]{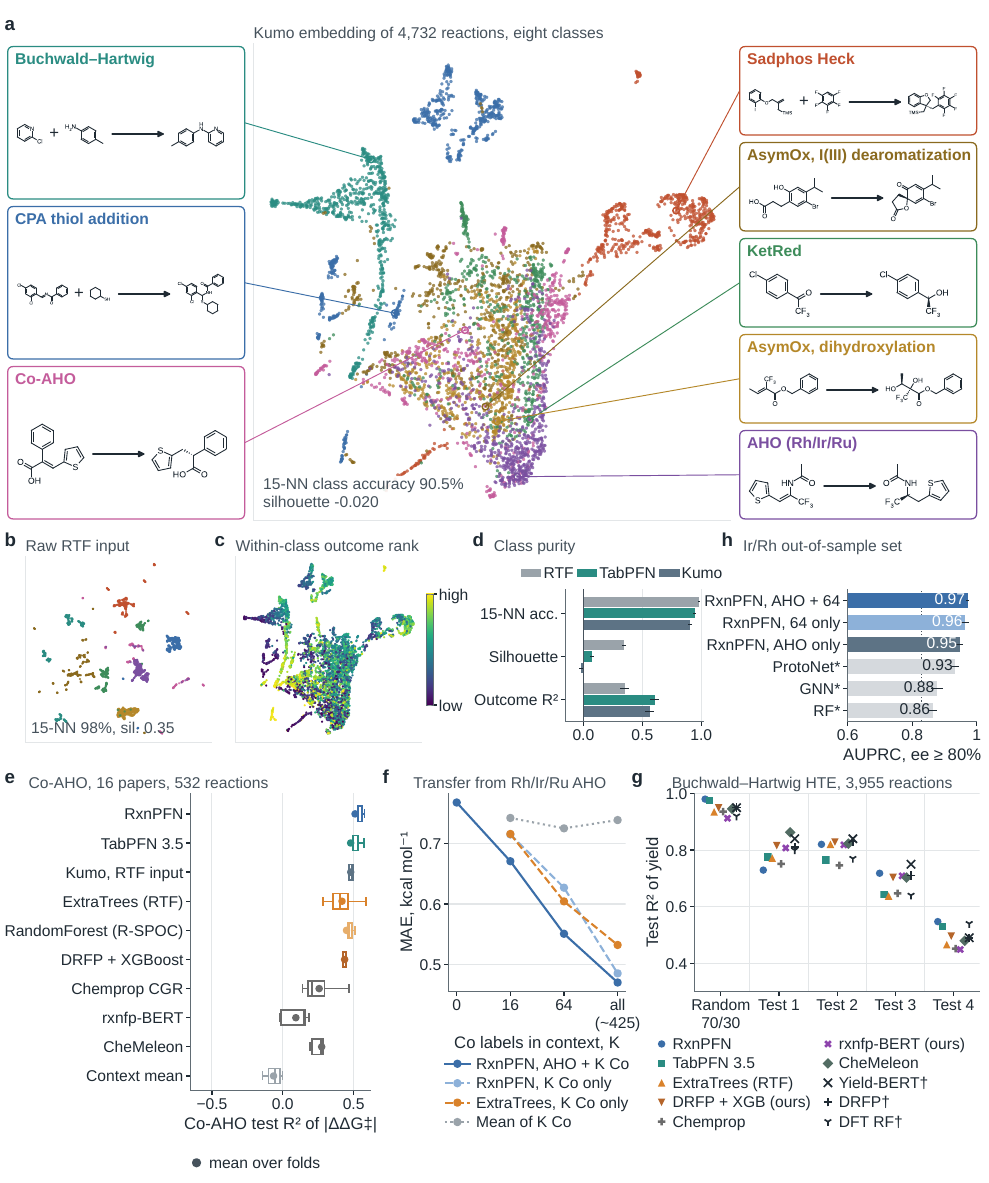}{figures/fig_newdata.png}
\caption{\textbf{Reaction similarity across classes and transfer to cobalt hydrogenation and Buchwald--Hartwig amination.} Legend on the next page.}
\label{fig:newdata}
\end{figure*}
\begin{figure*}[!t]
\ContinuedFloat
\captionsetup{font=footnotesize}
\caption[]{ \textbf{a}, Kumo-Tabular embeddings of 4,732 held-out reactions from eight public classes, generated in five class-stratified folds with outcomes z-scored within each class. Each scheme is the class medoid, connected to the same reaction on the map. \textbf{b}, The same reactions in the raw role--transformation fingerprint. Numbers, leave-one-out 15-nearest-neighbour class accuracy and cosine silhouette. \textbf{c}, The map of \textbf{a}, coloured by outcome rank within the class. \textbf{d}, Class purity and outcome structure of the two embeddings and of the raw input, with cluster-bootstrap 95\% intervals. \textbf{e}, Co-AHO ($|\Delta\Delta G^\ddagger|$, 532 reactions, 16 papers): test $R^2$ over five substrate-grouped folds, using each fold's Co training reactions and 9,358 non-Co AHO precedents. Boxes, median and interquartile range; dots, mean over folds. \textbf{f}, Co-AHO mean absolute error against the number of Co labels, with and without the AHO precedents. \textbf{g}, Buchwald--Hartwig HTE (3,955 reactions, yield): test $R^2$ on the ten standard random splits and the four additive out-of-sample tests; crosses, published values on the same splits. \textbf{h}, AUPRC at 80\% ee on the 246-reaction out-of-sample Ir/Rh hydrogenation set (bars, mean over ten draws; lines, 95\% interval); grey bars, published models; dotted line, prevalence of the positive class. Full numbers in Supplementary Tables~\ref{tab:newdata-co}--\ref{tab:newdata-bh}.}
\end{figure*}

%% file: tex/figs/fig_insight.tex
\begin{figure*}[p]
\centering
\figinclude[0.96\linewidth]{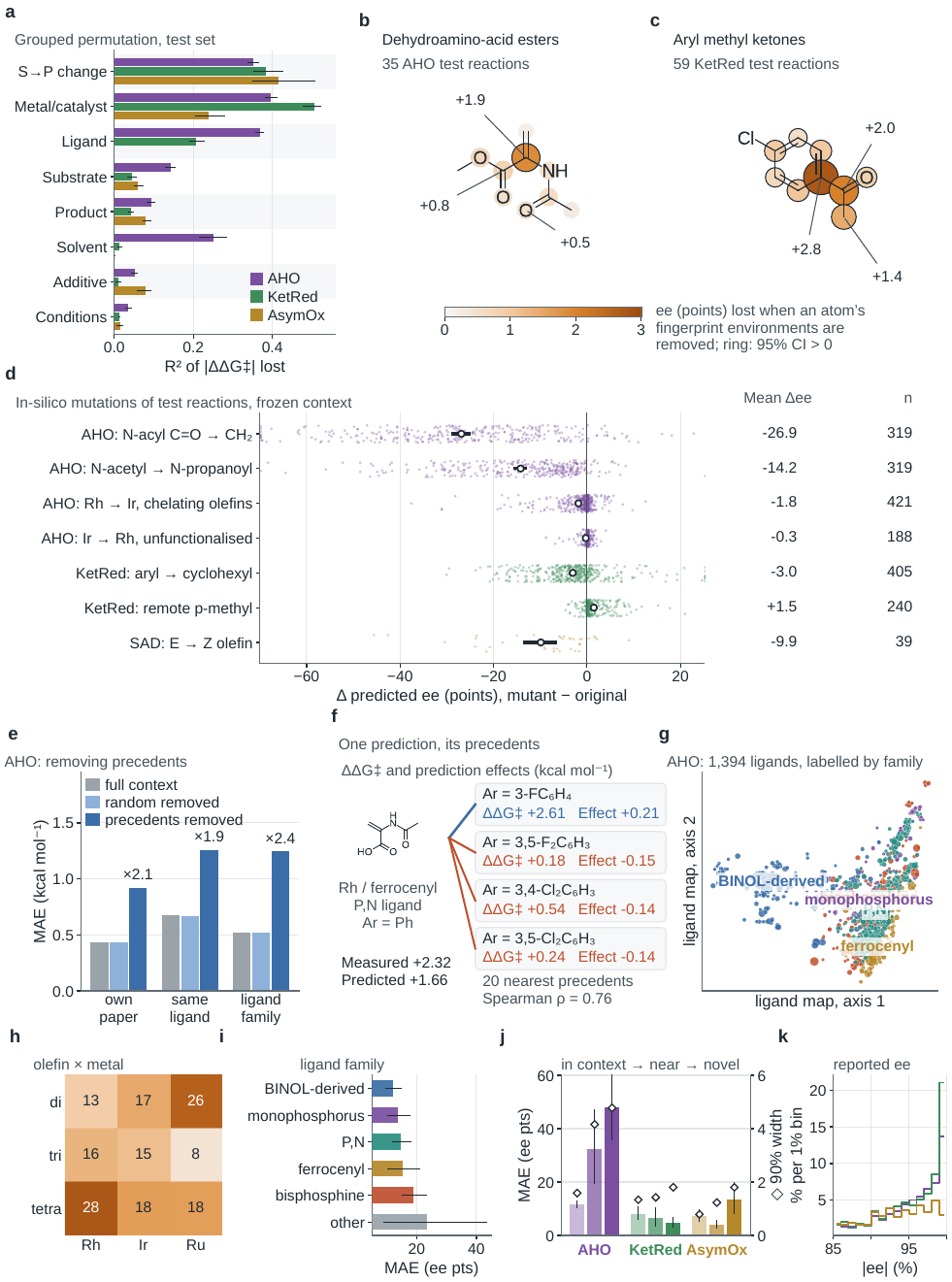}{figures/fig_insight.png}
\caption{\textbf{Structural features and measured precedents that influence predicted enantioselectivity.} Legend on the next page.}
\label{fig:insight}
\end{figure*}
\begin{figure*}[!t]
\ContinuedFloat
\captionsetup{font=footnotesize}
\caption[]{ Per-dataset RxnPFN models (one context per dataset); network weights never change, and counterfactual and attribution results describe the model's beliefs, not experiments.
\textbf{a}, Grouped permutation importance: decrease in test $R^2$ of $|\Delta\Delta G^\ddagger|$ when all RTF columns of one role are permuted (ten permutations; bars, mean; lines, 2.5--97.5\% range), one row of bars per role and one colour per dataset. With recorded product stereochemistry supplied, permuting the change block costs 45 points of AHO sign accuracy whereas every other role leaves it within 0.5 points, because the change block encodes the recorded product configuration.
\textbf{b,c}, Atom attribution over all test reactions containing the fragment (AHO methyl dehydroamino-acid esters, $n=35$; KetRed aryl methyl ketones, $n=59$), drawn on the most frequent member; the scale between them gives the ee lost. Every Morgan, atom-pair and torsion count of an atom is removed from substrate and product, the reaction re-encoded and re-predicted; colour, mean predicted ee lost (ring, bootstrap 95\% interval above zero).
\textbf{d}, In-silico mutations of every eligible test reaction, predicted with the same frozen context; points are reactions, white circle and bar the mean and bootstrap 95\% interval, and the right columns give the mean ee change and the number of reactions. The N-acetyl to N-propanoyl and remote $p$-methyl edits are controls.
\textbf{e}, Removing from the AHO context the reactions of a query's own publication (40 papers, 374 queries), the same ligand (40 ligands, 299 queries) or the same ligand family (6 families, 940 queries) roughly doubles the error of the affected queries, whereas the same number of random removals has no effect (paired Wilcoxon $p=1.8\times10^{-12}$, $7.8\times10^{-11}$, $0.031$).
\textbf{f}, A median-error AHO test reaction and the four largest effects among its 20 closest precedents, ranked by substrate and ligand Tanimoto similarity. All four come from the same paper and differ only in the P-aryl group; each box gives the precedent's measured $\Delta\Delta G^\ddagger$ and its effect on the prediction, and the effect tracks the residual of the precedent (Spearman $\rho=0.76$).
\textbf{g}, AHO ligand space: training-fitted Morgan coordinates projected to two dimensions; marker area $\propto\sqrt{\text{reactions}}$, coloured and labelled by ligand family, with the same colours in \textbf{i}.
\textbf{h}, AHO test error by olefin substitution and metal (cells with $\geq5$ reactions).
\textbf{i}, AHO test error by ligand family (bootstrap 95\% intervals).
\textbf{j}, Error (bars, bootstrap 95\% interval) and mean 90\% predictive-interval width (diamonds, right axis) for test reactions whose substrate is in the context, near (nearest-context Morgan Tanimoto similarity at or above the median of the rest) or novel.
\textbf{k}, Distribution of reported $|\mathrm{ee}|$ above 85\% in the three literature datasets (1\% bins).}
\end{figure*}

%% file: tex/figs/fig6.tex
\begin{figure*}[tbp]
\centering
\figinclude[0.90\linewidth]{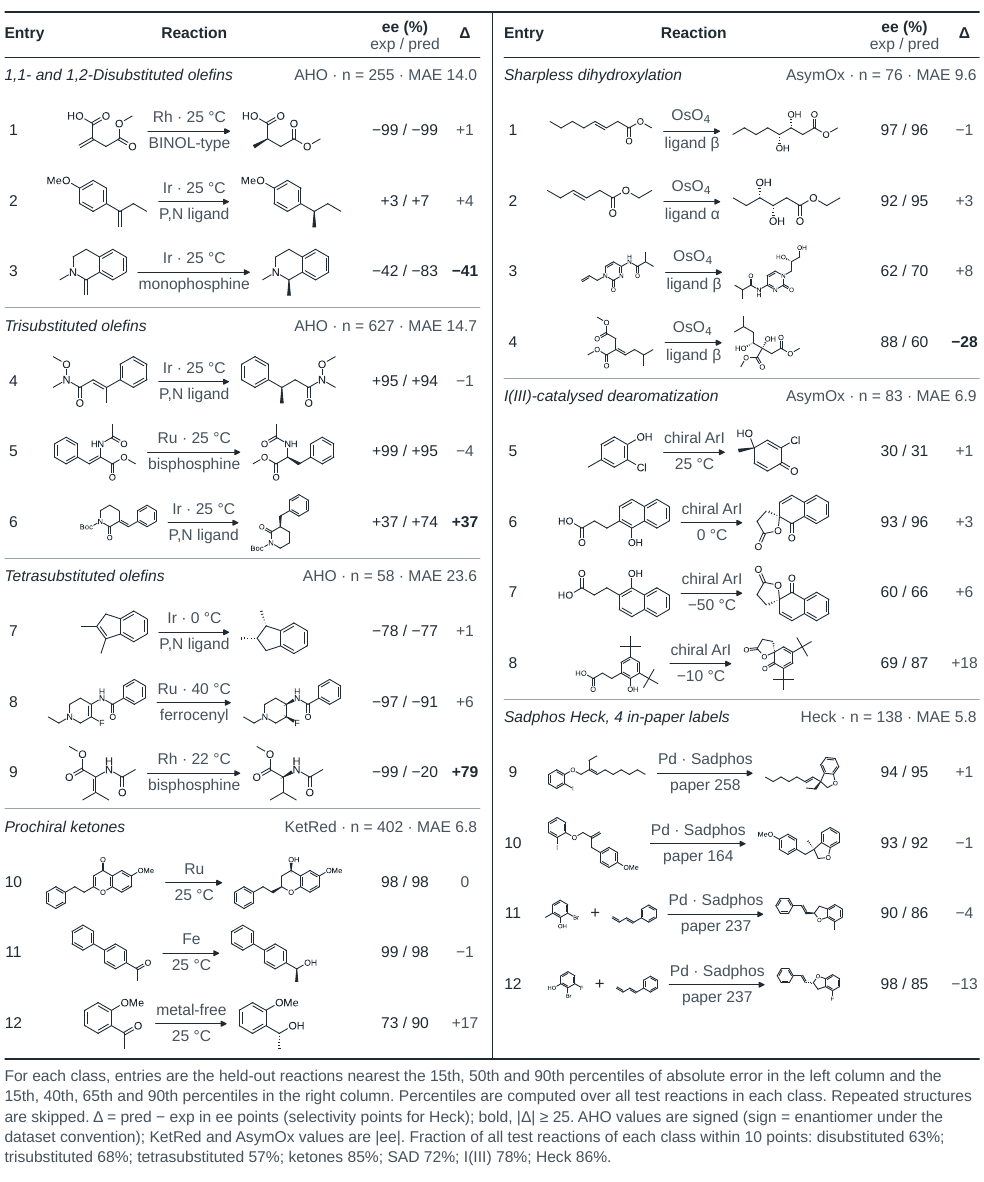}{figures/fig8_cases_ee.png}
\caption{\textbf{Measured and predicted selectivity of individual reactions.} Held-out reactions from seven reaction classes, three per class and four in the shorter right-hand column, chosen by a fixed rule given in the footnote below the table. For the benchmark classes the prediction is RxnPFN's pooled $\Delta\Delta G^\ddagger$, converted to ee at each reaction's recorded temperature (25~$^\circ$C when none is recorded); the Sadphos Heck class shows Kumo with four labelled reactions from the held-out paper in the context, on the reported selectivity scale. Structures are taken from the literature sources in Supplementary Note~\ref{app:data}.}
\label{fig:cases}
\end{figure*}

%% file: tex/4_discussion.tex
The result that generalizes is not a selectivity value but the procedure that produced it. A frozen model, a reaction described by the chemical roles of its components and by the bonds it changes, and a readout matched to what the label records together give held-out literature predictions that exceed models trained for that purpose, in seconds with fixed network weights. A method that predicts from precedents cannot reach beyond them, and we have treated that as a property to be measured rather than as a defect of the implementation. Its accuracy is a function of how close a query sits to its precedents, and that distance can be estimated before a prediction is relied on. For a field that produces data one reaction at a time, this is the useful form of a model: one that is re-pointed at a new campaign by adding a few rows rather than by retraining.

The published record of asymmetric catalysis can serve as a set of precedents for choosing a catalyst. In olefin hydrogenation, records that specify the product configuration encode the observed major enantiomer directly; when a planned synthesis specifies only the product connectivity, or only the starting materials, catalyst and conditions, the chiral catalyst and substrate still identify the favoured enantiomer in most cases. The harder question is how strongly the catalyst will discriminate between the two faces of a substrate. Here the identities of the ligand, metal complex and substrate all contribute.

A useful precedent can be as specific as a change in the P-aryl group of a ferrocenyl ligand. In the hydrogenation examined here, removing a fluorinated analogue changed the prediction in a direction that could be traced to its measured selectivity. Other comparisons expose the limits of that reasoning. The enamide carbonyl participates in rhodium coordination, and removing it lowers predicted selectivity, but a related edit that retains the carbonyl also lowers the prediction. Likewise, the historical choice of ligand is so closely tied to the metal that the records give little guidance on exchanging rhodium for iridium at fixed ligand.

The first measurements on a new system can change which precedents are informative. In the Sadphos Heck studies, they improved the ordering of reactions within a paper as well as correcting differences in average selectivity between papers. In a scope table with a common ligand or substrate, a measured reference reaction provided a useful starting point for estimating the effect of changing its partner.

These results also point to the experiments that would make the published record more informative. Complete ligand screens retain poorly selective combinations alongside highly selective ones. Reporting ee at the measured precision matters most for the most selective reactions, where rounding has the greatest effect on the inferred free-energy difference. For yield, records of nominally identical reactions can differ substantially. Reporting the experimental details behind both high and low yields would make it easier to judge which precedents apply to a new synthesis. As curated reaction databases and automated screening campaigns continue to expand, a training-free predictor that reads new measurements directly into context offers a practical path towards quantitative, continuously updated catalyst selection across asymmetric synthesis.

%% file: tex/5_methods.tex
\subsection{Datasets}
\label{sec:setup}
The benchmark combines published measurements of asymmetric olefin hydrogenation, ketone reduction and oxidation (Table~\ref{tab:data}). For hydrogenation we use the records of Hong and co-workers \citep{Xu2021AHO} with the signed $\Delta\Delta G$ values and the test partition released with ChemAHNet \citep{Cheng_2025}. KetRed comprises reductions of prochiral ketones to secondary alcohols. AsymOx comprises Sharpless asymmetric dihydroxylations \citep{Ocampo2025} and hypervalent-iodine-catalysed oxidative dearomatizations of phenols and naphthols. Both collections report the magnitude of selectivity. We convert ee to $\Delta\Delta G$ at the reported reaction temperature, using 25~$^\circ$C where the temperature is unavailable (2,586 KetRed records). Sources, structure checks and file provenance are given in Supplementary Note~\ref{app:data}. The Sadphos Heck collection and the published sparse-data studies used in the adaptation and transfer experiments are described with those experiments below.

\begin{table}[t]
\centering\small
\caption{\textbf{Benchmark datasets.} Validation is used only for model selection; final predictions use the final training context.}
\label{tab:data}
\begin{tabular}{llrrrrl}
\toprule
Dataset & Reaction class & Records & Final train & Validation & Test & Label\\
\midrule
AHO & Asymmetric olefin hydrogenation & 9,478 & 8,538 & 1,016$^{a}$ & 940 & signed $\Delta\Delta G$\\
KetRed & Asymmetric ketone reduction & 4,232 & 3,430 & 400 & 402 & $|\Delta\Delta G|$\\
AsymOx & Dihydroxylation; dearomatization & 1,516 & 1,213 & 144 & 159 & $|\Delta\Delta G|$\\
\midrule
Total & & 15,226 & 13,181 & 1,560 & 1,501 & \\
\bottomrule
\end{tabular}

\smallskip
\footnotesize $^{a}$Held out from the AHO training pool by substrate--product pair for selection and returned to the context afterwards.
\end{table}

\subsection{Problem setup}
A reaction record is $R_i=(S_i,P_i,M_i,L_i,V_i,A_i,\mathbf c_i)$, where $S,P,M,L,V,A$ are the substrate, product, metal-containing catalyst species, ligand, solvent and additive, and $\mathbf c_i$ collects the reaction conditions. Its selectivity label is the free-energy difference
\begin{equation}
 y=RT\log\frac{1+e}{1-e},\qquad e=\tanh\!\left(\frac{y}{2RT}\right),
 \label{eq:selectivity}
\end{equation}
where $e\in(-1,1)$ is the enantiomeric excess as a fraction under a fixed enantiomer ordering. Datasets that report only $|e|$ determine $|y|$ but not its sign; AHO labels are signed, whereas KetRed and AsymOx labels are magnitudes. We keep the stored labels exactly and never impute missing signs. Given a labelled context $\mathcal D=\{(R_i,y_i)\}_{i\in\mathcal I}$ and a query reaction $R_q$, the task is to predict $y_q$ without training any model parameters on $\mathcal D$.

\subsection{Role--transformation fingerprint}
\label{sec:representation}
Let $\mathbf m_r\in\mathbb N^{2048}$ be a radius-3, chirality-aware Morgan count fingerprint of role $r$ and $\psi(v)=\operatorname{sign}(v)\log(1+|v|)$. For each role we fit a truncated SVD on the training reactions and keep 32 coordinates:
\begin{equation}
 h_{\mathrm{role}}=\mathop{\bigoplus}_{r\in\{S,P,M,L,V,A\}}\operatorname{SVD}_{r,32}\!\big(\psi(\mathbf m_r)\big)\in\mathbb R^{192}.
\end{equation}
To make the transformation visible, we split the product-minus-substrate difference $\mathbf d=\mathbf m_P-\mathbf m_S$ into gained and lost environments and compress them jointly,
\begin{equation}
 h_{\mathrm{change}}=\operatorname{PCA}_{96}\!\big(\log(1+[\max(\mathbf d,0),\max(-\mathbf d,0)])\big)\in\mathbb R^{96}.
 \label{eq:change}
\end{equation}
These blocks describe hashed environments, not an atom-mapped reaction centre. Chirality-aware AtomPair counts of the six roles and of the product-minus-substrate difference add $7\times24=168$ SVD coordinates ($h_{\mathrm{pair}}$), and 19 metadata columns encode temperature, source, structure validity and parsed conditions with explicit missingness flags ($h_{\mathrm{meta}}$). The resulting \textbf{role--transformation fingerprint (RTF)}, named for its six chemical roles and the substrate-to-product transformation, is
\begin{equation}
 h(R)=h_{\mathrm{role}}\oplus h_{\mathrm{change}}\oplus h_{\mathrm{pair}}\oplus h_{\mathrm{meta}}\in\mathbb R^{475}.
 \label{eq:representation}
\end{equation}
We also use a 435-dimensional \textbf{fingerprint--descriptor encoding (FDE)} that combines a 192-coordinate PCA of the R-SPOC descriptor (ECFP, MACCS and RDKit properties aggregated over catalyst, conditions and product-minus-substrate) with compact Morgan, AtomPair and topological-torsion role blocks and metadata. All projections and imputers are fitted on the permitted training reactions only, separately for every setting, fold and dataset; implementation details are in Supplementary Note~\ref{app:implementation}.

\textit{Rank control.} A projection fitted on a small context can retain components whose singular values are numerical noise; reactions outside the context then project onto these components with very large coordinates, which destabilizes in-context models (trees are unaffected). Every SVD and PCA therefore keeps only components whose singular value exceeds $10^{-6}$ times the largest, and afterwards every column whose standard deviation over the context is below $10^{-6}$ times the largest column standard deviation is removed. The rule uses no labels and was fixed before scoring. On the main benchmark it changes no reported result (validation $R^2$ within $0.0004$ per dataset). It removes the failure of the frozen model on the Zahrt out-of-catalyst splits, where 24 context catalysts give about 40 distinct molecules. In the Pd-to-Ni study, preprocessing with rank control raises delta leave-one-out $R^2$ from $-0.552$ to $0.311$. In a separate input comparison, replacing the external SVD with raw inputs raises $R^2$ from $-0.381$ to $0.305$ (Supplementary Table~\ref{tab:hong-debug}). A transductive variant that also fits the projections on query structures was evaluated and not adopted (Supplementary Note~\ref{app:benchmark}).

\subsection{In-context prediction with a frozen tabular model}
We use the released Kumo-Tabular Large regressor \citep{Kumo2026,KumoAPI2026} with eight preprocessing estimators and a fixed set of seeds. Each representation coordinate is a table column. Cell embeddings use learned Fourier features, projection and missingness indicators; context labels receive linear embeddings, whereas query labels are not encoded. A six-stage table encoder alternates attention across features within a reaction with column-wise attention across reactions, using context keys and inducing points for column attention. Four learned summary tokens form each row representation. The $4\times5$ grid in Fig.~\ref{fig:method}c is schematic and denotes three context rows, one query row and five feature tokens. Context row representations also receive a label embedding before the separate 24-layer in-context Transformer. Each layer applies RMSNorm and attention followed by a residual addition, then RMSNorm and a GELU feed-forward network followed by a second residual addition. All rows supply attention queries, but only labelled context rows supply keys and values. Context and query representations are updated through the stack. A final normalization and prediction head return 999 quantiles $\widehat Q_q(\tau_j)$, $\tau_j=j/1000$, of the predictive distribution $p_\theta(y_q\mid h_q,\mathcal D)$. Selectivity and yield or conversion use separate passes through the frozen model, and predictions are averaged over the eight preprocessing estimators. The point prediction is the quantile-grid mean
\begin{equation}
 \widehat y_q=\frac{1}{999}\sum_{j=1}^{999}\widehat Q_q(\tau_j).
 \label{eq:point}
\end{equation}
The parameters $\theta$ are never updated. We refer to this predictor, which uses a single head on the RTF, as \textbf{\rxndirect{}}; it serves as the ablation of the next step.

\textit{Illustrated reaction and context.} The query in Fig.~\ref{fig:method}a--c is the KetRed test reaction whose absolute error in $|\Delta\Delta G|$ under per-dataset RxnPFN is closest to the median across all 402 test reactions, restricted to drawable records with all chemical roles recorded and the most frequent KetRed precatalyst family, [Ir(cod)Cl]$_2$. It is the asymmetric hydrogenation of 2-hydroxy-1-(thiophen-2-yl)ethan-1-one to the (\textit{R})-1,2-diol, with a ferrocenyl P,N,N ligand, K$_2$CO$_3$ and 2-propanol; temperature and hydrogen pressure are shown as recorded. The observed values are 99.0\% ee and 99.0\% conversion, and the predictions are 98.6\% ee and 96.0\% conversion. Prediction uses all 3,430 KetRed training reactions. For the seven examples in the figure, we averaged the percentile ranks of ee and conversion among eligible training records and selected distinct drawable substrates whose average ranks were closest to 0.10, 0.23, 0.36, 0.50, 0.64, 0.77 and 0.90. Ties were resolved by substrate similarity to the query and then by record identifier.

\subsection{Magnitude--direction decomposition}
\label{sec:signmag}
A mixed collection contains signed labels $y=\pm|y|$ and magnitude-only labels $|y|$. Regressing $y$ directly treats every magnitude-only label as a positive signed label and asks a single head to learn both how selective a reaction is and which enantiomer it favours. RxnPFN instead uses two context heads with the same frozen weights:
\begin{itemize}[leftmargin=*,itemsep=1pt,topsep=2pt]
\item a \emph{magnitude head} with context $\{(h_i,|y_i|)\}$ over all training reactions, giving $\widehat m_q$;
\item a \emph{direction head} with context $\{(h_i,\operatorname{sign}(y_i))\}$ over signed training reactions only, giving $\widehat s_q$.
\end{itemize}
The soft readout is
\begin{equation}
 \widehat y_q=\begin{cases}\widehat m_q\cdot\operatorname{clip}(\widehat s_q,-1,1) & \text{if the query's dataset reports signed labels},\\ \widehat m_q & \text{if it reports magnitudes.}\end{cases}
 \label{eq:signmag}
\end{equation}
Because $\mathbb E[M\mid x]\,\mathbb E[S\mid x]\neq\mathbb E[MS\mid x]$ in general, Eq.~\eqref{eq:signmag} is an approximation with a clear task meaning rather than the mean of a joint model; the shrinkage of $\widehat s_q$ towards zero for uncertain directions reduces the squared-error cost of a wrong sign. A hard readout that uses $\operatorname{sign}(\widehat s_q)$ was evaluated alongside and was worse on validation ($0.770$ versus $0.792$). The two heads may use different inputs; the selected pipeline uses the FDE input for magnitude and the RTF input for direction. We call the resulting pipeline \textbf{\rxnpfn}. For a magnitude-only dataset modelled on its own, RxnPFN reduces to its magnitude head. More generally, the decomposition is switched off, and the single head used, when the minority sign makes up less than 10\% of the signed context labels; this threshold was fixed before scoring and does not affect any benchmark result.

\subsection{Reference-relative response}
\label{sec:delta}
Small selectivity campaigns are often designed as a ligand screen on one substrate followed by a substrate scope with one reference ligand $L_0$. If the reference reaction $(L_0,P_j)$ has been measured, we regress the deviation
\begin{equation}
 r(L_i,P_j)=y(L_i,P_j)-y(L_0,P_j),\qquad \widehat y(L_i,P_j)=\widehat r(L_i,P_j)+y(L_0,P_j),
 \label{eq:delta}
\end{equation}
with the same inputs, context membership and frozen weights; only the response changes. The reference value is a measured training label, never a query label. The transformation is applicable only when a measured reference exists for each query. To test whether it generalizes, we also fixed a single admission rule before scoring: the context catalyst observed with the most substrates serves as reference, and a query is admitted only if that catalyst was measured on its substrate.

\subsection{Mirror augmentation}
\label{sec:mirror}
The enantiomer of a reaction, obtained by inverting every stereocentre of every role, has the opposite signed selectivity and the same magnitude. For reactions whose catalyst carries assigned tetrahedral stereocentres and no possible biaryl axis or metallocene, we generate the mirror image with RDKit by inverting every tetrahedral centre of every role and add it to the context with negated signed label (magnitude labels unchanged); all label-free transforms are then refitted on the augmented context. For AHO this adds 4,463 of the 8,538 training reactions, giving a 13,001-reaction context. Axially and planarly chiral catalysts are not mirrored.

\subsection{Predictive intervals}
\label{sec:uq-method}
Native intervals at level $1-\alpha$ use the sorted quantiles $\ell_q=\widehat Q_q(\alpha/2)$ and $u_q=\widehat Q_q(1-\alpha/2)$. To test their calibration we hold out a calibration set and apply conformalized quantile regression \citep{Romano2019}: with scores $s_i=\max\{0,\ell_i-y_i,y_i-u_i\}$ on calibration reactions and $\widehat s$ their $\lceil(n+1)(1-\alpha)\rceil$-th order statistic, the adjusted interval is $[\ell_q-\widehat s,\,u_q+\widehat s]$. In the pooled setting the model uses a 9,739-reaction proper-fit context and 2,426 calibration reactions grouped by substrate--product pair within each dataset; all 1,501 test reactions are kept, and the smaller context lowers pooled test $R^2$ of \rxndirect{} to $0.797$ (two-head pipeline $0.814$). In the per-dataset setting each dataset uses only its own proper-fit and calibration reactions. For selective prediction, a query is accepted when its native 90\% interval is narrower than the median width on calibration reactions, a threshold fixed before test labels are read. \textit{Sign-mixture distribution.} For the two-head pipeline, scaling the magnitude quantiles by $\operatorname{clip}(\widehat s_q,-1,1)$ collapses the two possible directions into one distribution. We instead form the mixture
\begin{equation}
 p(y_q)=\pi_q\,M_q(y_q)+(1-\pi_q)\,M_q(-y_q),\qquad \pi_q=\tfrac{1}{2}\big(1+\operatorname{clip}(\widehat s_q,-1,1)\big),
 \label{eq:mixture}
\end{equation}
where $M_q$ is the distribution given by the 999 magnitude quantiles; in practice the mixture is represented by the $2\times999$ weighted atoms $\pm\widehat Q^{(m)}_q(\tau_j)$. Intervals are central intervals of this mixture; the point prediction remains Eq.~\eqref{eq:signmag}. A variant with Platt-calibrated $\pi_q$ fitted on calibration reactions was evaluated alongside. Interval quality is summarized by empirical coverage, the interval score \citep{Romano2019} and the continuous ranked probability score (CRPS), computed on the weighted atoms.

\subsection{Evaluation setup}
\label{sec:protocol}
For the main RxnPFN benchmark, we select the configuration by pooled mean-squared error on 1,560 validation reactions using a context of 12,165 fitting reactions. With the configuration fixed, we refit the input projections on the final context of 13,181 reactions and score predictions on the 1,501 held-out test reactions.

\textit{What the released split measures.} The AHO collection consists of ligand screens, in which one substrate is typically measured with many catalysts, and the partition is drawn per recorded reaction rather than per substrate. Of the 940 test reactions, 717 share a substrate and product with a training reaction, 814 share at least the substrate, and 354 share substrate and ligand. Of the test reactions, 151 repeat all eight modelled fields of at least one training record. For 98 of those reactions, at least one such training record also has the same reported ee. What the released split does not test is a substrate--product pair absent from the context altogether; the pair-grouped re-splits address this case and give a mean pooled $R^2=0.765$. In the \emph{per-dataset} setting, the main comparison, each dataset is modelled on its own: its training reactions are the only context (or training set) and every label-free projection is refitted on them. In the \emph{pooled} setting one model or context covers all three datasets, and only pooled metrics over all test reactions are reported for it. We report $R^2=1-\mathrm{SSE}/\mathrm{SST}$, MAE and RMSE in kcal\,mol$^{-1}$, and the macro average of the three per-dataset $R^2$ values. Pooled $R^2$ includes between-dataset variance: on the test set 27.0\% of the total sum of squares lies between dataset means, and predicting each dataset's training mean already gives pooled $R^2=0.268$. Confidence intervals come from a paired cluster bootstrap over substrate--product pairs (1,000 draws, and 4,000 for the contrasts that involve the selected pipeline), stratified by dataset when pooled. MAE in ee percentage points is computed after back-conversion with Eq.~\eqref{eq:selectivity} at each reaction's temperature. Seed variability is assessed over five model seeds, which change the estimator ensemble and the order of features. For re-split robustness, each dataset is additionally re-split five times at random, grouped by substrate--product pair and with the original test fraction, and all main arms are rerun per dataset; differences are summarized by their mean, a $t$-interval over the five splits and the number of wins. Backbone generality is tested by applying the RTF input and the magnitude--direction readout unchanged to TabPFN 3.5, TabICLv2 and TabDPT 1.3.

\subsection{Baselines}
Tree models (ExtraTrees, RandomForest, XGBoost; \citealp{Geurts2006,Breiman_2001,Chen2016}) use fixed hyperparameters with the R-SPOC descriptor or with the RTF. Pretrained in-context baselines (TabPFN 3.5, TabICLv2, TabDPT 1.3, Xiaomi-TabLDM and Kumo; \citealp{Hollmann2025,TabICL2026,TabDPT2026,Xiaomi2026,Kumo2026}) use frozen weights, eight estimators and the full context, with R-SPOC in their native form and after a shared 256-coordinate PCA. Reaction baselines are (i) the differential reaction fingerprint DRFP (2,048 bits, radius 3) with XGBoost in the published settings \citep{Probst_2022}; (ii) Chemprop \citep{Heid_2023} with a condensed-graph reaction (CGR) encoding of atom-mapped substrate-to-product reactions plus a reagent graph and temperature, trained for 80 epochs on the fitting reactions with the validation set used for checkpoint selection; (iii) the pretrained rxnfp BERT encoder \citep{Schwaller_Yield_2021} with a regression head, fine-tuned end to end on reaction SMILES with a standardized target and early stopping on validation, the learning rate chosen once on pooled validation from $\{2\times10^{-5},10^{-4}\}$; and (iv) an MLP on frozen 256-dimensional rxnfp embeddings with source indicators. Chemprop, rxnfp-BERT and the MLP are averaged over three seeds in both settings. Because Chemprop and rxnfp-BERT select their checkpoints on validation, they see fewer labelled reactions than the in-context models, which use all training reactions as context. On the small campaigns, Chemprop is a multicomponent message-passing network over the molecule graphs (these reactions have no CGR atom mapping), and both deep models use a fixed inner validation split of 15\% of the context; all settings were fixed before scoring. Complete baseline results are in Supplementary Note~\ref{app:baselines}.

\textit{Additional reaction baselines.} ReactionT5v2-yield \citep{Sagawa_2025} is fine-tuned end to end on reaction SMILES, with a standardized regression target and validation-based early stopping (at most 200 epochs). CheMeleon-initialized Chemprop v2 \citep{Burns_2025} uses three molecular components and temperature, for 50 epochs with validation checkpoint selection. Both use three seeds and are run separately under the pooled and per-dataset settings. ChemAHNet is retrained both on each collection and on the pooled fitting reactions using its released code and validation-selected checkpoints (three seeds). HCat-GNet \citep{AguilarBejarano2025} is evaluated in all four settings using the original labels and test memberships: three inner folds partition the non-test records, three validation-selected model predictions are averaged within each seed, and metrics are summarized across three seeds. We reuse the original AHO and KetRed graph inputs and inner-fold memberships; AsymOx uses the same substrate/catalyst/reagent roles with absent roles omitted. Its graph inputs omit the same two unparsable KetRed fitting structures as RXNGraphormer and retain every held-out reaction. Thus HCat-GNet uses inner-fold validation within the non-test records, whereas the other trained reaction models use the fixed fitting/validation split. Metric means over seeds are distinguished from metrics computed after averaging predictions.

\textit{RXNGraphormer for selectivity.} The released pretrained encoder \citep{Xu2025RXNGraphormer} is fine-tuned separately on AHO, KetRed, AsymOx and their pooled fitting set, using three seeds. We retain the four-layer GCN, four Transformer layers, width 256, two attention heads, batch size 32 and AdamW/Noam schedule of the yield reproduction below. Each run lasts 600 epochs; the checkpoint with minimum validation MAE is selected before test prediction. Targets are scaled using only the fitting-set minimum and maximum, then inverted to kcal\,mol$^{-1}$ without clipping. Inputs concatenate the substrate and available condition components on the reactant side, with the recorded product on the product side. Invalid condition components are omitted after fragment-level parsing checks. Two unparsable KetRed fitting structures are omitted; all 1,560 validation and 1,501 test records are retained. Unused gradient-norm diagnostics and per-epoch test evaluation are disabled without changing optimization.

\textit{Yield-specific deep baselines.} CARL \citep{Lu2025CARL} uses the released molecular featurization and network, with a repaired training loop that preserves the fixed data partitions. We use Adam ($10^{-3}$), batch size 64, 300 epochs, hidden dimension 256, the published prediction/equilibrium/reconstruction loss weights, validation-MAE checkpoint selection and ten Monte Carlo dropout passes at inference. Unknown atom vocabulary entries use a catch-all index; two unparsable KetRed fitting records are omitted by the featurizer, while all validation and test reactions are retained. RXNGraphormer \citep{Xu2025RXNGraphormer} starts from the released pretrained encoder and uses its Buchwald--Hartwig fine-tuning configuration: four graph-convolution and four Transformer layers, hidden dimension 256, two attention heads, batch size 32, AdamW, Noam scheduling and 600 epochs; the checkpoint with lowest validation MAE is scored. Unparsable condition components are omitted from the RXNGraphormer input without dropping test reactions. Both use three seeds and the fixed Y1 fit/validation/test allocations. Predictions are clipped to $[0,100]$ percentage points before scoring. These yield-specific runs predict conversion or yield; the separate selectivity evaluation is described above.

\textit{Buchwald--Hartwig graph baseline.} BH-RGNN \citep{Zhao2025BHRGNN} uses the released graph architecture and the authors' 100 masked Mordred descriptors per molecular role. On each of the ten standard random splits we train with three seeds for 2,000 epochs (Adam, $10^{-4}$, batch 64), selecting the epoch on a fixed 15\% holdout from the training context. BatchNorm statistics are recomputed on training inputs before evaluation, and inference uses evaluation mode. The test labels are never used to choose the epoch. We average metrics across seeds within a split, then report the mean and s.d. across the ten splits. The authors' internal 70/30 split is only a reproduction control and is excluded from this comparison.

\textit{Fig.~\ref{fig:main} examples and displays.} The training-reaction rule chooses the most frequent drawable substrate, its most frequent catalyst and the record nearest that group's median absolute selectivity. Histogram counts include every stored label. For product examples, products with at most 22 heavy atoms are eligible; the nearest measured label to each of the 1/6, 1/2 and 5/6 quantiles of a collection's test set is chosen, requiring distinct product structures after stereochemistry is removed and using row identifier to break ties. Predictions and errors do not enter this rule. Fig.~\ref{fig:main}d retains the first selected case in quantile order for AHO and KetRed. The AsymOx case follows the low-error selection described below. Each scheme shows its substrate and product with an abbreviated catalyst label. Measured ee is the recorded value, signed according to the benchmark target for AHO; predicted ee is $100\tanh(\widehat{\Delta\Delta G^\ddagger}/(2RT))$, using the recorded temperature (298.15~K for room-temperature or missing entries) and $R=0.0019872036$~kcal\,mol$^{-1}$\,K$^{-1}$. These schemes illustrate individual cases and do not specify every reaction condition. The full held-out scatter plots retain all test reactions. Catalyst-ranking and computation protocols are unchanged; repeated-split results are in Supplementary Fig.~\ref{fig:resplits}.

\subsection{Reproduction of ChemAHNet}
\label{sec:chemahnet}
ChemAHNet \citep{Cheng_2025} was retrained from scratch with the released code, since the released checkpoint was not available. Inputs are the released reaction strings (reactant, solvent and catalyst SMILES); the target is the signed $\Delta\Delta G^\ddagger$, which equals our benchmark label for all 9,478 reactions, and the test set is the released set of 940 reactions, identical to ours. The model uses the released hyperparameters (eight heads, seven layers, dropout 0.2, batch 128, Adam with learning rate $5\times10^{-4}$ and weight decay $10^{-5}$, linear warm-up over 500 steps, pretrained ChemBERTa token embeddings, at most 250 epochs with patience 80) and is trained on our 7,522 fitting reactions with the checkpoint selected by $R^2$ on the 1,016 validation reactions, over three seeds. The published value ($R^2=0.591$, RMSE $1.27$ kcal\,mol$^{-1}$) and the published per-reaction predictions, from which the sign accuracy is computed, are taken from the original Supplementary Information and data release. For the Zahrt thiol-addition splits we additionally ran the released script, which selects the checkpoint on the test set, alongside a variant with a 10\% inner validation split (Supplementary Note~\ref{app:benchmark}).

\textit{Figure 2 display examples and uncertainty.} The AHO and KetRed examples in Fig.~\ref{fig:main}d retain the earlier measured-quantile choices, which did not use prediction error. The AsymOx example was explicitly selected after evaluation to illustrate an accurate prediction: eligible test reactions have measured ee of 20--95\%, absolute ee error at most two percentage points, and at most 22 heavy atoms in each drawn substrate and product. Among the 29 eligible reactions, we select the one whose error is closest to their median, breaking ties by row identifier. It has measured ee 92.0\% and predicted ee 91.0\%. This outcome-selected illustration does not estimate average performance, and no model, context or test metric is changed by it. For each of the four columns in Fig.~\ref{fig:main}b, the same substrate--product-pair resamples are applied to every model; the pooled resamples are stratified by source. Metrics are computed separately for each saved seed prediction and then averaged within each resample; they are not computed from an across-seed prediction ensemble. Each box therefore describes resampling uncertainty on the fixed test reactions rather than only variability across three seeds.

\textit{Figure 3 drawn examples.} The three chemistry boxes in the strip of Fig.~\ref{fig:ablation} use fixed rules applied to the stored records and the saved predictions, never to the reported metrics. \emph{Signed labels.} Among AHO training records sharing a substrate, the most frequent eligible substrate is kept and the largest-magnitude record is drawn for each of two ligands that are mirror images in the stored ligand field. \emph{Product information.} The test reaction is the one whose increase in absolute error when the product is removed is closest to the median of that increase over AHO test reactions with all three inputs available; the product, its stereochemistry-free constitution and the substrate alone are drawn with their predicted $\Delta\Delta G^\ddagger$. \emph{Mirror pair.} The eligible test reaction whose label magnitude is closest to the median of the strict mirror probe is drawn with every tetrahedral centre inverted; the mirrored prediction is shown before and after mirror augmentation. These choices select illustrations after evaluation and change no model context, prediction or metric.

\textit{Reaction case tables.} Within each class, we compute absolute-error percentiles over all test reactions and draw the nearest eligible reaction to each percentile, skipping repeated structures. Eligible structures must parse and total at most 42 heavy atoms across the drawn species. Figure~\ref{fig:cases} uses the 15th, 50th and 90th percentiles on the left and the 15th, 40th, 65th and 90th on the right. Supplementary Figs.~\ref{fig:cases-more-a}, \ref{fig:cases-more-b} and \ref{fig:cases-more-y} use the 5th, 25th, 45th, 60th, 75th and 95th. Tables report stored predictions (from the pooled single-run RxnPFN evaluation for the benchmark selectivity classes, and per-task Kumo on the RTF input for conversion and yield). AHO signs follow the dataset's enantiomer convention. Fractions within 10 points use all test reactions in each class.

\subsection{Additional evaluations}
\label{sec:additional}
\textit{Information controls.} The product-constitution input removes all product atom and bond stereochemistry before featurization, and the product-free input removes the product role; all projections are refitted. \textit{Mirror test.} For test reactions whose catalyst carries assigned tetrahedral stereocentres and no possible biaryl axis or metallocene, every tetrahedral centre of the record is inverted, inputs are recomputed with the transforms fitted on the dataset's training reactions, and the reaction is predicted with the unchanged context. \textit{High-ee sensitivity.} Reactions with $|\mathrm{ee}|\geq99.5\%$ are flagged as clipped; we score without the clipped test reactions, cap all context labels at the 99\%-ee value of their temperature, or remove clipped reactions from the context. \textit{Grouped splits.} Three-fold splits hold out ligand families (generic Bemis--Murcko scaffold of the ligand or, if absent, of the metal catalyst), AHO publications (by DOI) or substrate scaffolds, with random three-fold splits over substrate--product pairs as the reference. \textit{Label budget.} Source-stratified random subsets of 32, 128, 512 and 2,048 training reactions, five per budget, serve as the context. \textit{Catalyst ranking.} Test reactions with identical dataset, substrate, product constitution, solvent, additive, temperature and pressure form a group; groups with at least three distinct catalysts are scored by $|\Delta\Delta G|$, with top-1 and top-3 hit rates, regret and Spearman correlation, against the analytical expectation of a random choice.

\textit{Overview application plots.} Figure~\ref{fig:method}d assembles the existing evaluations without changing their memberships. The B--H yield panel contains all 11,880 held-out predictions across ten random splits, with $R^2$ averaged over splits; catalyst selection uses 82 matched test groups. The AHO selectivity and KetRed conversion panels contain all 940 and 402 per-dataset test predictions, respectively, and their displayed $R^2$ and MAE are computed from the stored two-decimal outcomes. Parity hexagons encode log observation counts. The interval panel shows empirical versus nominal coverage of native central intervals for AHO selectivity with the sign-mixture readout, KetRed conversion and AsymOx yield, without post-hoc conformal recalibration. The AHO interval evaluation uses a separate 6,064-reaction proper-fit context and is distinct from the point-prediction evaluation. At 90\% nominal coverage, the three observed coverages are 90.6\%, 93.5\% and 90.6\%, respectively. The adaptation panel shows mean Sadphos Heck MAE over twelve held-out papers and ten label orders at zero, one, two and four added labels. The reference predicts the other-paper context mean at zero labels and the mean of the added labels thereafter. The percentage reduction is $100(1-\mathrm{MAE}_{4}/\mathrm{MAE}_{0})$.

\textit{Heck adaptation.} The Sadphos Heck collection is a compilation of 703 published Pd/Sadphos-catalysed Heck-type reactions from 13 papers (Sadphos ligand class; cited here via Singh et al.\ \citep{Li_Zhang_2024,Singh_2025}) after excluding 14 reactions without a reported value and one unparsable structure. Selectivity is modelled on its reported scale. For each paper with at least twelve reactions, the other papers form the base context, $K\in\{0,1,2,4,8,16\}$ labelled reactions of the held-out paper are added in ten label-independent nested random orders, and the held-out query set stays fixed across label budgets within each order (twelve reactions per paper, six for the 14-reaction paper). The mean of the $K$ added labels and the base-context mean are reference predictors.

\textit{Heck campaigns.} For each of the same twelve papers, a simulated campaign starts from two random reactions of the paper (ten starts per paper), adds the remaining reactions of the paper to a candidate pool with hidden labels, and lets a policy choose the next reaction to be revealed, up to 20 choices; the other papers stay in the context. Policies are greedy selection by the predicted value (frozen model or ExtraTrees on the RTF input), an upper confidence bound $\widehat y+(\widehat Q_{0.9}-\widehat Q_{0.5})$ from the native quantiles, and random order. A campaign that does not reach the paper's best reaction within the 22 revealed reactions is recorded as 23, one more than the budget, which is what the censored bars in Fig.~\ref{fig:generalization}f show. The primary metric, fixed beforehand, is the number of revealed reactions until the paper's best one is found; the recovered fraction of the three best reactions within the budget was added afterwards as a secondary metric. Policies are compared over papers with Wilcoxon signed-rank tests.

\textit{Yield and conversion.} Two literature endpoints use the same encodings and frozen weights: KetRed conversion and AsymOx reported yield with the benchmark splits; conversion is never relabelled as yield. We do not report yield or conversion for the AHO collection: its publication-disjoint splits contain few papers and near-saturated labels, and no model, including the training mean, gave a meaningful prediction. Because yield and conversion are bounded, predictive quantiles are truncated to [0,100]\%; this leaves coverage unchanged and narrows intervals. Replicate disagreement is the pooled standard deviation of reported values among reactions with identical substrate, product, catalyst, ligand, solvent, additive, temperature and pressure.

\textit{Published sparse-data studies.} For Gallarati et al.\ \citep{Gallarati_2026} we use their official memberships with their 2,268 raw mechanistic descriptors, the union of their selected features, or chirality-aware Morgan and AtomPair counts of ligand and substrate (106 coordinates for Case~1, no product). For the Pd-to-Ni study \citep{Xu2025NiTransfer} we reuse all 21 leave-one-out folds with their selected Pd support sets (120--141 reactions per fold), and for the Co study \citep{Zhang2023ChiralAcid} we use fixed ten-fold cross-validation of the 59 source reactions and leave-one-out over the ten target reactions. Each study uses only its own measured reactions as context.

\textit{Extrapolation stress test.} The Zahrt chiral phosphoric acid benchmark \citep{Zahrt_2019} is evaluated on its standard out-of-substrate, out-of-catalyst and combined splits (384 context reactions) and on the ChemAHNet hold-outs, reported in Supplementary Note~\ref{app:benchmark}.

\subsection{Cobalt hydrogenation, Buchwald--Hartwig and the embedding map}
\label{sec:newdata_methods}
\textit{Co-AHO.} The cobalt set is the 532-reaction Co-catalysed asymmetric olefin hydrogenation collection released with the meta-learning study of Singh and Hern{\'a}ndez-Lobato \citep{Singh_2025} (16 publications, 366 substrates, 59 ligands). Roles follow the AHO convention of the benchmark (substrate olefin, product, ligand, metal written as ligand.[Co$^+$], solvent, additive), and the target is $|\Delta\Delta G^\ddagger|$ computed from the reported ee at the reported temperature (the source ee carries no configuration). We evaluate five-fold cross-validation grouped by canonical substrate (substrate groups assigned at random) and leave-one-publication-out over the 16 papers. For transfer, the AHO reactions of the benchmark without Co catalysts are added to the context (9,358 reactions; none comes from the 16 Co papers), with all Co training reactions of a fold or with $K=16$ or $64$ of them (three random draws per fold), and with no Co label ($K=0$). Singh and Hern{\'a}ndez-Lobato score a binary label (ee $\ge 80\%$) by AUPRC; we report it alongside the regression metrics, using the ee implied by the predicted $|\Delta\Delta G^\ddagger|$ as the score. Their RF/GNN/ProtoNet scripts read cluster files from absolute paths of the authors' machine that are not released, so they could not be rerun; we quote their published values only where the task is the same.

\textit{Out-of-sample Ir/Rh set.} The official out-of-sample test set of the same study \citep{Singh_2025} contains 246 Ir- and Rh-catalysed hydrogenations (163 Ir, 83 Rh) from five 2022--2023 publications that are absent from the benchmark (no AHO reaction shares a DOI). Its product column is empty, so products were generated by hydrogenating the non-aromatic C=C bond (211 reactions); for the 35 chromones, which RDKit writes as aromatic rings, the C=C bond of the pyranone ring was reduced. The generator reproduces the recorded constitution of all 1,766 AHO substrates and of 530 of the 532 Co products. Products of the AHO context were reduced to their constitution as well, so that no stereochemical information enters either side. Following the published protocol, ten random draws provide 64 support and 128 query reactions; the support is added to the AHO context (or used alone), and AUPRC at 80\% ee is averaged over the draws. The published models are meta-trained on the Ir, Rh and Co reactions and adapted on the support set, whereas RxnPFN uses the AHO collection as context without training; the comparison therefore matches the query task but not the training data.

\textit{Buchwald--Hartwig.} The high-throughput set of Ahneman et al.\ \citep{Ahneman_2018} contains 3,955 Pd-catalysed couplings of 15 aryl halides with p-toluidine under 4 ligands, 3 bases and 22 isoxazole additives (3,955 of the 3,960 combinations). The role mapping is: substrate, aryl halide and p-toluidine; product, the C--N coupling product generated with the reaction template of Schwaller et al.\ \citep{Schwaller_Yield_2021}; catalyst, the Pd(II) precatalyst; ligand, the phosphine; solvent, DMSO (constant); additive, the base and the isoxazole. The target is yield (\%). We use the standard splits released with the Yield-BERT study \citep{Schwaller_Yield_2021}: ten random 70/30 splits (2,767 training, 1,188 test reactions) and the four additive out-of-sample tests (Tests 1--4; 3,057/3,055/3,058/3,055 training reactions; every test additive is absent from training). Published $R^2$ values are taken from Table 2 of the DRFP study \citep{Probst_2022}, which collects the DFT-descriptor random forest of Ahneman et al., Yield-BERT and DRFP on identical splits. Chemprop on this set uses a multicomponent graph network over the aryl halide and amine, the ligand and the base--additive pair, because no atom mapping is available.

\textit{Baselines on the new sets.} RxnPFN is run with its default recipe (FDE input with rank control; the decomposition rule selects the single-head readout because all targets are non-negative). TabPFN~3.5 uses the same input; Kumo is also run on the RTF. Tree models use the R-SPOC descriptor and the RTF with the settings of the benchmark; DRFP with XGBoost uses the yield settings of Probst et al.\ \citep{Probst_2022}. Chemprop (CGR reaction graph with reagents and temperature, or the multicomponent network above), the fine-tuned rxnfp-BERT regressor and CheMeleon-initialized Chemprop use the small-data recipes of the Heck experiments, with 15\% of the training reactions as validation for checkpoint selection, three seeds on the grouped Co folds and the additive tests and one seed on the other splits. Differences to RxnPFN are tested with a paired cluster bootstrap (2,000 draws; clusters are substrate--product pairs, or reactions for Buchwald--Hartwig).

\textit{Embedding map.} The maps show 4,732 reactions from eight public classes, with up to 600 per class (AHO, Co-AHO, KetRed, the two AsymOx subsets, Sadphos Heck, the Denmark CPA thiol additions and Buchwald--Hartwig). The original embedding run included 4,765 reactions across nine classes. The RTF is fitted to this pooled set without outcome labels. Each class's endpoint ($|\Delta\Delta G^\ddagger|$, the native Heck selectivity or the Buchwald--Hartwig yield) is z-scored within the class and serves as the common label. TabPFN~3.5 is fitted in five class-stratified folds, and the final-layer test-token embeddings (\texttt{get\_embeddings}) of the held-out fold are averaged over the eight estimators; for Kumo, the input of the in-context output head is averaged in the same way. UMAP (30 neighbours, minimum distance 0.3, cosine metric, fixed seed) is applied to the embeddings and, as the baseline, to the raw RTF. The overview map in Fig.~\ref{fig:method}d instead reuses a fixed t-SNE projection of the pooled Kumo embeddings (perplexity 30, cosine metric, PCA initialization, automatic learning rate, 1,500 iterations, Barnes--Hut method, the same seed). The projections use no class or outcome labels. For both maps we omit the 33 reactions from the additional class, retain the original coordinates of the eight public classes, and compute the following metrics on the displayed reactions. Class purity is measured in the full space by leave-one-out 15-nearest-neighbour accuracy of the class label and by the cosine silhouette. Outcome structure is measured by the $R^2$ of each reaction's within-class outcome against the mean outcome of its 15 nearest same-class neighbours. We exclude the query by row identity before selecting neighbours for either accuracy or outcome prediction, including when distances are tied. Intervals come from a cluster bootstrap (2,000 draws), paired between spaces. The structures in Fig.~\ref{fig:newdata}a are the TabPFN~3.5 class medoids (minimum summed within-class cosine distance), located by row identity on the Kumo map.

\textit{Computation.} All in-context predictions use one GPU and no gradient step; main point predictions were run on an H200 and some auxiliary experiments on an H100 with the same weights. Compute times are listed in Supplementary Table~\ref{tab:cost}.

%% file: tex/7_appendix.tex
\section{Data Sources}
\label{app:data}
\input{tex/dataset_sources}

\section{Implementation Details}
\label{app:implementation}

\paragraph{Representation.}
We used chiral Morgan count fingerprints of radius 3, folded to 2,048 positions. The gained/lost block concatenates the positive and negative parts of the product-minus-substrate difference into a 4,096-coordinate vector before the 96-coordinate centered PCA (no variance scaling). AtomPair fingerprints use 2,048 folded positions. Role-wise SVDs are fitted on the training reactions with each molecule weighted by its number of occurrences, which is equivalent at the covariance level to fitting on repeated reaction-level rows; the implementation computes a 64-coordinate basis and keeps the leading 32. Products keep their recorded stereochemical annotation; invalid or missing structures are encoded by validity flags rather than replaced. RDKit 2026.03.1 is used \citep{RDKit2026}.

\paragraph{Metadata.}
The 19 metadata columns derive from 20 definitions, including temperature, three source indicators, six structure-validity flags, five optional condition values (pressure, time, ligand loading, catalyst loading, transfer-hydrogenation text) and their five missingness flags---after removing one column that is constant on training reactions. The first numeric token of each optional condition is parsed, imputed with the training median and transformed by a signed logarithm. Catalyst identifiers, literature identifiers, row indices and outcome labels are never used as inputs.

\paragraph{FDE input.}
The 435-dimensional FDE input combines 192 PCA coordinates of the R-SPOC descriptor with Morgan, AtomPair and topological-torsion coordinates of the six roles and their product-minus-substrate view, plus metadata. R-SPOC contains ECFP, MACCS and RDKit-property blocks aggregated over catalyst, pooled conditions and product-minus-substrate.

\paragraph{Backbone.}
Kumo-Tabular Large is used as released, without modification. Kumo embeds feature groups with learned Fourier features and missingness projections and adds target embeddings to context cells only. Six stages alternate column-wise induced attention, which uses only context keys and values, with attention across the features of each row. Four 256-dimensional readout tokens form a 1,024-dimensional row representation, followed by 24 in-context Transformer layers with 16 heads in which context and query rows attend only to context rows. A two-layer GELU head outputs the 999 quantiles. These components follow the released implementation \citep{KumoAPI2026}; we do not modify them. Each of the eight estimators applies the official preprocessing and uses at most 500 input columns; our compact inputs are retained in full. Main point predictions run on an H200 in the released mixed-precision recipe; the uncertainty and descriptor-baseline experiments run on an H100 with the same weights.

\paragraph{Decomposition.}
In the final pipeline the magnitude context has 13,181 reactions and the direction context the 8,538 signed AHO reactions (12,165 and 7,522 during selection). Both heads reuse the frozen weights. The temperature-normalized control divides the magnitude by the parsed $RT$ scale before regression and multiplies it back at prediction.

\paragraph{Compute cost.}
Supplementary Table~\ref{tab:cost} lists the cost of each step.

\begin{table}[htbp]
\centering\small
\caption{Compute cost on the main benchmark (13,181 context reactions, 1,501 queries; one H200 GPU, 16 CPU threads; median of three repetitions, excluding imports and the first warm-up). Feature construction is shared by all models.}
\label{tab:cost}
\setlength{\tabcolsep}{3.7pt}
\begin{tabular}{lrrrr}
\toprule
Step & Fit (s) & Predict (s) & Peak GPU (GB) & Trained parameters\\
\midrule
Count fingerprints, 13,979 molecules (CPU) & 9.4 & -- & -- & 0\\
R-SPOC blocks, 10,008 substrate--product strings (CPU)$^a$ & 62.8 & -- & -- & 0\\
Role projections from scratch (CPU) & 122.2 & -- & -- & 0\\
\midrule
\rxnpfn{}, two heads (GPU) & 10.9 & 1.40 & 12.5 & 0\\
\rxndirect{} (GPU) & 7.2 & 0.76 & 12.8 & 0\\
TabPFN~3.5 (GPU) & 2.7 & 9.03 & 5.5 & 0\\
ExtraTrees, 800 trees (CPU) & 52.2 & 0.12 & -- & all\\
XGBoost, 600 trees (CPU) & 5.8 & 0.004 & -- & all\\
DRFP + XGBoost, fit and predict (CPU) & 26.8 & -- & -- & all\\
rxnfp-BERT, fine-tuned, one seed (GPU) & 152.6 & -- & -- & all\\
Chemprop CGR, 80 epochs, one seed (GPU) & $\approx$2,660 & -- & -- & all\\
\bottomrule
\end{tabular}

\smallskip
\footnotesize $^a$ Extrapolated from 1,000 timed strings. For the frozen models ``fit'' is context preprocessing and encoding; no gradient step is taken. \rxndirect{} answers a single query in 0.29 s and a batch of 100 in 0.26 s once the context is encoded.
\end{table}

\section{Quantitative Details of the Main Results}
\label{app:details}
\input{tex/results_details}

\input{tex/cases_more_figure.tex}

\section{Complete Baselines}
\label{app:baselines}

\begin{figure}[htbp]
\centering
\includegraphics[width=0.70\linewidth]{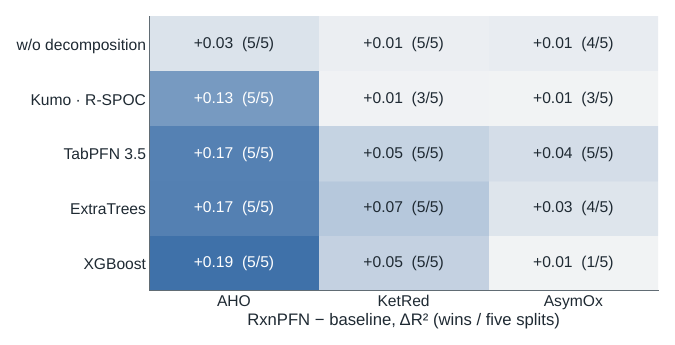}
\caption{\textbf{Robustness over repeated grouped splits.} Mean test $\Delta R^2$ of RxnPFN minus each baseline over five reaction-pair-grouped splits of each collection. Parentheses give the number of splits on which RxnPFN has higher $R^2$. All fitting, representation selection and evaluation follow the per-dataset protocol.}
\label{fig:resplits}
\end{figure}

\begin{figure}[htbp]
\centering
\figinclude{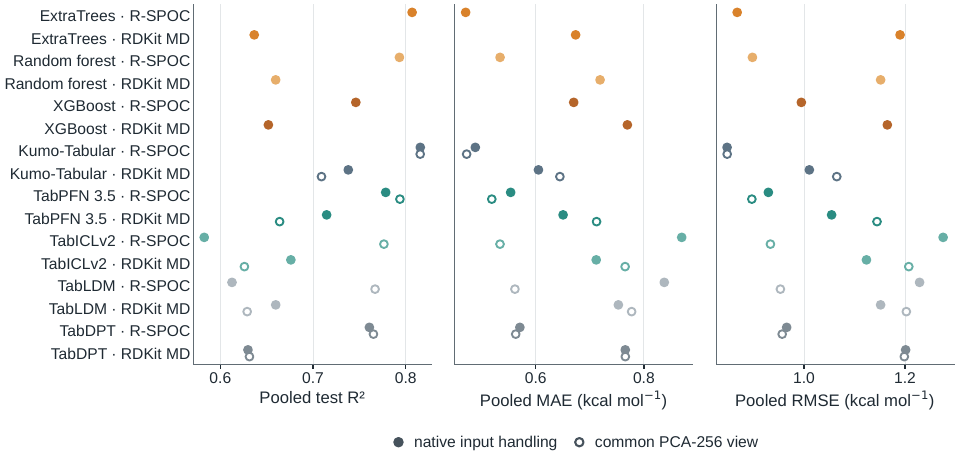}{figures/fig_baselines.pdf}
\caption{\textbf{Baselines on identical test reactions, pooled setting.} Pooled test $R^2$, MAE and RMSE over all 1,501 test reactions for tree models and five frozen pretrained tabular models with R-SPOC or its numerical RDKit-descriptor block (MD), with each model's native input handling (filled) or a shared training-fitted 256-coordinate PCA of the chemistry block plus four metadata columns (open). All models were fitted once on the three sources together, so no per-source values are shown.}
\label{fig:baselines}
\end{figure}

\begin{figure}[htbp]
\centering
\figinclude{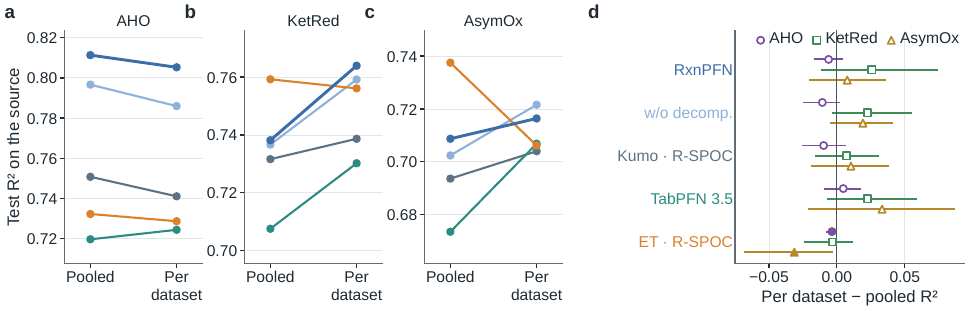}{figures/figS_settings.png}
\caption{\textbf{Comparison of training settings.} \textbf{a}--\textbf{c}, Test $R^2$ on each source for the same model and input trained, or given a context, on that source alone (per dataset) or on all three sources together (pooled; the pooled model's predictions scored on the source's test reactions). \textbf{d}, Per-dataset minus pooled $R^2$ with paired 95\% cluster-bootstrap intervals (2,000 draws); filled markers exclude zero. Line colours follow the model labels of panel d.}
\label{fig:settings}
\end{figure}

Tree hyperparameters are fixed before scoring: ExtraTrees uses 800 trees, all features per split, leaf size one and no bootstrap; RandomForest uses 400 bootstrap trees and one third of the features per split; XGBoost uses 600 depth-six histogram trees with learning rate 0.05 and unit $L_2$ regularization. A fixed seed is used throughout. The MD descriptor keeps only the 651 numerical RDKit properties of R-SPOC under the same aggregation, plus temperature and three source indicators (532 non-constant columns on final training). Supplementary Table~\ref{tab:baselines-all} lists pooled test $R^2$ for all inputs and views. Native results reflect each model's own feature handling---Kumo, for instance, caps each estimator at 500 columns and TabDPT applies its own projection---so the shared PCA view is the fairer comparison across backbones. Supplementary Fig.~\ref{fig:baselines} shows the same comparison across the three metrics.

\begin{table}[htbp]
\centering\small
\caption{Pooled test $R^2$ of the evaluated baseline configurations. n.e., not evaluated: tree baselines were run on native descriptors; PCA256 was the shared-input control for the pretrained tabular models.}
\label{tab:baselines-all}
\begin{tabular}{lrrrr}
\toprule
Model & R-SPOC native & MD native & R-SPOC PCA256 & MD PCA256\\
\midrule
ExtraTrees & 0.807 & 0.637 & n.e. & n.e.\\
RandomForest & 0.793 & 0.660 & n.e. & n.e.\\
XGBoost & 0.746 & 0.652 & n.e. & n.e.\\
Kumo & 0.816 & 0.738 & 0.816 & 0.709\\
TabPFN 3.5 & 0.778 & 0.715 & 0.794 & 0.664\\
TabICLv2 & 0.583 & 0.676 & 0.776 & 0.626\\
Xiaomi-TabLDM & 0.613 & 0.660 & 0.767 & 0.629\\
TabDPT 1.3 & 0.761 & 0.630 & 0.765 & 0.631\\
\bottomrule
\end{tabular}
\end{table}

In an earlier comparison with the same splits and R-SPOC-based inputs, LimiX-2, TabFM, TabAbax and a PriorTune/LightGBM pipeline reached best test $R^2$ between $0.801$ and $0.806$, below Kumo. Supplementary Table~\ref{tab:bootstrap} gives the paired source-stratified intervals for the direct comparisons.

\begin{table}[htbp]
\centering\small
\caption{Paired source-stratified bootstrap over substrate--product groups (1,000 resamples; 4,000 for the last two rows). $\Delta$ = first minus second model.}
\label{tab:bootstrap}
\begin{tabular}{lrr}
\toprule
Comparison & $\Delta R^2$ & 95\% interval\\
\midrule
\rxndirect{} (RTF) vs ExtraTrees (R-SPOC) & 0.038 & [0.018, 0.064]\\
\rxndirect{} (RTF) vs Kumo (R-SPOC) & 0.030 & [0.017, 0.044]\\
\rxndirect{} (RTF) vs ExtraTrees (RTF) & 0.031 & [0.020, 0.043]\\
\rxndirect{} (FDE) vs ExtraTrees (FDE) & 0.024 & [0.011, 0.037]\\
\rxndirect{} vs ExtraTrees, product constitution only & 0.096 & [0.073, 0.120]\\
\rxndirect{} vs ExtraTrees, no product & 0.097 & [0.071, 0.123]\\
\rxnpfn{} (selected) vs \rxndirect{} (RTF) & 0.010 & [0.001, 0.019]\\
\rxnpfn{} (selected) vs \rxnpfn{} (RTF magnitude) & $-0.002$ & [$-0.007$, 0.003]\\
\bottomrule
\end{tabular}
\end{table}

\paragraph{Fine-tuning the backbone.}\label{app:lora} To test whether fine-tuning a pretrained tabular model adds to in-context prediction, we trained LoRA adapters for TabPFN 3.5 on the 12,165 fitting reactions for 14 epochs (672 updates) and selected epoch 2 on validation. With matched context and two estimators, test $R^2$ changes from $0.808$ (frozen weights) to $0.808$ (LoRA; $\Delta R^2=+0.0001$) and MAE from $0.472$ to $0.472$. Fine-tuning therefore gives no measurable gain here, which supports investing in the input representation and response construction instead.

\section{Information Controls}
\label{app:controls}
Supplementary Table~\ref{tab:controls} gives the per-dataset results of the product-information controls shown in Fig.~\ref{fig:ablation}c. The product-constitution input removes all product atom and bond stereochemistry; the no-product input zeros the product fingerprints and validity flag; all projections are refitted on the unchanged training reactions. The permutation control shuffles training labels within each source once. In the pooled setting, \rxndirect{} drops from $0.845$ to $0.793$ (constitution) and $0.786$ (no product) and still leads ExtraTrees on the same coordinates by $+0.096$ $[+0.073,+0.120]$ and $+0.097$ $[+0.071,+0.123]$, a larger margin than with the product; the full RxnPFN pipeline gives $0.794$ and $0.791$, so without product stereochemistry the decomposition adds little over \rxndirect{} ($+0.002$ and $+0.005$), because the direction head loses most of its information.

\begin{table}[htbp]
\centering\small
\caption{Product-information controls on the RTF coordinates. \emph{Pooled}: one model for all sources, $R^2$ over all 1,501 test reactions. \emph{Per dataset}: one context per source with projections refitted on it, test $R^2$ of that source. For KetRed and AsymOx \rxnpfn{} reduces to its FDE magnitude head. Pooled-model controls without source indicators give $R^2=0.842$, and within-source label permutation gives $0.269$.}
\label{tab:controls}
\begin{tabular}{llrrrr}
\toprule
& & Pooled & \multicolumn{3}{c}{Per dataset}\\
\cmidrule(lr){3-3}\cmidrule(lr){4-6}
Model & Input & $R^2$ & AHO & KetRed & AsymOx\\
\midrule
\rxnpfn & full record & 0.855 & 0.805 & 0.764 & 0.716\\
\rxnpfn & product constitution only & 0.794 & 0.716 & 0.754 & 0.697\\
\rxnpfn & no product & 0.791 & 0.729 & 0.725 & 0.686\\
\rxndirect & full record & 0.845 & 0.786 & 0.759 & 0.722\\
\rxndirect & product constitution only & 0.793 & 0.716 & 0.738 & 0.699\\
\rxndirect & no product & 0.786 & 0.712 & 0.746 & 0.712\\
ExtraTrees & full record & 0.814 & 0.751 & 0.701 & 0.698\\
ExtraTrees & product constitution only & 0.696 & 0.570 & 0.678 & 0.703\\
ExtraTrees & no product & 0.688 & 0.556 & 0.702 & 0.694\\
\bottomrule
\end{tabular}
\end{table}

\section{Additional Benchmark Analyses}
\label{app:benchmark}
\input{tex/figs/figS_extrapolation.tex}

Component ablation and benchmark retrieval use single runs. At 13,181 and 12,165 labels, paired three-seed pooled $R^2$ averaged $0.855$ and $0.839$ for RxnPFN and $0.842$ and $0.825$ without decomposition.

\paragraph{Seed variability.} Across five model seeds, which change the estimator ensemble and the order of features, pooled RxnPFN reaches $0.856\pm0.002$ (range $0.854$--$0.858$) and \rxndirect{} $0.841\pm0.003$, a mean gain of $0.015$ from the decomposition. ExtraTrees varies by less than $0.001$. In the single-run comparison, the selected pipeline improves on \rxndirect{} with the RTF input by $\Delta R^2=+0.010$ (95\% interval $[+0.001,+0.019]$); the RTF-magnitude variant has a slightly higher test score ($0.856$), but the difference between the two is within noise ($-0.002$, interval $[-0.007,+0.003]$), and we keep the variant chosen on validation.

\paragraph{Rank control and the rejected transductive variant.} Beyond the rank control adopted as default (Methods), we evaluated a transductive variant that fits the role projections on the structures of context and query reactions together, using no query labels. It lowered the per-dataset validation $R^2$ of RxnPFN on KetRed ($-0.006$) and AsymOx ($-0.011$) and did not improve the Zahrt out-of-catalyst split (MAE $0.372$ with rank control alone, $0.378$ with the variant), so it was not adopted under the pre-registered rule; it does help trees on that benchmark.

\paragraph{Zahrt chiral phosphoric acid benchmark.} On the standard out-of-substrate, out-of-catalyst and combined splits of the thiol-addition benchmark \citep{Zahrt_2019} (384 context reactions; 216, 304 and 171 queries), the published MAEs are $0.161$, $0.211$ and $0.238$ kcal\,mol$^{-1}$ for ASO/AEIF descriptors with SVM \citep{Zahrt_2019}, $0.137$, $0.254$ and $0.262$ for MFF \citep{Sandfort_2020} and $0.135$, $0.215$ and $0.212$ for ECFP with XGBoost \citep{Asahara2022} (Supplementary Fig.~\ref{fig:extrapolation}). RxnPFN with rank-controlled RTF projections reaches $0.217$, $0.372$ and $0.576$, and \rxndirect{} $0.173$, $0.399$ and $0.549$: projections fitted on 24 context catalysts do not represent new catalysts. With the single-role encoding the frozen model reaches $0.166$, $0.253$ and $0.354$, and with the R-SPOC descriptor $0.141$, $0.252$ and $0.308$ (ExtraTrees on R-SPOC $0.153$, $0.262$ and $0.297$). On new catalysts the frozen model therefore reaches the level of MFF only with uncompressed or descriptor inputs, and on the combined split it remains behind the published methods. On random 600/475 splits all models reach MAE of about $0.14$ kcal\,mol$^{-1}$.

\paragraph{ChemAHNet reproduction on the Zahrt splits.} With the released thiol-addition script, which selects the checkpoint on the test set, the retrained ChemAHNet reproduces the published $R^2$ on the random 600/475 split ($0.909$ versus $0.918$), the imine ($0.908$ versus $0.926$), transformation ($0.895$ versus $0.918$), thiol ($0.587$ versus $0.657$) and catalyst splits ($0.778$ versus $0.774$). With checkpoints selected on a 10\% validation split of the training reactions instead, the random, imine and transformation splits are similar ($0.886\pm0.012$, $0.892\pm0.002$ and $0.877\pm0.034$), but the thiol and catalyst splits fall to $-0.920\pm0.190$ and $0.042\pm0.548$. On ChemAHNet's own hydrogenation benchmark, validation-based selection gives $0.492\pm0.032$ against the published $0.591$; the published test set and our test set are identical, but the published training run and its selection procedure are not available, so we report both numbers. The original study also reports $88.9\%$ accuracy for classifying the configuration of the major enantiomer, $R^2=0.797$ in ten-fold cross-validation on a single Rh/BINOL-phosphite subset, and results on three further small organocatalytic datasets; these are different tasks from the signed regression compared here.

\paragraph{Pooled versus per-dataset context.} Pooling the three datasets into one context helps the frozen model where the labels are signed and hurts it slightly on the magnitude-only datasets: RxnPFN reaches $0.811$ on AHO when pooled versus $0.805$ per dataset, but $0.738$ versus $0.764$ on KetRed and $0.709$ versus $0.716$ on AsymOx (Supplementary Fig.~\ref{fig:settings}). Trees show the opposite pattern on AsymOx, where pooled ExtraTrees reaches $0.737$ against $0.706$ when trained on AsymOx alone, presumably because the explicit source indicator lets a tree separate the datasets while it still borrows split structure from the larger collections. In practice a frozen in-context model can use either setting at no extra cost; for a small magnitude-only campaign the campaign's own reactions are the better context. The decomposition acts only on signed labels, so its gain is confined to AHO in both settings ($+0.019$ $[+0.004,+0.036]$ per dataset).

\paragraph{Direction and strength.} Pooled $R^2$ on this mixed benchmark rewards correct enantiomer direction more than accurate selectivity strength. AHO labels are signed and roughly half are negative, while the other datasets are magnitudes; a reference predictor that knows only the true sign and the training mean $|\Delta\Delta G|$ of each dataset reaches pooled test $R^2=0.763$, compared with $0.268$ for the per-dataset mean alone. Supplementary Table~\ref{tab:components} reports the two components separately. Pooled over all datasets, $R^2$ of $|\Delta\Delta G|$ is $0.588$ against $0.855$ for the signed values, with a margin over ExtraTrees on strength ($0.588$ versus $0.487$) and \rxndirect{} slightly better on this component ($0.610$). The constitution rows correspond to the setting before an experiment, when the major enantiomer is not known.

\begin{table}[htbp]
\centering\footnotesize
\caption{Direction and strength components on the test set. \emph{Pooled}: one model for all datasets, metrics over all 1,501 test reactions. \emph{Per dataset}: a model given only AHO training reactions as context, scored on the 940 AHO test reactions; sign accuracy over the 904 with non-zero labels. ee MAE in percentage points after back-conversion at each reaction's temperature. ``Constitution'' removes product stereochemistry; ``none'' removes the product.}
\label{tab:components}
\setlength{\tabcolsep}{3.5pt}
\begin{tabular}{llrrrrrrr}
\toprule
& & \multicolumn{3}{c}{Pooled} & \multicolumn{4}{c}{Per dataset, AHO}\\
\cmidrule(lr){3-5}\cmidrule(lr){6-9}
Model & Product & $|\Delta\Delta G|$ $R^2$ & Signed $R^2$ & ee MAE & Sign acc. & $|\Delta\Delta G|$ $R^2$ & Signed $R^2$ & ee MAE\\
\midrule
RxnPFN & full & 0.588 & 0.855 & 12.1 & 0.969 & 0.479 & 0.805 & 15.2\\
\rxndirect{} & full & 0.610 & 0.845 & 12.8 & 0.963 & 0.495 & 0.786 & 16.3\\
ExtraTrees (RTF) & full & 0.487 & 0.814 & 14.6 & 0.965 & 0.306 & 0.751 & 18.6\\
Kumo (R-SPOC) & full & 0.564 & 0.816 & 14.7 & 0.956 & 0.425 & 0.741 & 18.9\\
\midrule
RxnPFN & constitution & 0.518 & 0.794 & 16.6 & 0.897 & 0.324 & 0.716 & 22.3\\
\rxndirect{} & constitution & 0.556 & 0.793 & 17.0 & 0.894 & 0.410 & 0.716 & 22.3\\
ExtraTrees (RTF) & constitution & 0.351 & 0.696 & 21.9 & 0.862 & 0.100 & 0.570 & 30.5\\
RxnPFN & none & 0.508 & 0.791 & 16.6 & 0.903 & 0.334 & 0.729 & 21.8\\
\rxndirect{} & none & 0.550 & 0.786 & 17.2 & 0.889 & 0.399 & 0.712 & 22.8\\
ExtraTrees (RTF) & none & 0.290 & 0.688 & 22.4 & 0.862 & 0.009 & 0.556 & 31.3\\
\bottomrule
\end{tabular}
\end{table}

\paragraph{Component ablation.} Supplementary Table~\ref{tab:ladder} lists validation and test scores along the design path, from ExtraTrees on R-SPOC to RxnPFN, including the hard readout and a temperature-normalized magnitude head; Supplementary Table~\ref{tab:blocks} lists the drop-one-block ablation plotted in Fig.~\ref{fig:ablation}a. The magnitude/direction pipeline shows the same block ordering as \rxndirect{}.

\begin{table}[htbp]
\centering\small
\caption{Component ablation on the same splits. Validation $R^2$ ($n=1{,}560$) is the selection criterion; test $R^2$ ($n=1{,}501$) is reported once per configuration.}
\label{tab:ladder}
\begin{tabular}{lrrr}
\toprule
Configuration & Validation $R^2$ & Test $R^2$ & Test MAE\\
\midrule
ExtraTrees, R-SPOC & -- & 0.807 & 0.471\\
Kumo, R-SPOC & -- & 0.816 & 0.488\\
\rxndirect, FDE (435) & 0.746 & 0.831 & 0.448\\
\rxndirect, RTF (475) & 0.744 & 0.845 & 0.442\\
\quad + magnitude--direction decomposition, hard readout & 0.770 & -- & --\\
\quad + magnitude--direction decomposition, soft readout & 0.792 & 0.856 & 0.431\\
\quad + FDE magnitude head (\rxnpfn, selected) & 0.795 & 0.855 & 0.431\\
\quad + temperature-normalized magnitude head & 0.793 & -- & --\\
\bottomrule
\end{tabular}
\end{table}

\begin{table}[htbp]
\centering\footnotesize
\caption{Drop-one-block ablation of the 475-dimensional RTF input (pooled $R^2$; validation context 12,165, test context 13,181). ``Mag./dir.'' uses the RTF input for both heads; ``$-$ metadata'' removes the 18 non-temperature metadata columns.}
\label{tab:blocks}
\begin{tabular}{lrrrrrr}
\toprule
& & \multicolumn{2}{c}{\rxndirect{}} & \multicolumn{2}{c}{Mag./dir.} & ExtraTrees\\
\cmidrule(lr){3-4}\cmidrule(lr){5-6}\cmidrule(lr){7-7}
Input & Width & Val. & Test & Val. & Test & Test\\
\midrule
Full RTF & 475 & 0.744 & 0.845 & 0.792 & 0.856 & 0.814\\
$-$ Morgan roles & 283 & 0.749 & 0.848 & 0.794 & 0.856 & 0.812\\
$-$ gained/lost & 379 & 0.717 & 0.843 & 0.787 & 0.856 & 0.809\\
$-$ AtomPair & 307 & 0.696 & 0.830 & 0.743 & 0.841 & 0.802\\
$-$ metadata & 457 & 0.740 & 0.842 & 0.790 & 0.853 & 0.810\\
\bottomrule
\end{tabular}
\end{table}

\paragraph{High-selectivity labels.} Labels at the top of the ee scale are poorly resolved: at 298~K, ee values of 99\% and 99.9\% correspond to 3.1 and 4.5 kcal\,mol$^{-1}$, so between them the label moves by about 1.4 kcal\,mol$^{-1}$ although the two values are within typical measurement error, and 19.7\% of KetRed reactions lie at or above the 99\% level, with a maximum of 9.03 kcal\,mol$^{-1}$ that reflects a reported ee of essentially 100\%. 295 reactions have $|\mathrm{ee}|\geq99.5\%$ (36 in test). Scoring the unchanged predictions without these 36 test reactions raises pooled RxnPFN from $0.855$ to $0.865$; per dataset, AHO rises from $0.805$ to $0.822$, whereas KetRed falls from $0.764$ to $0.696$, because its eleven clipped test reactions lie at the top of the label range, so removing them lowers the label variance more than the residual error. Changing the context labels lowers the score: capping all training labels at the 99\%-ee value of their temperature gives $0.838$ pooled, and removing the clipped reactions from the context gives $0.825$, both on the full test set. Per dataset the same changes give $0.802$ and $0.791$ on AHO but $0.561$ and $0.471$ on KetRed, where clipped labels are most frequent. The clipped labels are thus informative ordinal information that the model uses, and treating them as exact or dropping them both have a cost; a censored likelihood is the principled alternative but is not available in a frozen in-context regressor. Separately, 235 of the 402 KetRed test reactions report no temperature and were converted at the fixed 25~$^\circ$C used wherever a temperature is unavailable; RxnPFN predicts them less accurately than the 167 with a recorded temperature (MAE $0.376$ against $0.269$~kcal\,mol$^{-1}$; $R^2$ $0.743$ against $0.814$), so part of the KetRed error lies in the target rather than in the prediction.

\section{Representation, Context and Embedding Exploration}
\label{app:exploration}
Curating the context helps only where the library is much larger than the number of precedents a prediction needs. Deduplicating the training inputs improves validation slightly ($0.749$ versus $0.744$) but gives test $R^2=0.844$ against $0.845$ with the full context, so the duplicate rows were not what limited it. Per-query retrieval is the effective form of curation (Supplementary Table~\ref{tab:retrieval}): on four of the six collections the retrieved context exceeded the whole table, and on the pooled 13,181-reaction library 1,000 retrieved reactions, 7.6\% of it, gave $0.848$ against $0.855$ for the whole library, where a random 1,000 gave $0.660$. On the two collections where retrieval did not help, ketone reduction and the OrgAIcat aldol reactions, retrieved rows were still far better than random ones of the same size, but the whole table remained better still, so the discarded rows carried information that their nearest neighbours did not replace.
Nine additional chemical inputs---rooted stereochemistry, reactive-site environments, topological partner properties, role-wise Tanimoto maps, catalyst assembly, explicit-unit conditions, chemistry classes and a label-supervised projection---and fifteen embedding variants, including a label-free autoencoder, did not exceed the RTF input on validation. The frozen model therefore benefits from an informative and compact representation of the whole record more than from additional hand-crafted views.

\begin{table}[htbp]
\centering\footnotesize
\caption{\textbf{Retrieval against a size-matched random context.} Each query is scored with the $K$ most similar measured reactions of its context, ranked by cosine similarity in the RTF input, and with $K$ reactions drawn at random from the same context. Both arms score the same queries at the same context size, so the difference isolates which rows were supplied rather than how many. ``Full'' is the whole context. ``Retained'' is $K$ as a fraction of that context. The first three collections and the pooled rows use the primary magnitude--direction readout of Table~\ref{tab:main}, scored on their whole test set with each collection's own training rows as the library; the pooled rows draw one library of all 13,181 training reactions, so a query may retrieve any class. The OrgAIcat and Buchwald--Hartwig rows are scored on a fixed query sample (80 per fold over five random folds, and 60 per split over three splits) with that fold's training rows as the library; their targets are magnitudes, for which the primary readout reduces to the magnitude head, which is what was used. $R^2$ is computed on the queries scored and against the target of that collection.}
\label{tab:retrieval}
\begin{tabular}{lrrrrrr}
\toprule
Collection & Queries & Full & $K$ & Retained & Retrieved & Random\\
\midrule
AHO (8,538) & 940 & 0.805 & 1,000 & 11.7\% & 0.807 & 0.566\\
 & & & 2,000 & 23.4\% & \textbf{0.817} & 0.684\\
 & & & 4,000 & 46.8\% & 0.809 & 0.744\\
\midrule
KetRed (3,430) & 402 & 0.764 & 1,000 & 29.2\% & 0.753 & 0.708\\
 & & & 2,000 & 58.3\% & 0.760 & 0.671\\
\midrule
AsymOx (1,213) & 159 & 0.716 & 512 & 42.2\% & 0.713 & 0.603\\
 & & & 1,000 & 82.4\% & \textbf{0.719} & 0.698\\
\midrule
OrgAIcat Michael (7,412) & 400 & 0.816 & 1,000 & 13.5\% & 0.817 & 0.643\\
 & & & 2,000 & 27.0\% & \textbf{0.818} & 0.704\\
 & & & 4,000 & 54.0\% & 0.815 & 0.763\\
\midrule
OrgAIcat aldol (10,803) & 400 & 0.754 & 1,000 & 9.3\% & 0.738 & 0.489\\
 & & & 2,000 & 18.5\% & 0.741 & 0.623\\
 & & & 4,000 & 37.0\% & 0.752 & 0.719\\
\midrule
Buchwald--Hartwig (2,767) & 180 & 0.983 & 512 & 18.5\% & 0.981 & 0.905\\
 & & & 1,000 & 36.1\% & 0.984 & 0.938\\
 & & & 2,000 & 72.3\% & \textbf{0.984} & 0.955\\
\midrule
Pooled (13,181) & 1,501 & 0.855 & 1,000 & 7.6\% & 0.848 & 0.660\\
 & & & 2,000 & 15.2\% & 0.854 & 0.722\\
 & & & 4,000 & 30.3\% & \textbf{0.855} & 0.781\\
\bottomrule
\end{tabular}
\end{table}

\begin{table}[htbp]
\centering\small
\caption{\textbf{Local regression on RTF and FDE inputs.} We evaluated cosine $k$-nearest neighbours with uniform weights or cosine similarities floored at $10^{-6}$ ($k\in\{10,50,200,10^3\}$), linear ridge and ridge on radial basis features from up to 2,000 context anchors. Ridge penalties and radial basis scales were screened on the pooled validation reactions using projections fitted on the final context. The table reports the highest observed test $R^2$ among the evaluated configurations for each query set, and the corresponding MAE in kcal\,mol$^{-1}$.}
\label{tab:local}
\begin{tabular}{llrrrr}
\toprule
Query set & Best local learner & $R^2$ & MAE & RxnPFN $R^2$ & RxnPFN MAE\\
\midrule
AHO & $k$-NN, $k=10$, cosine-weighted, standardized RTF & 0.574 & 0.890 & 0.805 & 0.504\\
KetRed & $k$-NN, $k=10$, cosine-weighted, raw RTF & 0.612 & 0.419 & 0.767 & 0.331\\
AsymOx & Ridge on radial basis features, 1,213 anchors & 0.675 & 0.315 & 0.712 & 0.264\\
Pooled & $k$-NN, $k=10$, cosine-weighted, standardized RTF & 0.687 & 0.727 & 0.856 & 0.430\\
\bottomrule
\end{tabular}
\end{table}

Supplementary Table~\ref{tab:features} lists the additional chemical inputs evaluated on the full validation set with the same backbone; only explicit-unit condition repair was also tested. Supplementary Table~\ref{tab:context} gives the context policies. The embedding variants of Supplementary Table~\ref{tab:embedding-all} test catalyst- and reaction-weighted dimension budgets, 8,192-bin Morgan counts, training-only TF--IDF, a substrate--catalyst kernel, supervised and metric-learned coordinates, a random projection and a label-free autoencoder; supervised projections overfit even with context-only labels and inner early stopping. Approximate Gasteiger charges and graph-shell volumes are topological proxies, not conformer-derived steric parameters.

\begin{table}[htbp]
\centering\small
\caption{Additional chemical inputs (validation $R^2$, $n=1{,}560$).}
\label{tab:features}
\begin{tabular}{lrr}
\toprule
Input & Validation $R^2$ & Test $R^2$\\
\midrule
FDE (reference) & 0.746 & 0.831\\
RTF (reference) & 0.744 & 0.845\\
Explicit-unit condition repair & 0.744 & 0.840\\
Rooted stereochemistry & 0.743 & --\\
Reactive/donor site environments & 0.743 & --\\
Chemistry-class annotations & 0.741 & --\\
Site/stereo Tanimoto & 0.740 & --\\
Topological partner properties & 0.735 & --\\
Catalyst-component assembly & 0.733 & --\\
Training-label PLS projection & 0.709 & --\\
Role-wise Tanimoto replacement & 0.704 & --\\
\bottomrule
\end{tabular}
\end{table}

\begin{table}[htbp]
\centering\small
\caption{Context policies with the RTF input.}
\label{tab:context}
\begin{tabular}{lrrr}
\toprule
Context & Validation $R^2$ & Test $R^2$ & Test context size\\
\midrule
Full training set & 0.744 & 0.845 & 13,181\\
Deduplicated training inputs & 0.749 & 0.844 & 11,795\\
512 most similar reactions per query & 0.742 & 0.835 & 512\\
\bottomrule
\end{tabular}
\end{table}

\begin{longtable}{llllrr}
\caption{Embedding variants. ``Supplied'' uses the recorded product stereochemistry; ``constitution'' removes product stereo. $R^2$ on all validation or test reactions.}\label{tab:embedding-all}\\
\toprule Set & Product & Stage & Variant & Width & $R^2$\\\midrule\endfirsthead
\toprule Set & Product & Stage & Variant & Width & $R^2$\\\midrule\endhead
\bottomrule\endfoot
1 & Supplied & validation & catalyst\_budget & 463 & 0.736\\
1 & Supplied & validation & hires\_balanced & 455 & 0.731\\
1 & Supplied & validation & hires\_catalyst & 463 & 0.728\\
1 & Supplied & validation & hires\_hybrid & 499 & 0.737\\
1 & Supplied & validation & hybrid\_budget & 499 & 0.736\\
1 & Supplied & validation & joint\_chemical\_kernel & 483 & 0.709\\
1 & Supplied & validation & metric128 & 499 & 0.562\\
1 & Supplied & validation & metric64 & 499 & 0.546\\
1 & Supplied & validation & reaction\_budget & 495 & 0.727\\
1 & Supplied & validation & supervised32 & 499 & 0.488\\
1 & Supplied & validation & tfidf\_hires & 479 & 0.740\\
1 & Supplied & test & hybrid\_budget & 499 & 0.840\\
1 & Supplied & test & joint\_chemical\_kernel & 483 & 0.833\\
1 & Supplied & test & metric128 & 499 & 0.739\\
1 & Supplied & test & tfidf\_hires & 479 & 0.839\\
1 & Constitution & test & tfidf\_hires & 479 & 0.780\\
2 & Supplied & validation & autoencoder16 & 491 & 0.742\\
2 & Supplied & validation & early\_metric16 & 491 & 0.522\\
2 & Supplied & validation & early\_supervised16 & 491 & 0.568\\
2 & Supplied & validation & random16 & 491 & 0.741\\
2 & Supplied & test & autoencoder16 & 491 & 0.844\\
2 & Supplied & test & early\_supervised16 & 491 & 0.774\\
2 & Constitution & test & autoencoder16 & 491 & 0.787\\
\end{longtable}

\section{Uncertainty Details}
\label{app:uncertainty}
\input{tex/uq_details}

\input{tex/substrate_coverage}

\paragraph{Pooled calibration.} In the pooled setting native 90\% intervals of \rxndirect{} have mean width $2.012$ kcal\,mol$^{-1}$. Pooled and source-wise conformal corrections raise 90\% coverage to 91.7\% and 92.1\% without improving the interval score ($2.747$ native, $2.752$ calibrated). For the clipped two-head readout (before the sign-mixture correction of Eq.~\eqref{eq:mixture}), the interval score rises from $2.747$ to $4.526$ and width correlates less with error ($\rho=0.317$); conformal correction restores pooled coverage to 93.6\% with intervals 1.5 times wider than those of \rxndirect{}. Accepting the 65.0\% of test reactions whose width is below the calibration median lowers pooled MAE of \rxndirect{} from $0.525$ to $0.297$ kcal\,mol$^{-1}$. Coverage on the present split is an empirical observation; reaction grouping and dataset heterogeneity mean that conformal exchangeability assumptions are not guaranteed.

For $N$ test reactions, empirical coverage and mean width are $\widehat{\mathrm{cov}}=N^{-1}\sum_i\mathbf 1\{y_i\in[\ell_i,u_i]\}$ and $\overline w=N^{-1}\sum_i(u_i-\ell_i)$. The interval score at miscoverage $\alpha$ is
\begin{equation}
 \mathrm{IS}_{\alpha}=\frac1N\sum_i\Big[(u_i-\ell_i)+\tfrac{2}{\alpha}(\ell_i-y_i)_+ +\tfrac{2}{\alpha}(y_i-u_i)_+\Big].
\end{equation}
The 999-quantile CRPS approximation (twice the mean pinball loss) is $0.359$ for Kumo versus $1.117$ for the empirical target distribution of the same context, and the weighted interval score over $\alpha\in\{0.2,0.1,0.05\}$ is $0.199$.

\section{Yield and Conversion}
\label{app:yield}
We apply the same frozen model to two single-endpoint tasks without pooling yield and conversion (Supplementary Table~\ref{tab:yield-all}). AsymOx contributes 1,516 reported yields and KetRed 4,232 conversions under their original splits (1,213/144/159 and 3,430/400/402). The AsymOx conversion column duplicates yield and is not used. Inputs are the RTF, R-SPOC reduced to 256 PCA coordinates, or MD reduced to 256 coordinates, each fitted on the task's training reactions; predictions and intervals are clipped to $[0,100]$.

Kumo is better than ExtraTrees on AsymOx yield ($R^2=0.645$ versus $0.611$ with RTF input; $0.662$ with MD) and on KetRed conversion ($0.434$ versus $0.371$; $0.450$ with R-SPOC). Native 90\% Kumo coverage is 90.57\% for AsymOx yield and 93.28\% for KetRed conversion.

\begin{table}[htbp]
\centering\small
\caption{All yield and conversion configurations. RTF, role--transformation fingerprint; R, R-SPOC; MD, numerical RDKit descriptors. MAE and RMSE in percentage points.}
\label{tab:yield-all}
\begin{tabular}{lllrrr}
\toprule
Task & Input & Model & $R^2$ & MAE & RMSE\\
\midrule
AsymOx reported yield & R PCA & ExtraTrees & 0.518 & 11.80 & 16.48\\
AsymOx reported yield & R PCA & Kumo & 0.606 & 11.21 & 14.90\\
AsymOx reported yield & MD PCA & ExtraTrees & 0.619 & 10.85 & 14.66\\
AsymOx reported yield & MD PCA & Kumo & 0.662 & 10.38 & 13.80\\
AsymOx reported yield & RTF & ExtraTrees & 0.611 & 10.74 & 14.80\\
AsymOx reported yield & RTF & Kumo & 0.645 & 10.38 & 14.14\\
KetRed conversion & R PCA & ExtraTrees & 0.324 & 10.06 & 17.38\\
KetRed conversion & R PCA & Kumo & 0.450 & 8.57 & 15.68\\
KetRed conversion & MD PCA & ExtraTrees & 0.272 & 11.22 & 18.04\\
KetRed conversion & MD PCA & Kumo & 0.382 & 9.24 & 16.62\\
KetRed conversion & RTF & ExtraTrees & 0.371 & 8.96 & 16.77\\
KetRed conversion & RTF & Kumo & 0.434 & 8.18 & 15.90\\
\bottomrule
\end{tabular}
\end{table}

ReactionT5v2-yield, CheMeleon, ChemAHNet, CARL \citep{Lu2025CARL} and RXNGraphormer \citep{Xu2025RXNGraphormer} are evaluated on the same KetRed conversion and AsymOx yield test reactions. Supplementary Table~\ref{tab:yield_modern} reports mean metrics over seeds; Fig.~\ref{fig:yield_main}b instead scores the averaged predictions and shows reaction-pair bootstrap intervals. Every comparison aligns predictions by reaction identity and verifies the targets. The RXNGraphormer test loader preserves the prepared source-row order; it is not aligned by sorting or matching outcome values.

\input{tex/recent_yield_baselines.tex}

\section{Transfer to Public Aldol, Michael Addition and C--H Activation Collections}
\label{app:external-pilot}
We evaluated the frozen model on the two OrgAIcat collections (13,360 aldol reactions and 9,194 Michael additions) \citep{OrgAIcat2026} and on 1,651 Michael additions catalysed by $N,N'$-dioxide/metal complexes \citep{TangMichael2026}. We encoded both reactants in the substrate role and fitted the 192-coordinate R-SPOC projection and the Morgan, atom-pair and torsion role blocks on each fold's context. We converted ee to $|\Delta\Delta G^\ddagger|$ at the recorded temperature (25~$^\circ$C where missing), capped the target at 4.5005~kcal\,mol$^{-1}$ and predicted its magnitude with four frozen Kumo ensemble members.

In random five-fold cross-validation we assigned reaction records to folds. For grouped five-fold cross-validation we assigned all records of a given organocatalyst constitution to one fold. In the $N,N'$-dioxide collection we grouped by ligand constitution across metals. Stereoisomers shared a group in both cases. OrgAIcat used ten folds for its random benchmark and clipped absolute ee at 99.9\% before conversion to free energy. Its authors also reported a separate catalyst-skeleton analysis \citep{OrgAIcat2026}.

\begin{table}[htbp]
\centering\small
\caption{\textbf{Selectivity prediction on three public collections.} Scores are means over five test folds. The standard deviation (s.d.) is computed across the five $R^2$ values. Both MAE columns are in kcal\,mol$^{-1}$. The baseline predicts the mean target value of each fold's context.}
\label{tab:pilot-transfer}
\begin{tabular}{llrrrr}
\toprule
Collection & Split & $R^2$ & s.d. & MAE & Baseline MAE\\
\midrule
Aldol & Random & 0.737 & 0.019 & 0.280 & 0.637\\
 & Grouped & 0.164 & 0.074 & 0.555 & 0.639\\
Michael addition & Random & 0.802 & 0.007 & 0.249 & 0.682\\
 & Grouped & 0.299 & 0.181 & 0.514 & 0.678\\
$N,N'$-dioxide/metal Michael & Random & 0.780 & 0.029 & 0.297 & 0.724\\
 & Grouped & 0.021 & 0.162 & 0.721 & 0.735\\
\bottomrule
\end{tabular}
\end{table}

With held-out organocatalysts, RxnPFN's MAE remained below the context-mean baseline for both aldol and Michael additions. With held-out $N,N'$-dioxide ligands, the two errors were similar (0.721 and 0.735~kcal\,mol$^{-1}$).

A fourth collection moves outside organocatalysis altogether. The $\beta$-C(sp$^3$)--H activation study of Hoque et al.\ released 220 reactions with structures and \%~ee in its \texttt{EnsembleAC} repository \citep{Hoque2025CSP3H}. Each record is a dot-separated reaction string; the two coupling partners occupy the substrate role, the palladium precatalyst and the chiral ligand the catalyst roles, and the remaining fragments are separated into counter-ions (additive) and solvent by their formal charge. No product and no temperature are recorded, so the product role is left empty and ee is converted at 25~$^\circ$C. The collection has 78 ligands, 63 of them appearing once, so a catalyst-held-out fold would leave almost no precedents; only the random split is reported.

A fifth collection is the high-throughput screen of rhodium-catalysed asymmetric hydrogenation behind the solvent, pressure and temperature effects of Kalikadien et al.\ \citep{Kalikadien2024}. Its released response file lists 1,248 reactions with substrate and ligand identifiers but no structures; the identifier-to-SMILES tables in the associated \texttt{obelix-ml-pipeline} repository cover every code used, so 1,224 reactions with a finite ee were retained. The solvent is recorded by name and the two that occur, 1,2-dichloroethane and methanol, were mapped explicitly. No product is recorded, so that role is left empty. Ligand stereochemistry includes ferrocenyl and biaryl-atropisomeric motifs for which CIP labelling fails on very large graphs; those four stereocentre descriptor columns are set to zero for the affected ligands, which continue to contribute through the Morgan, atom-pair and torsion blocks. With 190 ligands over 1,224 reactions the rows per ligand are too few for a held-out fold, so again only the random split is reported.

\begin{table}[htbp]
\centering\small
\caption{\textbf{Two further collections scored on the free-energy scale.} Mean test $R^2$ and MAE over five random folds, with the s.d.\ of $R^2$ across folds. All three arms share the split and the RTF coordinates, so the ExtraTrees column differs from RxnPFN only in being fitted. ``Precedents'' is the mean context size per fold.}
\label{tab:transfer-new}
\begin{tabular}{lrrrrrrr}
\toprule
Collection & Reactions & Precedents & \rxnpfn{} $R^2$ & s.d. & ExtraTrees $R^2$ & \rxnpfn{} MAE & Mean $R^2$\\
\midrule
$\beta$-C(sp$^3$)--H activation & 220 & 176 & 0.570 & 0.019 & 0.566 & 0.229 & $-0.013$\\
Rh hydrogenation screen & 1,224 & 979 & \textbf{0.811} & 0.043 & 0.732 & 0.248 & $-0.003$\\
\bottomrule
\end{tabular}
\end{table}

Together these two collections locate the condition under which the frozen model earns its place. The encoding transfers to a bond-forming class it has never seen, but with about 176 precedents per fold it only reaches the accuracy of a fitted tree on the same input. With about 979 precedents it exceeds that tree by $0.079$, in every fold. The collections differ in both chemistry and size. The larger gain in the denser collection suggests that the supply of measured precedents helps determine when the frozen model is worth using.

The rhodium conjugate-addition and biaryl-arylation collections of Aguilar-Bejarano et al.\ were used to evaluate the HCat-GNet baseline compared in Table~\ref{tab:main} and were excluded from this transfer analysis.

\section{ScopeMap and SynAD Data}
\label{app:external}
\begin{figure}[htbp]
\centering
\figinclude{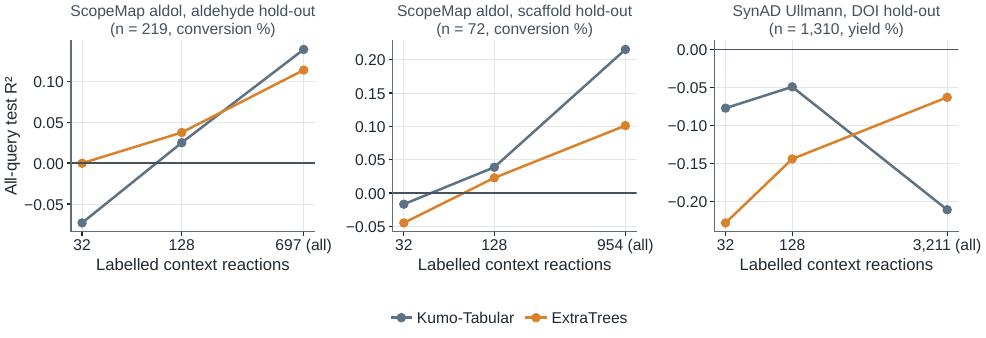}{figures/fig_external_risk.pdf}
\caption{\textbf{ScopeMap and SynAD.} All-query test $R^2$ of Kumo and ExtraTrees against the number of labelled context reactions for the ScopeMap aldol aldehyde and scaffold hold-outs (conversion) and the SynAD Ullmann DOI hold-out (yield). These are static evaluations, not the original active-learning protocols.}
\label{fig:external}
\end{figure}

We re-evaluate the released data of ScopeMap \citep{Li2026ScopeMap} and SynAD \citep{Tan2026SynAD} at the released repository commits as static regressions with each case's own training context; we do not reproduce their sampling or applicability-domain procedures. Supplementary Fig.~\ref{fig:external} gives the label-budget curves and Supplementary Table~\ref{tab:external-all} the per-configuration errors. ScopeMap's aldol table has 1,091 aldehyde/conversion entries; we use only the aldehyde structure (32 Morgan SVD coordinates) and split either by canonical aldehyde (697/175/219) or by Bemis--Murcko scaffold (954/65/72). SynAD's Ullmann ligand dataset has 5,495 yields from 107 publications, split by DOI into 3,211/974/1,310 reactions; eight molecular roles and conditions give 214 features, with products and identifiers excluded. Context budgets of 32, 128 and all labels share a feature basis fitted on the unlabelled training inputs.

Kumo is better than ExtraTrees on both aldol splits but the absolute accuracy is modest ($R^2=0.139$ and $0.216$ with the full context). On the Ullmann DOI split both models have negative $R^2$ ($-0.211$ for Kumo, $-0.063$ for ExtraTrees), and smaller contexts are not uniformly worse ($-0.077$ with 32 labels). Width-based selection still helps: retaining the 25\% of Ullmann queries with the narrowest native intervals lowers MAE from 23.45 to 16.85 percentage points. The SynAD article reports $R^2=0.90$ at 12.3\% retained coverage \citep{Tan2026SynAD}, which is a different protocol from our full-test scores.

\begin{table}[htbp]
\centering\small
\caption{ScopeMap and SynAD label-budget configurations. MAE and RMSE in percentage points.}
\label{tab:external-all}
\begin{tabular}{llrrrr}
\toprule
Case & Model & Context labels & $R^2$ & MAE & RMSE\\
\midrule
Aldol, aldehyde groups & ExtraTrees & 32 & $-0.000$ & 9.73 & 16.85\\
Aldol, aldehyde groups & ExtraTrees & 128 & 0.038 & 10.59 & 16.53\\
Aldol, aldehyde groups & ExtraTrees & 697 & 0.114 & 10.18 & 15.86\\
Aldol, aldehyde groups & Kumo & 32 & $-0.073$ & 12.15 & 17.46\\
Aldol, aldehyde groups & Kumo & 128 & 0.025 & 11.26 & 16.64\\
Aldol, aldehyde groups & Kumo & 697 & 0.139 & 10.62 & 15.64\\
Aldol, scaffold groups & ExtraTrees & 32 & $-0.045$ & 15.30 & 23.26\\
Aldol, scaffold groups & ExtraTrees & 128 & 0.023 & 14.14 & 22.50\\
Aldol, scaffold groups & ExtraTrees & 954 & 0.101 & 14.13 & 21.58\\
Aldol, scaffold groups & Kumo & 32 & $-0.017$ & 15.19 & 22.95\\
Aldol, scaffold groups & Kumo & 128 & 0.039 & 14.92 & 22.32\\
Aldol, scaffold groups & Kumo & 954 & 0.216 & 13.99 & 20.16\\
Ullmann, DOI groups & ExtraTrees & 32 & $-0.229$ & 21.84 & 28.58\\
Ullmann, DOI groups & ExtraTrees & 128 & $-0.144$ & 22.69 & 27.58\\
Ullmann, DOI groups & ExtraTrees & 3,211 & $-0.063$ & 21.36 & 26.58\\
Ullmann, DOI groups & Kumo & 32 & $-0.077$ & 20.04 & 26.76\\
Ullmann, DOI groups & Kumo & 128 & $-0.049$ & 21.70 & 26.41\\
Ullmann, DOI groups & Kumo & 3,211 & $-0.211$ & 23.45 & 28.38\\
\bottomrule
\end{tabular}
\end{table}

\section{Campaign and Sparse-Data Study Details}
\label{app:campaigns}
\input{tex/figs/figS_smalldata_uq.tex}

\paragraph{Nature sparse-data benchmark, Case~2.} On Case~2 (nickel cross-coupling) of Gallarati et al.\ \citep{Gallarati_2026} the author-feature model matches MLR on the ten official out-of-sample ligand--reaction combinations (MAE $0.289$ versus $0.295$) and is weaker on the new reaction H and in leave-one-reaction-out evaluations. Only the ten out-of-sample combinations have a measured same-reaction training reference; predicting the deviation from it raises MAE to $0.318$ with the author-selected descriptors. A more general variant that pairs each query with four training references chosen from the descriptors leaves the C--H leave-one-reaction-out MAE unchanged ($0.261$) and increases it on the Ni leave-one-reaction-out ($0.276$ versus $0.265$), out-of-sample ligand ($0.326$ versus $0.289$) and new-reaction tasks ($0.219$ versus $0.202$). The fixed admission rule of Methods recovers L6 for Case~1 but admits a reference for almost no other query: 2 of 10 new Ni ligands, 9 of 10 Co targets, and none of the leave-one-reaction-out queries or the 21 Ni targets, where the substrate is held out by construction. Where it applies outside Case~1 it does not help (Co targets: $R^2=0.068$ with $\Delta y$ versus $0.119$ without; Ni ligands, two queries: MAE $0.373$ versus $0.182$). On Case~1, applying the reference-relative transformation to the authors' models helps them less than it helps the frozen model: SISSO with $\Delta y$ has MAE $0.347$, and the best author variant, MLR with differenced descriptors and $\Delta y$, has $0.120$. Supplementary Table~\ref{tab:smalldata} summarizes the sparse-data studies; further Nature results are in Supplementary Note~\ref{app:nature}.

\paragraph{Pd-to-Ni transfer and Co chiral acids.} With our role fingerprints and only the 20 Ni support reactions, \rxndirect{} reaches $R^2=0.280$ in the Pd-to-Ni study \citep{Xu2025NiTransfer}; adding 138 Ni reactions from our own benchmark, mostly hydrogenations, did not help (Supplementary Note~\ref{app:hong}). In the Co study \citep{Zhang2023ChiralAcid}, ExtraTrees has the lowest MAE in source cross-validation although \rxndirect{} has the highest $R^2$. The authors' 27 descriptors were selected with all 69 labels, including the targets, so the target comparison favours the reference.

\begin{table}[htbp]
\centering\small
\caption{Sparse-data catalyst campaigns, original evaluation memberships. MAE in kcal\,mol$^{-1}$. ``Reference'' marks reference-relative responses; DT, decision tree; SVR, support-vector regression. Input handling of the frozen model differs between the two Co rows: $^{a}$native input ($R^2=0.724$ with an external SVD of the fingerprints); $^{b}$external SVD ($R^2=-0.013$ with native input); see Supplementary Table~\ref{tab:hong-debug}.}
\label{tab:smalldata}
\begin{tabular}{lllrr}
\toprule
Case & Method & $n$ & $R^2$ & MAE\\
\midrule
\multirow{7}{*}{\makecell[l]{Nature Case 1\\(official OOS)}} & Authors' MLR (published predictions) & 9 & $-0.091$ & 0.158\\
 & Authors' SISSO (published predictions) & 9 & -- & 0.108\\
 & Authors' MLR, differenced descriptors + reference & 9 & -- & 0.120\\
 & Authors' SISSO + reference & 9 & -- & 0.347\\
 & Additive reference formula (no model) & 9 & -- & 0.096\\
 & \rxndirect, fingerprints & 9 & -- & 0.252\\
 & \rxndirect, fingerprints + reference & 9 & \textbf{0.678} & \textbf{0.078}\\
\midrule
\multirow{4}{*}{\makecell[l]{Pd$\rightarrow$Ni Suzuki\\(21-fold LOO)}} & Authors' DT delta, MF descriptors & 21 & 0.630 & 0.107\\
 & \rxndirect, author RDKit descriptors, delta & 21 & 0.590 & 0.129\\
 & \rxndirect, role fingerprints, Ni context only & 21 & 0.280 & 0.153\\
 & 0.5\,DT + 0.5\,\rxndirect{} (fixed ensemble) & 21 & \textbf{0.713} & \textbf{0.102}\\
\midrule
\multirow{5}{*}{\makecell[l]{Co chiral acid}} & Authors' SVR, 27 selected features, 10-fold CV & 59 & 0.719 & 0.182\\
 & ExtraTrees, role fingerprints, 10-fold CV & 59 & 0.773 & \textbf{0.153}\\
 & \rxndirect, role fingerprints, 10-fold CV$^{a}$ & 59 & \textbf{0.790} & 0.166\\
 & Authors' SVR, delta, target LOO & 10 & \textbf{0.775} & \textbf{0.095}\\
 & \rxndirect, role fingerprints, delta, target LOO$^{b}$ & 10 & 0.267 & 0.150\\
\bottomrule
\end{tabular}
\end{table}

\begin{figure}[htbp]
\centering
\figinclude{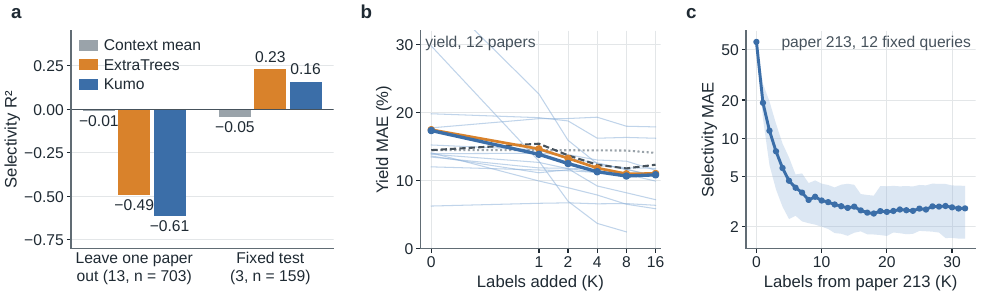}{figures/figS_heck.png}
\caption{\textbf{Sadphos Heck collection, further results.} \textbf{a}, Selectivity $R^2$ under leave-one-paper-out evaluation (13 papers, $n=703$) and on the fixed paper-disjoint test split ($n=159$) for the context mean, ExtraTrees and Kumo on the RTF input. \textbf{b}, Yield MAE (percentage points) on fixed held-out reactions from twelve papers (twelve queries per paper, six for paper 217) against the number $K$ of added labelled reactions from that paper (thin lines, Kumo per paper; thick, mean over papers for Kumo and ExtraTrees; dashed, mean of the added labels; dotted, context mean). \textbf{c}, Paper~213, examined because its zero-shot error was large: selectivity MAE on twelve fixed queries for $K=0$--$32$ (mean and 10--90\% range over ten random orders, log scale).}
\label{fig:heck}
\end{figure}

\paragraph{Sadphos Heck collection.} On the fixed paper-disjoint test split (159 reactions, 138 of them from one paper), ExtraTrees is better than Kumo ($0.230$ versus $0.159$; Supplementary Fig.~\ref{fig:heck}a). The zero-shot predictions for paper~213 are far off: its main ligand gives a mean selectivity of 97.05 in that paper but 37.67 in the six reactions with the same ligand elsewhere. Keeping the 659 reactions of the other papers as context and adding $K$ labelled reactions from paper~213 in ten random nested orders, MAE on twelve never-added reactions falls from $57.428$ at $K=0$ to $19.003$ with one label, $5.818$ with four and $3.257$ with eight, and stays between $2.673$ and $2.795$ for $K=16$--$32$ (Supplementary Fig.~\ref{fig:heck}c). $R^2$ on the twelve queries stays negative because their selectivities are nearly constant, so MAE is the informative metric here. Paper~213 was chosen after its errors were observed, which motivated the all-paper protocol of Fig.~\ref{fig:generalization}d,e.

\input{tex/transfer_controls}

\paragraph{Heck candidate list.} Across four Heck reaction types (papers 180, 200, 258 and 264), we predicted selectivity and yield for 2,498 substrate--ligand entries using all 703 historical reactions as context. Counting substrate and ligand stereochemistry gave 2,493 distinct pairs, of which 2,491 were absent from the context. These candidate predictions are excluded from performance estimates. For the Narasaka--Heck/silylation series (P12, paper 264), the archived structures encode an unsubstituted $O$-benzoyl oxime, whereas Tables~S1 and S5 of the original study use $O$-(2,3,4-trifluorobenzoyl) oximes \citep{CaoHeck2024}. The P12 results in Fig.~\ref{fig:generalization}d use the archived structures.

\section{Nature Sparse-Data Benchmark}
\label{app:nature}
We use the official repository of Gallarati et al.\ \citep{Gallarati_2026} at the released repository state with all official train/query memberships and the stored positive $\Delta\Delta G$ values; Supplementary Table~\ref{tab:nature-direct} compares absolute-response predictions on those memberships. Inputs are the 2,268 raw mechanistic descriptors, the union of the authors' selected features (13 for Case~1, 14 for Ni, 17 for CH) or ligand and substrate fingerprints (106 coordinates for Case~1, 82 for Ni); fingerprint projections are fitted on the context only. Note that the ``$R^2$'' in the authors' code is the squared Pearson correlation: for the official Case~1 linear predictions it is $0.760$, while the coefficient of determination is $-0.091$.

\begin{table}[htbp]
\centering\small
\caption{Direct (absolute-response) comparison on official memberships. MAE in kcal\,mol$^{-1}$. LORO, leave one reaction out; the MLR for Ni LORO is rebuilt with the published algorithm because per-row predictions are not released.}
\label{tab:nature-direct}
\begin{tabular}{lrrrrr}
\toprule
Task & $n$ & Authors' MLR & Raw descriptors & Author features & Fingerprints\\
\midrule
Case 1, official OOS & 9 & 0.158 & 0.497 & 0.491 & 0.252\\
Case 1, later measured candidates & 3 & 0.277 & 0.306 & 0.240 & 0.148\\
Case 2, new ligands & 10 & 0.295 & 0.370 & 0.289 & 0.382\\
Case 2, new reaction H & 17 & 0.184 & 0.288 & 0.202 & 0.229\\
LORO, CH A--C & 38 & 0.199 & 0.376 & 0.261 & 0.408\\
LORO, Ni D--G & 46 & 0.226 & 0.404 & 0.265 & 0.283\\
\bottomrule
\end{tabular}
\end{table}

\paragraph{Reference-relative ablation.} Supplementary Table~\ref{tab:nature-delta} applies the reference transformation of Eq.~\eqref{eq:delta} to all methods on the nine Case~1 queries. For the authors' models we keep their published feature sets and their cross-validated ensemble weighting (2-fold $\times$ 5 repeats, inverse CV-RMSE weights); ``$\Delta X$ and $\Delta y$'' additionally replaces the descriptors by their differences from the reference reaction. Freezing the original ensemble weights instead gives $\Delta y$ MAE $0.581$/$0.361$ and $\Delta X$+$\Delta y$ MAE $0.130$/$0.168$ for MLR/SISSO. Our reconstruction of the absolute MLR/SISSO models ($0.165$/$0.109$) differs slightly from the published predictions ($0.158$/$0.108$; maximum row difference $0.022$).

\begin{table}[htbp]
\centering\small
\caption{Reference-relative ablation on Nature Case 1 (nine official OOS reactions). MAE in kcal\,mol$^{-1}$.}
\label{tab:nature-delta}
\begin{tabular}{lrrrr}
\toprule
Method & Absolute & $\Delta y$ & $\Delta X$ and $\Delta y$ & $\Delta y$ RMSE / $R^2$\\
\midrule
Authors' MLR (reconstructed) & 0.165 & 0.571 & 0.120 & 0.597 / $-11.333$\\
Authors' SISSO (reconstructed) & 0.109 & 0.347 & 0.166 & 0.371 / $-3.758$\\
Kumo, raw descriptors & 0.497 & 0.173 & -- & 0.223 / $-0.715$\\
Kumo, author features & 0.491 & 0.128 & -- & 0.169 / 0.012\\
Kumo, fingerprints & 0.252 & \textbf{0.078} & -- & 0.096 / 0.678\\
\bottomrule
\end{tabular}
\end{table}

\paragraph{Checks on the direct results.} Because the direct Kumo results were weak, we checked the input pipeline before interpreting them. Old and new adapters agree to $1.2\times10^{-7}$ on identical inputs, target shift/scale inversion to $7\times10^{-7}$, cached and direct execution exactly, single and batched queries to $1.4\times10^{-4}$, and FP16 and FP32 to $3.3\times10^{-4}$. A linear synthetic task with 53 context rows and 13 features gives $R^2=0.999$, and 77,562 descriptor cells agree with the official master table. All nine Case~1 queries are over-predicted (mean bias $+0.252$ for fingerprints): the training mean is about 1.33 kcal\,mol$^{-1}$ and the test mean about 0.77, and one official Hirshfeld descriptor ranges over 0.63--1.80 in training but 4.97--8.44 in test. This shift motivated the reference-relative response. For Case~3, the 195 candidate workbooks lack measured values and per-round predictions, so it was not evaluated.

\section{Cobalt and Nickel Cases}
\label{app:hong}

\begin{figure}[htbp]
\centering
\figinclude{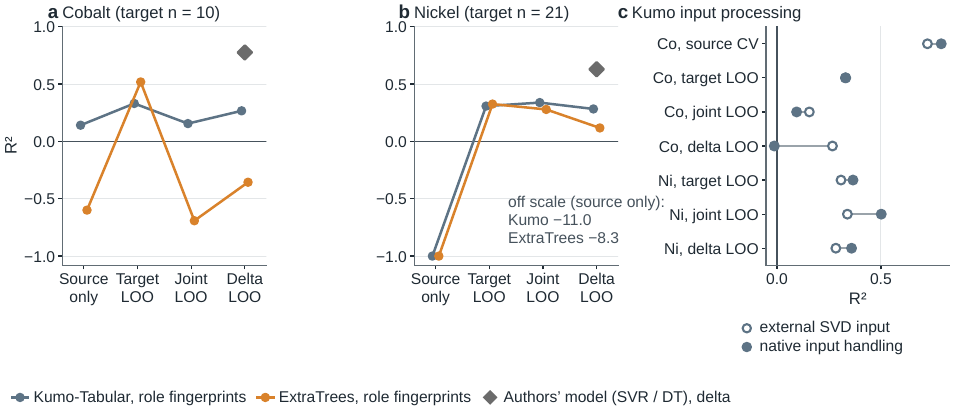}{figures/fig_hong_debug.pdf}
\caption{\textbf{Co and Ni cases.} \textbf{a},\textbf{b}, $R^2$ on the Co (10 targets) and Ni (21 targets) reactions for Kumo and ExtraTrees with role fingerprints under source-only, target leave-one-out, joint leave-one-out and delta leave-one-out protocols, with the authors' delta models (SVR, DT) as references. Ni source-only values are off scale and printed. \textbf{c}, Effect of compressing inputs with an external SVD versus native input handling on the frozen model.}
\label{fig:hong}
\end{figure}

Both cases use each study's measured data as context, not transfer from the main benchmark; Supplementary Fig.~\ref{fig:hong} collects the Co and Ni comparisons. For Co \citep{Zhang2023ChiralAcid}, 59 source reactions are evaluated with fixed local ten-fold cross-validation (the original random folds were not released), and the ten target reactions by leave-one-out with nine target labels in the context; 360 unmeasured virtual candidates are excluded. For Ni \citep{Xu2025NiTransfer}, we recover all 21 original leave-one-out folds with 20 Ni support reactions and the selected Pd reactions, and reproduce the authors' decision-tree delta model ($r=0.811$, MAE $0.107$, as published). The first comparisons compressed our inputs with an external SVD, which turned out to cost accuracy in several settings; Supplementary Table~\ref{tab:hong-debug} summarizes how input handling changes the result. The numerical calling path passed all checks (refit and fresh instances agree exactly; first-fold Ni predictions reproduce to $2\times10^{-5}$). Raw inputs do not help uniformly: for the ten Co targets the delta score falls from $0.267$ to $-0.013$. Among Ni context variants, role fingerprints with Ni context only give $R^2=0.280$, adding 138 internal Ni reactions gives $0.326$ (higher MAE), donor-local inputs with Pd and local Ni context give $0.496$, and decision-tree prior coordinates with a Kumo head give $0.604$; none exceeds the fixed 50/50 ensemble ($0.713$).

\begin{table}[htbp]
\centering\small
\caption{Effect of input handling on the frozen model in the Co and Ni cases (prediction $R^2$, all queries).}
\label{tab:hong-debug}
\begin{tabular}{lllrr}
\toprule
Case & Setting & Input & Initial & Revised\\
\midrule
Co & 10-fold CV ($n=59$) & role fingerprints, SVD $\rightarrow$ raw & 0.724 & 0.790\\
Co & target delta LOO ($n=10$) & role fingerprints, SVD $\rightarrow$ raw & 0.267 & $-0.013$\\
Ni & target LOO ($n=21$) & author MF, SVD $\rightarrow$ raw & 0.368 & 0.489\\
Ni & joint context LOO & author MF, SVD $\rightarrow$ raw & $-1.447$ & 0.468\\
Ni & delta LOO & author MF, SVD $\rightarrow$ raw & $-0.381$ & 0.305\\
Ni & delta LOO & author RDKit, default $\rightarrow$ standardization & 0.557 & 0.590\\
\midrule
Co & 10-fold CV & ExtraTrees, raw role fingerprints & \multicolumn{2}{r}{0.773}\\
Ni & delta LOO & authors' MF decision tree & \multicolumn{2}{r}{0.630}\\
\bottomrule
\end{tabular}
\end{table}

\section{\texorpdfstring{Co-AHO, the Out-of-Sample Hydrogenation Set and Buchwald--Hartwig}{Co-AHO, out-of-sample hydrogenation and Buchwald--Hartwig}}
\label{app:newdata}
\paragraph{Datasets.} Co-AHO is the 532-reaction Co-catalysed asymmetric hydrogenation collection of Singh and Hern{\'a}ndez-Lobato \citep{Singh_2025} (16 publications, 366 substrate structures, 59 ligands, 419 reactions at or above 80\% ee). Products in the source file are recorded for all 532 reactions; 530 match the constitution given by hydrogenating the non-aromatic C=C bond, and the two that differ (a methyl carbamate reported as acetamide, and a propargyl substrate reported as a coupled ester) are kept unchanged. Labels are $|\Delta\Delta G^\ddagger|=RT\,|\ln[(1+\mathrm{ee})/(1-\mathrm{ee})]|$ at the reported temperature, capped at 4.5005~kcal\,mol$^{-1}$; no configuration is recorded, so the sign is not modelled. Their official out-of-sample set contains 246 Ir (163) and Rh (83) reactions from five 2022--2023 publications, none of which appears in the benchmark; the product column is empty and the products were generated as described in Methods (211 by hydrogenating a single non-aromatic C=C bond, 35 chromones by reducing the pyranone ring of an aromatic Kekul\'e form). The generator reproduces the constitution of all 1,766 AHO substrates and of 530 of the 532 Co products. Neither Singh file overlaps the benchmark by DOI, so the AHO context for both tasks contains 9,358 and 9,478 reactions respectively. The Buchwald--Hartwig set is the 3,955-reaction screen of Ahneman et al.\ \citep{Ahneman_2018}, complete over 15 aryl halides $\times$ 4 ligands $\times$ 3 bases $\times$ 22 isoxazole additives (3,955 of the 3,960 combinations); yields are the reported GC yields in per cent.
\paragraph{Splits.} Co-AHO: five folds that keep every canonical substrate in one fold (366 substrates, 80--139 test reactions per fold), and 16 leave-one-publication-out folds (9--58 test reactions). Transfer folds add the non-Co AHO reactions to the context with the Co training rows of the same fold, with $K=16$ or $64$ random Co training rows (three draws), or with no Co row. Out-of-sample set: ten draws of 64 support and 128 query reactions from the 246, the support added to the AHO context or used alone. Buchwald--Hartwig: the ten released random 70/30 splits (2,767 training, 1,188 test reactions) and Tests~1--4 (3,057, 3,055, 3,058 and 3,055 training reactions; 898, 900, 897 and 900 test reactions with additives absent from training).
\paragraph{Published numbers.} Singh and Hern{\'a}ndez-Lobato report AUPRC at 80\% ee on ten random tasks (ProtoNet 0.9133, random forest 0.8369, GNN 0.8259) and on the out-of-sample set with 64 support and 128 query reactions (ProtoNet 0.9341, GNN 0.8764, random forest 0.8631). Their released scripts read cluster feature files from absolute paths on the authors' machine that are not part of the repository, so these models could not be rerun here. Their random-task protocol differs from our grouped folds (random tasks with 64 support and 128 query reactions, not substrate-held-out folds), so that comparison is not made. For Buchwald--Hartwig we use Table~2 of Probst et al.\ \citep{Probst_2022}, which reports the DFT-descriptor random forest of Ahneman et al.\ (0.92, 0.80, 0.77, 0.64, 0.54), Yield-BERT \citep{Schwaller_Yield_2021} (0.95, 0.84, 0.84, 0.75, 0.49), its augmented variant (0.97, 0.80, 0.88, 0.56, 0.43) and DRFP with XGBoost (0.95, 0.81, 0.83, 0.71, 0.49) for the random 70/30 split and Tests~1--4.
\paragraph{Embedding map.} Classes and row counts before the per-class cap: AHO 9,478, KetRed 4,232, AsymOx-SAD 688, AsymOx-ICat 828, Heck 703, CPA thiol additions 1,075, Co-AHO 532 and Buchwald--Hartwig 3,955; the map shows 600 reactions from each class except Co-AHO, for which all 532 are shown (4,732 reactions in total). The raw RTF projections were fitted on the pooled structures without labels. TabPFN~3.5 is fitted on each training fold with its balanced feature subsampling and pretraining limits disabled, and \texttt{get\_embeddings} returns the final-layer test-token representation for each of the eight estimators; the mean over estimators is used (1,024 dimensions). Kumo embeddings are the inputs of its in-context output head, averaged over the eight estimators (1,024 dimensions, as recorded in the extraction artefacts). Quality metrics are computed in the full space with the cosine metric, not in the 2-D map. The raw-input comparison uses the same procedure (PCA-free RTF, 126--468 columns depending on the pooled set).
\input{tex/newdata_si.tex}

%% file: tex/dataset_sources.tex
All benchmark records are experimental results reported in the chemical literature, taken from public research datasets and local literature compilations. We assemble them into a common six-role schema; no new laboratory measurements are introduced. The three source names denote provenance partitions.

\paragraph{KetRed: literature-curated asymmetric ketone reduction.} KetRed (internal file name: Chiral) consists of asymmetric reductions of prochiral ketones to secondary alcohols: an RDKit check of substrate and product finds a C=O to C--OH conversion in 4,229 of the 4,232 records (221 of them conjugated enones or quinolones whose C=C is reduced as well); two records cannot be parsed and one has inconsistent substrate and product.
The 4,232 records come from the \texttt{chiral\_carbonyl\_data.xlsx} literature compilation, which records substrate and product structures, catalyst and ligand, conditions, conversion, ee and a DOI or literature-title field. References are usually given at section headings rather than repeated for every row, so the number of rows is not the number of independent studies, and we do not forward-fill empty reference cells as if they were verified row-level citations. The original benchmark membership is retained.

\paragraph{AsymOx: two literature-derived asymmetric oxidations.} AsymOx (internal file name: Screened) groups two asymmetric oxidations with magnitude labels.
The 1,516 records comprise 688 Sharpless asymmetric dihydroxylations from the literature-derived collection released by Ocampo et al.\ \citep{Ocampo2025} (688 is the number usable in this benchmark, not the full database size), and 828 hypervalent-iodine-catalysed asymmetric dearomatizations from the \texttt{ICat.xlsx} compilation. Hypervalent-iodine(III)-catalysed asymmetric phenolic dearomatizations of this kind were compiled and modelled by Gao et al.\ \citep{Gao2024Iodine}; the record-level correspondence between \texttt{ICat.xlsx} and that compilation has not yet been verified, so we cite it as the source study of this reaction class rather than as the file's origin. All 688 SAD records convert an alkene into a 1,2-diol, and the ICat records are oxidative dearomatizations of phenols and naphthols to dienones and spirolactones. For 33 ICat records (2\%) the drawn product is inconsistent with such an oxidation---typically a dienone carbonyl recorded as an alcohol---which suggests structure-entry errors; we keep the records unchanged and flag them here. The AsymOx split is row-based.

\paragraph{AHO: asymmetric hydrogenation of olefins.}
AHO is the literature database of asymmetric olefin hydrogenations compiled by Hong and co-workers (12,619 reactions published 2000--2020; \citealp{Xu2021AHO}; asymcatml.net). We use the curated 9,478-reaction subset with a signed $\Delta\Delta G^\ddagger$ target released as \texttt{9478\_file.csv} with ChemAHNet \citep{Cheng_2025}, keep its 940-reaction test split and its signed target convention, and refer to it as AHO throughout.

File hashes, retained counts and bibliography fields are recorded in the aggregate evidence accompanying the manuscript source. Conversion fields are never relabelled as isolated yield, and reference identifiers are never used as predictive features. Supplementary Table~\ref{tab:datasets} lists each source with its record counts and the part it plays.

\begin{table}[htbp]
\centering\footnotesize
\setlength{\tabcolsep}{3pt}
\caption{\textbf{Every dataset used in this work.} Records are the number of reactions entering that analysis, which for some collections is smaller than the number published because of parsimony and label checks described in the corresponding note. ``Role'' states the question the collection answers. Sources are cited in the text and listed in the bibliography.}
\label{tab:datasets}
\begin{tabular}{@{}p{2.6cm}p{1.15cm}p{3.1cm}p{4.0cm}@{}}
\toprule
Dataset & Records & Reaction class & Role in this work\\
\midrule
\multicolumn{4}{@{}l}{\emph{Benchmark}}\\
AHO & 9,478 & Olefin hydrogenation & Main benchmark; signed $\Delta\Delta G^\ddagger$ \citep{Xu2021AHO,Cheng_2025}\\
KetRed & 4,232 & Ketone reduction & Main benchmark; magnitude labels\\
AsymOx & 1,516 & Dihydroxylation, dearomatization & Main benchmark; magnitude labels \citep{Ocampo2025,Gao2024Iodine}\\
\midrule
\multicolumn{4}{@{}l}{\emph{Transfer and interpolation limits}}\\
OrgAIcat aldol & 13,360 & Organocatalytic aldol & Random and catalyst-held-out transfer \citep{OrgAIcat2026}\\
OrgAIcat Michael & 9,194 & Organocatalytic Michael & Random and catalyst-held-out transfer \citep{OrgAIcat2026}\\
$N,N'$-dioxide Michael & 1,651 & Lewis-acid Michael & Random and ligand-held-out transfer \citep{TangMichael2026}\\
$\beta$-C(sp$^3$)--H activation & 220 & Palladium C--H activation & New bond-forming class \citep{Hoque2025CSP3H}\\
Rh hydrogenation screen & 1,224 & Rh hydrogenation HTE & Same class, dense precedents \citep{Kalikadien2024}\\
\midrule
\multicolumn{4}{@{}l}{\emph{Adaptation to new campaigns}}\\
Co-AHO & 532 & Cobalt hydrogenation & Substrate-held-out and leave-one-publication-out \citep{Singh_2025}\\
Buchwald--Hartwig & 3,955 & C--N coupling, yield & Ten random splits and additive-out-of-sample Tests 1--4 \citep{Ahneman_2018}\\
Sadphos Heck & 703 & Palladium Heck & Few-shot adaptation across papers \citep{Li_Zhang_2024}\\
Zahrt CPA thiol addition & 1,075 & Chiral phosphoric acid & Out-of-substrate and out-of-catalyst extrapolation \citep{Zahrt_2019}\\
Ir/Rh out-of-sample set & 246 & Iridium, rhodium hydrogenation & AUPRC at 80\% ee against published models \citep{Singh_2025}\\
\midrule
\multicolumn{4}{@{}l}{\emph{Published sparse-data studies}}\\
Gallarati C--H and Ni cases & 30 & C--H activation, cross-coupling & Reference-relative prediction with the authors' descriptors \citep{Gallarati_2026}\\
Pd-to-Ni transfer & 20 & Nickel cross-coupling & Transfer from palladium data \citep{Xu2025NiTransfer}\\
Cobalt chiral acids & 69 & Cobalt Br\o{}nsted acid & Transfer to new acids \citep{Zhang2023ChiralAcid}\\
\bottomrule
\end{tabular}
\end{table}

%% file: tex/results_details.tex
The main text reports only the key numbers. This note gives the full quantitative statements behind each Results subsection, in the order of the main text.

\paragraph{A frozen foundation model predicts enantioselectivity across three literature datasets without training.}

We compiled 15,226 enantioselectivity records from published experiments into three datasets (Fig.~\ref{fig:main}a and Methods). AHO, the asymmetric hydrogenation of olefins database \citep{Xu2021AHO}, contributes 9,478 reactions with signed $\Delta\Delta G^\ddagger$ labels in the curated subset and 940-reaction test split released with ChemAHNet \citep{Cheng_2025}; KetRed contains 4,232 asymmetric ketone reductions, and AsymOx 1,516 asymmetric oxidations (688 Sharpless dihydroxylations and 828 hypervalent-iodine-catalysed dearomatizations), both with magnitude labels. Following standard practice, each model is trained, or given a context, on one dataset at a time, and all design choices were selected on validation reactions only.

In this per-dataset setting RxnPFN is the best or tied-best model on AHO and KetRed and has the highest macro average (Fig.~\ref{fig:main}b and Table~\ref{tab:main}). Averaged over three seeds as in Table~\ref{tab:main}, RxnPFN reaches test $R^2=0.805$ on AHO, $0.767$ on KetRed and $0.712$ on AsymOx (macro average $0.761$), against $0.730$ for the best tree model (ExtraTrees on R-SPOC: $0.729$, $0.756$ and $0.706$), $0.728$ for the best other pretrained configuration (Kumo on R-SPOC: $0.741$, $0.739$ and $0.704$) and $0.691$ for Chemprop ($0.726$, $0.612$ and $0.734$ per-seed mean; XGBoost reaches $0.725$ on AsymOx). In the single-seed reference evaluation used for paired cluster-bootstrap contrasts (RxnPFN $0.805$, $0.764$ and $0.716$; macro $0.762$), the paired differences to ExtraTrees are $+0.077$ (95\% interval $[+0.044,+0.112]$) on AHO and $+0.012$ ($[-0.064,+0.094]$) on KetRed, and against the prediction-averaged three-seed Chemprop ensemble ($0.748$, $0.650$ and $0.752$) they are $+0.057$ ($[+0.022,+0.089]$), $+0.114$ ($[+0.061,+0.179]$) and $-0.036$ ($[-0.117,+0.042]$). A fine-tuned reaction language model and the differential reaction fingerprint are weaker on every dataset (rxnfp-BERT $0.591$, $0.580$ and $0.588$; DRFP with XGBoost $0.648$, $0.690$ and $0.684$; all paired intervals to RxnPFN above zero except DRFP on AsymOx). The predictions follow the measured values over the full range of each dataset (Fig.~\ref{fig:main}c). Across five model seeds RxnPFN varies by $\pm0.002$ in pooled test $R^2$ (Supplementary Note~\ref{app:benchmark}).

ChemAHNet was designed and trained for exactly this hydrogenation dataset, so its own benchmark is the most direct comparison with a dataset-specific deep model. On the 940 test reactions of the published split, ChemAHNet reports $R^2=0.591$ for the signed $\Delta\Delta G^\ddagger$ \citep{Cheng_2025}; retraining it from the released code with validation-based checkpoint selection gives $0.492\pm0.032$ over three seeds. RxnPFN, given the AHO training reactions as context without any training, reaches $0.805$ on the same reactions and predicts the favoured enantiomer for 96.9\% of them, against 86\% for the published predictions and 82--85\% for the retrained models (Fig.~\ref{fig:main}b). Higher figures in the original study refer to other tasks, namely the classification of the major configuration, ten-fold cross-validation within a single catalyst family and a separate organocatalysis benchmark (Methods and Supplementary Note~\ref{app:benchmark}).

The advantage is not an artefact of one split or one backbone. Over five additional random re-splits per dataset, grouped by substrate--product pair, RxnPFN reaches $0.745\pm0.036$ on AHO, $0.741\pm0.036$ on KetRed and $0.723\pm0.043$ on AsymOx (Supplementary Fig.~\ref{fig:resplits}). It beats every baseline in all five AHO splits (ExtraTrees by $+0.173$, 95\% $t$-interval $[+0.123,+0.223]$), beats ExtraTrees, TabPFN and XGBoost in all five KetRed splits (ExtraTrees by $+0.068$ $[+0.036,+0.100]$) and TabPFN in all five AsymOx splits ($+0.036$ $[+0.024,+0.049]$); against Kumo on R-SPOC on KetRed and against the tree models on AsymOx the differences are not significant. The recipe is also not specific to Kumo (Fig.~\ref{fig:main}f). Applied to three other frozen tabular foundation models, the RTF input with the magnitude--direction readout raises pooled test $R^2$ from $0.795$ to $0.840$ for TabPFN, from $0.780$ to $0.829$ for TabICL and from $0.766$ to $0.814$ for TabDPT, and on AHO per dataset from $0.724$ to $0.792$, $0.706$ to $0.785$ and $0.688$ to $0.750$; every gain has a paired interval above zero, and Kumo remains the strongest backbone.

The same frozen model can also take all three datasets as one context. In this pooled setting RxnPFN reaches test $R^2=0.856$ (MAE $0.430$, RMSE $0.750$ kcal\,mol$^{-1}$) over all 1,501 test reactions, averaged over five seeds, against $0.807$ for ExtraTrees on R-SPOC, $0.816$ for Kumo on R-SPOC, $0.794$ for TabPFN 3.5 and $0.795$ for Chemprop (Table~\ref{tab:main}). Pooling helps where labels are signed and slightly hurts the magnitude-only datasets, so for a small magnitude-only campaign its own reactions are the better context (Supplementary Fig.~\ref{fig:settings}).

On the enantiomeric-excess scale on which chemists read selectivity, using each reaction's temperature, the per-dataset models reach MAE $15.2$, $7.0$ and $7.8$ percentage points on AHO, KetRed and AsymOx, against $17.9$, $6.9$ and $8.0$ for ExtraTrees on R-SPOC (Fig.~\ref{fig:main}e). For catalyst selection, the relevant question is whether the model ranks candidates correctly under fixed substrate and conditions. In the 82 test groups with at least three distinct catalysts (396 reactions, 4.8 catalysts per group), scored by selectivity strength $|\Delta\Delta G|$, per-dataset RxnPFN identifies the best catalyst in 68.3\% of groups (random 25.6\%), places it among its top three in 96.3\%, and has a mean regret of $0.137$ kcal\,mol$^{-1}$ (random $0.832$), with mean Spearman correlation $0.616$ (Fig.~\ref{fig:main}g). ExtraTrees on R-SPOC reaches $65.9\%$, $0.256$ kcal\,mol$^{-1}$ and $0.523$, and on RTF $62.2\%$, $0.196$ kcal\,mol$^{-1}$ and $0.570$. \rxndirect{} is close ($67.1\%$, $0.153$ kcal\,mol$^{-1}$, $0.634$), and when AHO groups are ranked by signed $\Delta\Delta G$ it is better than RxnPFN (top-1 $74.6\%$ versus $67.8\%$, regret $0.098$ versus $0.130$ kcal\,mol$^{-1}$), so the decomposition does not improve the ranking of candidate catalysts. Because no weights are trained, a new task requires only feature construction and one forward pass (Fig.~\ref{fig:main}h): on one H200 GPU the full two-head pipeline encodes the 13,181-reaction context and predicts all 1,501 test reactions in about 12 s, after about 3 min of CPU feature construction, whereas one pooled training run takes about 27 s for DRFP with XGBoost, 2.5 min for the fine-tuned rxnfp-BERT and 44 min for Chemprop (Supplementary Table~\ref{tab:cost}).

\paragraph{The reaction encoding and a label-matched response drive accuracy, symmetry and calibration.}

With the backbone fixed, replacing the R-SPOC descriptor by the RTF raises pooled test $R^2$ from $0.816$ to $0.845$ (paired $\Delta R^2=+0.030$, 95\% interval $[+0.017,+0.044]$). The same coordinates help trees less: ExtraTrees reaches $0.814$, so the frozen model gains $+0.031$ ($[+0.020,+0.043]$) over a tree on identical input. Separating magnitude and direction gives the largest validation improvement of any design choice, from $0.744$ to $0.792$, once signed and magnitude labels are no longer conflated in one head; on test it adds $+0.010$ ($[+0.001,+0.019]$), and averaged over five seeds $+0.015$. The decomposition does not help trees in the same way: an ExtraTrees magnitude/direction pair on the same inputs reaches $0.815$ against $0.814$ for direct ExtraTrees. Removing one block of the RTF at a time shows that the AtomPair views carry the most information: without them validation $R^2$ of \rxndirect{} falls from $0.744$ to $0.696$ and test $R^2$ from $0.845$ to $0.830$ (Fig.~\ref{fig:ablation}a). The gained/lost environments matter on validation ($0.717$) but little on test ($0.843$), metadata contribute little, and the role-wise Morgan block is redundant once the other blocks are present (validation $0.749$, test $0.848$ without it). ExtraTrees is insensitive to every single removal (test $0.802$--$0.814$). Curating the context, retrieving similar reactions or adding nine further chemical inputs and fifteen embedding variants did not improve on the RTF (Supplementary Note~\ref{app:exploration}).

On AHO, the only dataset with signed labels, getting the favoured enantiomer right is much easier than getting the selectivity strength right (Fig.~\ref{fig:ablation}b). With AHO reactions as the only context, RxnPFN predicts the favoured enantiomer for 96.9\% of the 904 test reactions with non-zero labels, against 95.7\% for ExtraTrees on R-SPOC and 96.5\% on the RTF input, whereas $R^2$ of $|\Delta\Delta G|$ is $0.479$, against $0.805$ for the signed values. The frozen model keeps a clear margin over ExtraTrees on strength ($0.479$ versus $0.424$ on R-SPOC and $0.306$ on RTF), and \rxndirect{} is slightly better than RxnPFN on this component ($0.495$), consistent with the decomposition helping mainly through the direction. This matters for interpreting pooled scores: a reference predictor that knows only the true sign and the training mean $|\Delta\Delta G|$ of each dataset reaches pooled test $R^2=0.763$, compared with $0.268$ for the per-dataset mean alone, which is why per-dataset, sign and magnitude scores are reported (Supplementary Table~\ref{tab:components}).

The product is part of every reaction record, and its stereochemical annotation can reveal the major enantiomer. We therefore rebuilt the inputs with all product stereochemistry removed (constitution only) and with the product removed entirely (Fig.~\ref{fig:ablation}c). The dependence is concentrated in AHO, where removing product stereochemistry lowers $R^2$ of RxnPFN from $0.805$ to $0.716$ (\rxndirect{} $0.786$ to $0.716$) and its sign accuracy from $0.969$ to $0.897$; on KetRed and AsymOx the changes are small ($0.764$ to $0.754$ and $0.716$ to $0.697$ for the constitution-only input), and on AsymOx ExtraTrees is on par with the frozen model without product stereochemistry ($0.703$ versus $0.697$). Without the product the frozen model outperforms ExtraTrees on the same coordinates by a larger margin than with it: on AHO RxnPFN leads by $+0.146$ $[+0.111,+0.182]$ (constitution) and $+0.173$ $[+0.136,+0.215]$ (product-free). For screening before an experiment, when the major enantiomer is unknown, these product-free scores are the relevant ones. Labels at the measurement ceiling ($|\mathrm{ee}|\ge99.5\%$, stored as clipped $\Delta\Delta G^\ddagger$) matter mainly for KetRed: capping the context labels at 99\% ee or dropping the clipped context labels lowers the per-dataset test $R^2$ of RxnPFN from $0.764$ to $0.561$ and $0.471$, whereas on AHO and AsymOx both changes stay within $0.015$ (Fig.~\ref{fig:ablation}d). Scored without the clipped test reactions, $R^2$ is $0.822$, $0.696$ and $0.710$ on AHO, KetRed and AsymOx.

A model of enantioselectivity should also respect its mirror symmetry: the enantiomeric reaction must give the opposite sign and the same magnitude. We inverted every tetrahedral stereocentre of the 772 test reactions whose catalyst carries tetrahedral stereocentres and no biaryl axis or metallocene (501 AHO, 193 KetRed, 78 AsymOx), recomputed all inputs and predicted with the same per-dataset context (Fig.~\ref{fig:ablation}e,g). For AHO RxnPFN flips the predicted sign in 86.0\% of cases (\rxndirect{} 80.2\%, ExtraTrees on RTF 71.9\%), and the mirrored prediction correlates with the negated original prediction at $r=0.780$ ($0.671$, $0.568$); for the magnitude-only datasets the mirrored prediction correlates with the original at $r=0.925$ (KetRed) and $0.928$ (AsymOx). Against negated labels the mirrored AHO reactions reach only $R^2=0.453$, against $0.763$ for the original reactions, so the model learned part of its accuracy from recorded catalyst--product configurations rather than from a symmetric description of selectivity. Because the context of an in-context model is just a table of measured reactions, the symmetry can be supplied as data: adding the mirror images of the 4,463 eligible AHO training reactions, with negated labels, to the context raises the sign-flip rate to 99.2\%, the correlation with the negated original prediction to $r=0.996$ and $R^2$ of the mirrored reactions against negated labels to $0.740$, the level of the original reactions (Fig.~\ref{fig:ablation}f,g). The accuracy on the recorded test reactions is unchanged within noise ($0.797$ versus $0.805$; paired difference $-0.008$ $[-0.027,+0.008]$). Labels at the top of the ee scale are also poorly resolved; scoring without the clipped high-ee test reactions, or capping or removing such labels in the context, changes the conclusions little (Supplementary Note~\ref{app:benchmark}).

The two-head readout also changes what the model knows about its own errors. With each dataset given only its own proper-fit reactions as context and calibration reactions held out by substrate--product pair, native 90\% intervals of the single-head model \rxndirect{} cover 87.3\% of AHO, 90.0\% of KetRed and 93.1\% of AsymOx test reactions (Fig.~\ref{fig:ablation}j and Supplementary Fig.~\ref{fig:smalldata}d), and in the pooled setting native 80\%, 90\% and 95\% intervals cover 80.8\%, 90.8\% and 95.1\%. Scaling the magnitude quantiles by the clipped mean sign, however, ignores the bimodal uncertainty of the direction, and the resulting two-head intervals cover only 73.2\% of AHO test reactions at the nominal 90\% level. Treating the predictive distribution as a mixture of the magnitude distribution and its mirror image, weighted by the direction probability, restores coverage to 90.5\% and lowers the 90\% interval score from $6.33$ to $2.93$ kcal\,mol$^{-1}$, below the $3.38$ of the single-head model, with the best CRPS of the three readouts ($0.40$ against $0.49$ and $0.44$) and unchanged point predictions (Fig.~\ref{fig:ablation}h,i). Interval width also ranks errors: accepting the reactions whose single-head width is below the calibration median retains 69.7\%, 64.2\% and 51.6\% of AHO, KetRed and AsymOx test reactions and lowers their MAE from $0.645$ to $0.391$, $0.364$ to $0.223$ and $0.282$ to $0.151$ kcal\,mol$^{-1}$ (Fig.~\ref{fig:ablation}k and Supplementary Fig.~\ref{fig:smalldata}f).

\paragraph{Accuracy falls under extrapolation but recovers within a few labels from a new campaign.}

The benchmark splits follow the original sources and mostly measure interpolation among related reactions. With all settings frozen, we re-evaluated RxnPFN on three-fold splits that hold out ligand families, AHO publications or substrate scaffolds, with random folds over substrate--product pairs as the reference; each fold keeps about two thirds of the 15,226 reactions as context (Fig.~\ref{fig:generalization}b). Grouping costs every model a large part of its accuracy. RxnPFN falls from $0.765\pm0.008$ on the random reference to $0.611\pm0.063$ with ligand families held out, $0.498\pm0.017$ with AHO publications held out and $0.460\pm0.085$ with substrate scaffolds held out. Its lead is preserved under every grouping: ExtraTrees reaches $0.558$, $0.430$ and $0.405$, and TabPFN $0.596$, $0.441$ and $0.403$; RxnPFN is better than ExtraTrees in eight of the nine grouped folds. The decomposition helps most under the publication split ($0.498$ versus $0.452$ for \rxndirect{}), consistent with the direction being transferable across papers when the product is recorded. Predicting a new substrate class or a new paper remains much harder for all models, and the frozen model narrows but does not close this gap. On the out-of-sample splits of the Zahrt chiral phosphoric acid benchmark \citep{Zahrt_2019}, RxnPFN is close to published methods on new substrates (MAE $0.166$ kcal\,mol$^{-1}$ against $0.135$--$0.161$) and, with the single-role encoding (Methods), on new catalysts ($0.253$ against $0.211$--$0.254$), but not on the split that combines new substrates and catalysts (Supplementary Fig.~\ref{fig:extrapolation}).

Data efficiency is the central argument for in-context prediction. With 32, 128, 512, 2,048 and all context reactions (five random subsets per budget), RxnPFN leads at every budget, and the margin is largest with the fewest labels (Fig.~\ref{fig:generalization}a): with 32 context reactions it reaches $R^2=0.263\pm0.032$ against $0.192\pm0.061$ for ExtraTrees, $0.180$ for TabPFN and $-0.164$ for XGBoost; at 512 and 2,048 reactions it reaches $0.606$ and $0.730$ against $0.535$ and $0.663$ for ExtraTrees. On the same RTF input \rxndirect{} is close to TabPFN and ExtraTrees below 128 reactions and pulls ahead from 512 reactions on, so at small budgets most of the advantage comes from the decomposition rather than from the backbone. At no budget is accuracy high in absolute terms: on this heterogeneous benchmark a few hundred labels are not enough for any model.

The Sadphos Heck collection of 703 published reactions from 13 papers (Sadphos ligand class; cited here via Singh et al.\ \citep{Li_Zhang_2024,Singh_2025}), assembled independently of the benchmark, shows what happens when a new publication arrives. Its selectivity is reported on an unspecified scale (ee or enantiomeric ratio), so we model it as reported. Zero-shot transfer to an unseen paper fails: in leave-one-paper-out evaluation every model has a negative selectivity $R^2$, below the context-mean predictor ($-0.010$), namely $-0.612$ for Kumo and $-0.492$ for ExtraTrees on the RTF input (Supplementary Fig.~\ref{fig:heck}). The main ligand of paper 213 illustrates why: it gives a mean selectivity of 97.05 in the 37 reactions of that paper but 37.67 in the six reactions in which it appears in other papers. Adaptation inside the campaign, however, is fast. For every paper with at least twelve reactions (12 of 13), we kept the other papers as context, added $K$ labelled reactions from the held-out paper in ten random orders and scored twelve reactions that are never added (Fig.~\ref{fig:generalization}d,e). Averaged over papers, Kumo's selectivity MAE falls from $18.10$ at $K=0$---worse than the context mean ($10.75$)---to $9.22$ with one label, $5.93$ with four across all twelve papers and $4.83$ with sixteen across eleven papers, while ExtraTrees adapts more slowly ($13.13$, $7.29$ and $5.19$). The relevant control is the mean of the $K$ added labels, which captures a pure shift of the paper's level and stays between $10.8$ and $13.1$. Kumo beats this control in 9 of 12 papers with one label and in all 12 with four, against 5 and 10 of 12 for ExtraTrees, so the gain is more than recentring. In the four papers with fixed-query $R^2$ of at least $0.5$ at sixteen in-paper labels (Fig.~\ref{fig:generalization}g), $R^2$ goes from $0.61$ to $0.67$ (paper 164, MAE $5.33\to3.85$), $0.62$ to $0.61$ (258, MAE $5.97\to4.64$), $0.22$ to $0.61$ (180, MAE $2.82\to2.21$) and $-7.33$ to $0.58$ (200, MAE $16.50\to5.46$). For papers whose twelve held-out selectivities are nearly constant, such as 213, 217 and 264, $R^2$ stays negative even when the predictions are close, because the denominator of $R^2$ is close to zero; there MAE is the informative metric and is reported above. Trained deep models do not remove the zero-shot gap in the same way: Chemprop, retrained for every cell, is on par with the frozen model at $K=0$ ($16.78$ versus $17.51$ over the first three orders) and slightly worse once labels are added (e.g.\ $7.15$ versus $6.19$ at $K=4$); a fine-tuned rxnfp-BERT transfers better zero-shot ($10.84$ versus $17.51$ at $K=0$, lower in 9 of 12 papers) and is on par after one label ($8.99$ versus $9.77$, not significant). In simulated optimization campaigns that start from two random reactions of a paper and let the model choose the next twenty, greedy selection by RxnPFN recovers 74\% of each paper's three most selective reactions within the budget, against 47\% for random selection and 63\% for ExtraTrees (Wilcoxon $p=0.008$ against random; a secondary metric, defined after the primary one), whereas the number of reactions needed to reach the single most selective one does not differ from random ($12.9$ versus $13.6$), because many reactions of a paper lie within a few units of the maximum (Fig.~\ref{fig:generalization}f,h).

\paragraph{In sparse published studies, matching the response to the campaign design matters.}

We next applied the same frozen model to three published small-data studies, using each study's own measured reactions as context and its original evaluation membership (Fig.~\ref{fig:generalization}c and Supplementary Fig.~\ref{fig:smalldata}a--c). These tasks have tens of labels, and descriptor models designed by the original authors are strong references.

Gallarati et al.\ released mechanistic descriptors and official train/test memberships for several asymmetric reactions \citep{Gallarati_2026}. Applied directly, the frozen model is weaker than the authors' models: on the nine official out-of-sample reactions of Case~1 the fingerprint model has MAE $0.252$ kcal\,mol$^{-1}$ against $0.158$ for the published MLR and $0.108$ for SISSO, and all nine predictions are too high. The training set, however, is a ligand screen on substrate P0 plus a substrate scope with ligand L6, and the reference reactions L6/P16 and L6/P23 for all nine test queries are measured training reactions. Predicting the deviation from this reference, with the same inputs, context and weights, lowers MAE to $0.078$ kcal\,mol$^{-1}$ ($R^2=0.678$), below SISSO, below the best author variant with the same transformation ($0.120$) and below a fixed additive formula that needs no model ($0.096$); Chemprop and the fine-tuned rxnfp-BERT reach $0.201$ and $0.202$ with the same reference-relative target and $0.377$ and $0.519$ without it. This variant was developed after the direct out-of-sample errors were known, and nine queries are too few to establish a significant difference from the additive formula. The gain does not carry over to Case~2, where no comparable measured reference exists for most queries and the reference-relative response raises the error; a single fixed admission rule for references admits almost no query outside Case~1 (Supplementary Note~\ref{app:campaigns}). The reference-relative response is therefore useful when the campaign is designed around a measured reference, and not as a general correction.

For Pd-to-Ni transfer of atroposelective Suzuki couplings \citep{Xu2025NiTransfer}, we reused all 21 leave-one-out folds with their selected Pd support sets and reproduced the authors' decision-tree delta model ($R^2=0.630$, MAE $0.107$). With the authors' RDKit descriptors \rxndirect{} reaches $R^2=0.590$, and averaging the decision-tree and Kumo delta predictions with weights fixed at $0.5$ before scoring gives $R^2=0.713$ and MAE $0.102$, better than either model; we report this as an ensemble extension rather than as a result of the frozen model alone. For the 59 source reactions of the cobalt chiral-acid study \citep{Zhang2023ChiralAcid}, \rxndirect{} on raw role fingerprints reaches $R^2=0.790$ in ten-fold cross-validation, above the authors' SVR on 27 selected descriptors ($0.719$) and ExtraTrees ($0.773$). For the ten new target acids, however, the authors' delta SVR is clearly better ($R^2=0.775$ versus $0.267$), so in this case the frozen model does not transfer from source to target reactions as well as a task-specific model (Supplementary Note~\ref{app:campaigns}).

\paragraph{The same frozen model predicts yield and conversion.}

Selectivity is only one outcome of a reaction. Without any change to the backbone or the encoding, we gave the frozen model the reported conversions of the KetRed reductions and the reported yields of the AsymOx oxidations as targets (Fig.~\ref{fig:yield_main}a,b). Conversion is never relabelled as yield. We compared the frozen model, with the FDE input fixed as primary before scoring, with 21 other model and input configurations on the same task-specific splits (Fig.~\ref{fig:yield_main}b). On KetRed conversion RxnPFN reaches test $R^2=0.452$ (95\% interval $[0.329,0.571]$, MAE $8.1$ points), on par with Kumo on R-SPOC ($0.448$) and TabICLv2 ($0.440$) and significantly better than every tree ensemble (ExtraTrees $0.317$--$0.370$, random forest $0.360$--$0.366$, XGBoost $0.336$--$0.344$), DRFP + XGBoost ($0.325$) and the fine-tuned rxnfp-BERT ($0.274$, three-seed prediction average; per-seed means are in Supplementary Table~\ref{tab:yield_modern}), whose paired intervals all exclude zero; the three-seed prediction average of Chemprop reaches $0.343$. On AsymOx yield it reaches $0.651$ $[0.550,0.728]$ (MAE $10.3$ points), the highest of all models, ahead of random forest on R-SPOC ($0.639$), TabPFN ($0.632$) and ExtraTrees ($0.610$--$0.618$); the differences to the strongest baselines are within the bootstrap uncertainty, while those to DRFP + XGBoost ($0.545$), the three-seed prediction averages of Chemprop ($0.564$) and rxnfp-BERT ($0.569$), and TabICLv2 ($0.574$) are significant. The native quantile intervals of RxnPFN remain usable: at the nominal 90\% level they cover 93.5\% of KetRed and 90.6\% of AsymOx test reactions (Fig.~\ref{fig:yield_main}e). Interval width correlates with absolute error (Spearman $\rho=0.83$ and $0.49$, respectively).

Across records grouped by substrate, product, catalyst, ligand, solvent, additive, temperature and pressure, pooled within-group standard deviations are $14.1$ percentage points for KetRed conversion (290 groups) and $16.2$ for AsymOx yield (79 groups). Reaction time and catalyst loading can vary within these groups.

Within a campaign, yield behaves like selectivity. In the Heck collection, adding labels from a held-out paper lowers Kumo's yield MAE from $17.35$ at $K=0$ to $11.29$ at $K=4$, against $12.31$ for the mean of the added labels, beating that control in 9 of 12 papers (Fig.~\ref{fig:yield_main}f). In simulated campaigns, greedy selection by the frozen model reaches each paper's highest-yielding reaction after $10.3$ reactions on average, against $13.5$ for random selection, faster in 11 of 12 papers (mean difference $-3.1$ reactions, 95\% interval $[-5.4,-1.2]$; Wilcoxon $p=0.007$; Fig.~\ref{fig:yield_main}g).

\paragraph{Reaction-level cases span successes and failures.}

Figure~\ref{fig:cases} shows measured and predicted selectivity for individual reactions across seven reaction classes, selected by a fixed rule that spans each class's error distribution rather than its best cases. Most reactions fall within 10 ee points of the measurement (57--86\% per class). The largest errors cluster in tetrasubstituted alkene hydrogenation (26\% beyond 25 points), where AHO labels are signed and a wrong predicted direction produces a large ee error. Ketone reduction and the Heck adaptation setting have the fewest large errors. The corresponding yield and conversion cases are in Supplementary Fig.~\ref{fig:cases-more-y}.

%% file: tex/cases_more_figure.tex
\section{Further Held-Out Reaction Examples}
\label{app:cases-more}

\begin{figure*}[p]
\centering
\includegraphics[width=0.99\linewidth]{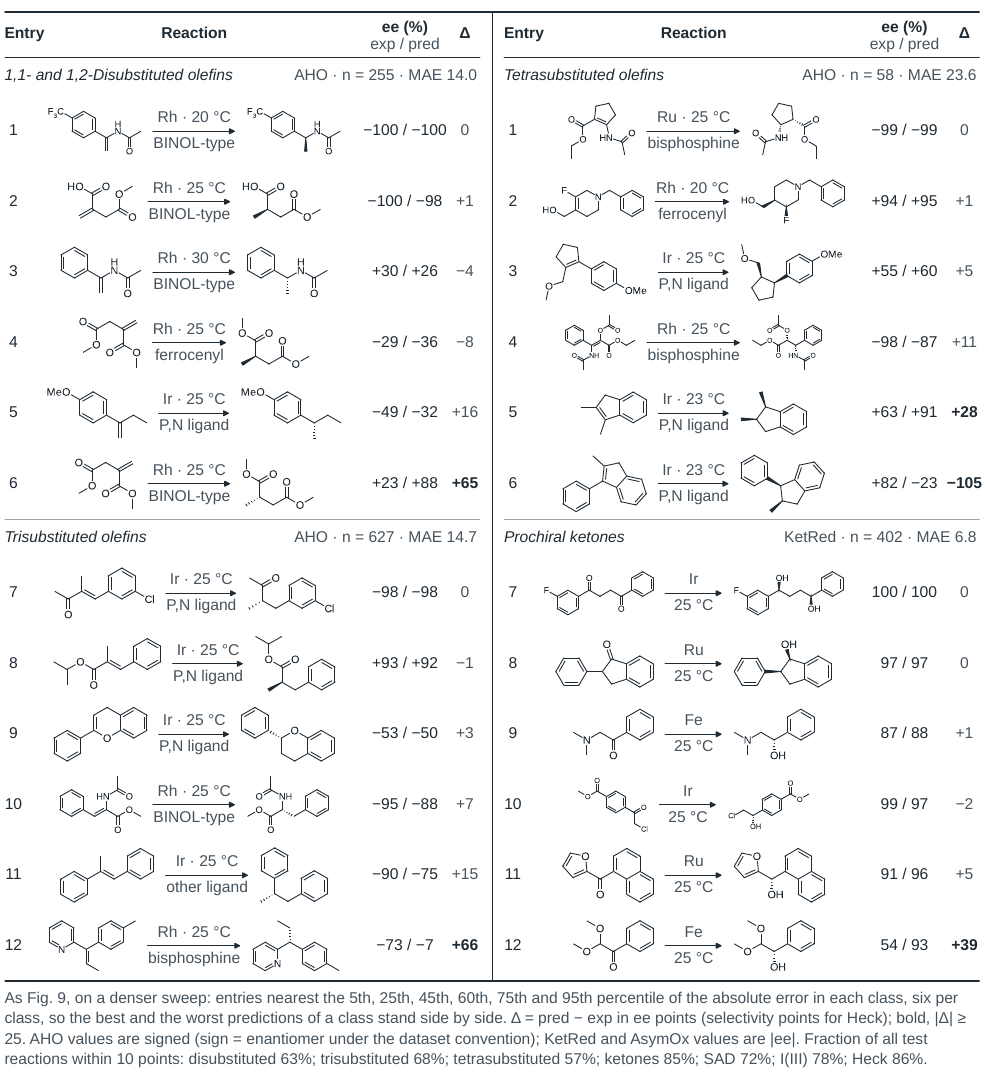}
\caption{\textbf{Further held-out reactions, olefin hydrogenation and ketone reduction.} Six reactions per class at the 5th, 25th, 45th, 60th, 75th and 95th percentile of the absolute error; conventions as in Fig.~\ref{fig:cases}. Each class header gives the dataset, the number of test reactions in the class and its mean absolute error. $\Delta$ = predicted $-$ measured in ee points; bold, $|\Delta|\ge25$. AHO values are signed, with the sign giving the enantiomer under the dataset convention; KetRed values are $|$ee$|$. The fractions of all test reactions of each class within 10 points are given in the footnote below the table.}
\label{fig:cases-more-a}
\end{figure*}

\begin{figure*}[p]
\centering
\includegraphics[width=0.99\linewidth]{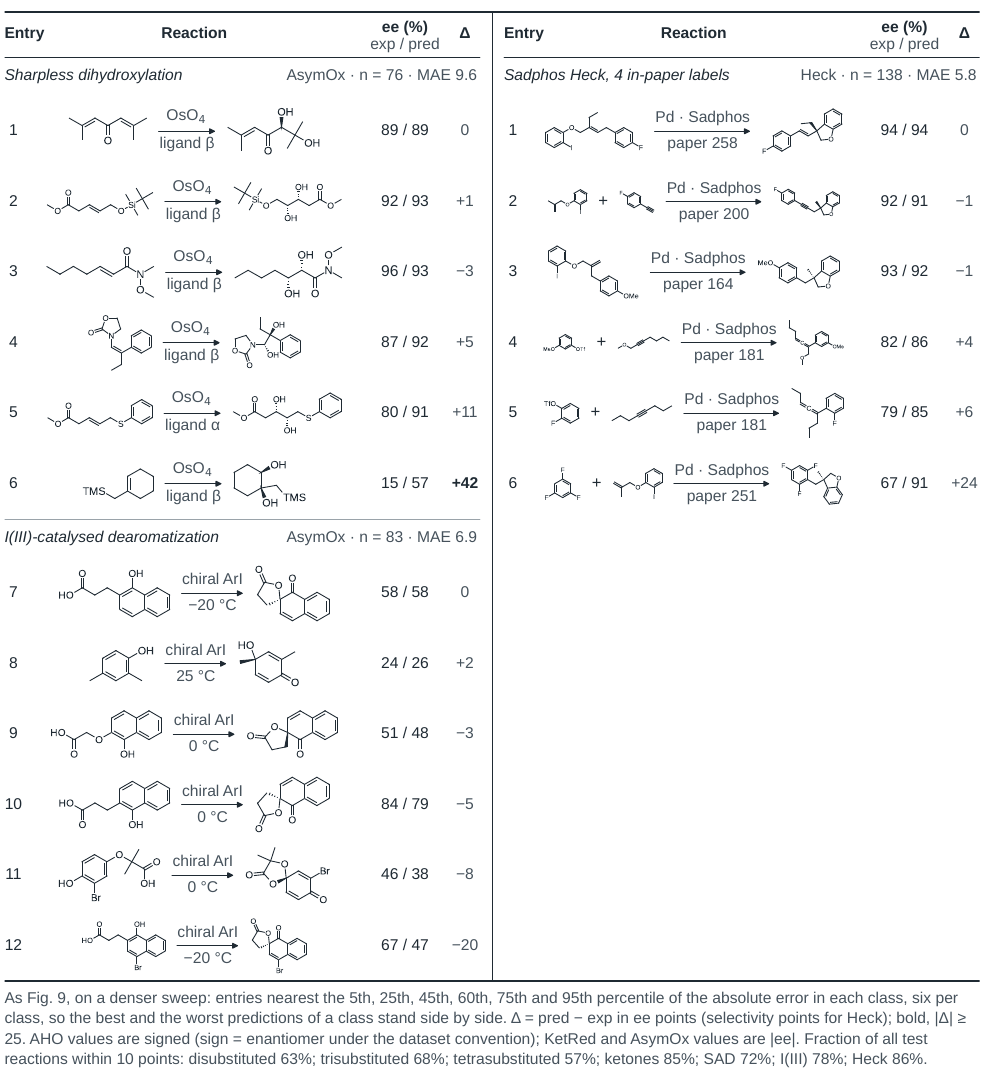}
\caption{\textbf{Further held-out reactions, oxidation and Heck selectivity.} As Supplementary Fig.~\ref{fig:cases-more-a}, for the two oxidation classes of AsymOx and for the Sadphos Heck queries with four labelled reactions from the held-out paper in the context. AsymOx values are $|\mathrm{ee}|$; Heck values are on the reported selectivity scale.}
\label{fig:cases-more-b}
\end{figure*}

\begin{figure*}[p]
\centering
\includegraphics[width=0.99\linewidth]{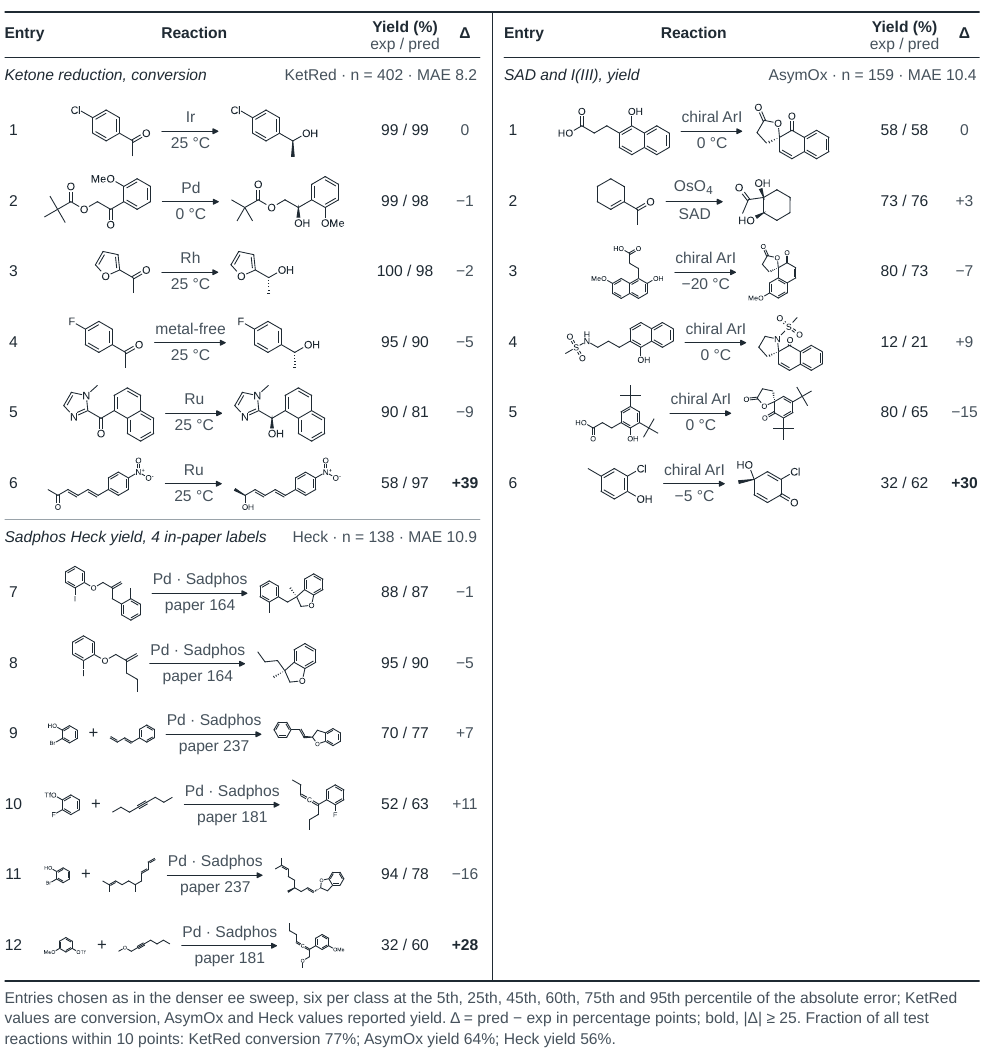}
\caption{\textbf{Further held-out reactions, conversion and yield.} As Supplementary Fig.~\ref{fig:cases-more-a}, six reactions per class for the KetRed conversions and the AsymOx and Sadphos Heck yields. KetRed values are conversion and the other two reported yield; the two endpoints are never pooled. $\Delta$ = predicted $-$ measured in percentage points; bold, $|\Delta|\ge25$.}
\label{fig:cases-more-y}
\end{figure*}

Figure~\ref{fig:cases} shows three reactions per class in the left column and four in the right, selected across the absolute-error distribution. The supplementary figures sample this distribution more densely, using six reactions per class at the 5th, 25th, 45th, 60th, 75th and 95th percentiles. They use the same held-out predictions and structure-size limit. Read down a class and the failure is usually a specific chemical difference---a substituent that no precedent covers, a substrate binding through an atom the model has not learned---rather than a uniform loss of accuracy. Supplementary Figs.~\ref{fig:cases-more-a} and \ref{fig:cases-more-b} give the selectivity classes and Supplementary Fig.~\ref{fig:cases-more-y} the conversion and yield classes.

%% file: tex/figs/figS_extrapolation.tex
\begin{figure*}[tbp]
\centering
\figinclude{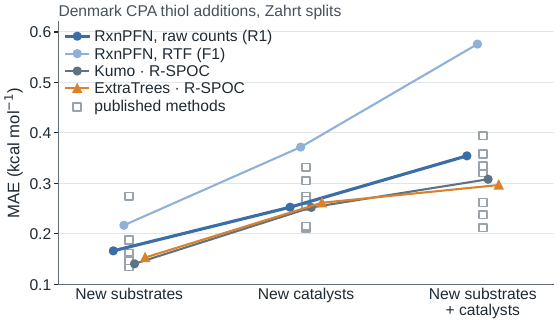}{figures/figS_extrapolation.png}
\caption{\textbf{Extrapolation stress test.} Denmark CPA-catalysed thiol additions \citep{Zahrt_2019} on the standard out-of-sample splits (384 context reactions; 216, 304 and 171 queries): MAE of RxnPFN on uncompressed role count fingerprints (R1) and on rank-controlled RTF projections (F1), of the frozen model and ExtraTrees on R-SPOC (mean of five seeds where applicable), and of published methods on the same splits (open squares; values as reported in the cited works). With only 24 context catalysts, SVD compression (F1) cannot represent unseen catalysts, whereas uncompressed role counts (R1) and R-SPOC match published methods on new substrates and new catalysts.}
\label{fig:extrapolation}
\end{figure*}

%% file: tex/uq_details.tex
A source-stratified group split reserves 1,458 AHO, 725 KetRed and 243 AsymOx reactions for calibration, leaving 6,064, 2,705 and 970 proper-fit reactions; substrate--product groups do not cross the two sets, and all 1,501 test reactions are kept. Projections and imputers are refitted on the 9,739 proper-fit reactions (474 non-constant features), and the same fitted predictor produces calibration and test quantiles. Corrections are fixed on calibration labels before test inference. For this pooled model, the 90\% pooled additive correction is $0.019$ kcal\,mol$^{-1}$ per side, and the source-wise corrections are $0.037$ (AHO), $0.010$ (KetRed) and $0$ (AsymOx, which needs no positive correction at any level); its native 90\% widths on the three sources are $2.307$, $1.608$ and $1.291$ kcal\,mol$^{-1}$. In the per-dataset setting each source uses only its own proper-fit reactions as context and its own calibration reactions; the 90\% CQR corrections of \rxndirect{} are then $0.034$, $0.012$ and $0$ kcal\,mol$^{-1}$ and the native widths $2.317$, $1.522$ and $1.211$ kcal\,mol$^{-1}$ for AHO, KetRed and AsymOx.

%% file: tex/substrate_coverage.tex
\begin{table}[htbp]
\centering
\caption{\textbf{Native 90\% prediction intervals by substrate similarity on the per-dataset test splits.} Each cell gives the number of measurements inside the interval divided by the number of test reactions, followed by the mean interval width in kcal\,mol$^{-1}$. The intervals and test groups are those used in Fig.~\ref{fig:insight}j.}
\label{tab:novelty-coverage}
\small
\begin{tabular}{lccc}
\toprule
Nearest substrate in context & AHO & KetRed & AsymOx\\
\midrule
Fingerprint match & 744/819,\ 1.58 & 205/223,\ 1.34 & 89/96,\ 0.79\\
Higher similarity & 57/61,\ 4.16 & 82/90,\ 1.42 & 30/33,\ 1.23\\
Lower similarity & 53/60,\ 4.78 & 85/89,\ 1.80 & 26/30,\ 1.80\\
\bottomrule
\end{tabular}
\par\medskip
\begin{minipage}{0.92\linewidth}\footnotesize
A match has nearest-context Morgan fingerprint Tanimoto similarity $\geq0.999$. The remaining reactions are split at the median similarity within their dataset (0.706, 0.704 and 0.667, respectively), with ties assigned to the higher-similarity group. These are subgroups of the original held-out reactions. AHO uses the sign-mixture interval, and KetRed and AsymOx use the native magnitude quantiles.
\end{minipage}
\end{table}

%% file: tex/recent_yield_baselines.tex
\begin{table}[htbp]\centering\footnotesize\setlength{\tabcolsep}{3pt}
\caption{\textbf{Deep baselines on yield and conversion.} Test $R^2$ (unitless) and MAE (percentage points). Trained models show mean $\pm$ s.d. across three seeds, with checkpoints selected on the fixed validation allocation. Frozen models and tree baselines have one fit. These means of metrics differ from the metrics of seed-averaged predictions in Fig.~\ref{fig:yield_main}.}
\label{tab:yield_modern}
\begin{tabular}{lrrrr}\toprule
 & \multicolumn{2}{c}{KetRed conversion ($n=402$)} & \multicolumn{2}{c}{AsymOx yield ($n=159$)}\\
\cmidrule(lr){2-3}\cmidrule(lr){4-5}
Model & $R^2$ & MAE & $R^2$ & MAE\\\midrule
RxnPFN (FDE, frozen) & 0.452 & 8.14 & 0.651 & 10.33\\
ExtraTrees, RTF & 0.370 & 8.83 & 0.610 & 10.75\\
DRFP + XGBoost & 0.325 & 10.24 & 0.545 & 11.46\\
Chemprop (CGR) & 0.318 $\pm$ 0.021 & 10.16 $\pm$ 0.21 & 0.540 $\pm$ 0.007 & 11.89 $\pm$ 0.25\\
rxnfp-BERT, fine-tuned & 0.196 $\pm$ 0.069 & 11.09 $\pm$ 1.28 & 0.528 $\pm$ 0.010 & 12.06 $\pm$ 0.18\\
ReactionT5v2-yield & 0.272 $\pm$ 0.016 & 10.35 $\pm$ 0.18 & 0.525 $\pm$ 0.004 & 12.30 $\pm$ 0.08\\
CheMeleon + Chemprop v2 & 0.197 $\pm$ 0.010 & 10.02 $\pm$ 0.21 & 0.602 $\pm$ 0.006 & 11.38 $\pm$ 0.09\\
ChemAHNet, retrained & 0.266 $\pm$ 0.055 & 10.61 $\pm$ 0.53 & 0.549 $\pm$ 0.017 & 11.71 $\pm$ 0.26\\
CARL & 0.178 $\pm$ 0.057 & 9.82 $\pm$ 0.63 & 0.574 $\pm$ 0.066 & 11.36 $\pm$ 0.81\\
RXNGraphormer & 0.274 $\pm$ 0.012 & 9.56 $\pm$ 0.31 & 0.064 $\pm$ 0.071 & 16.92 $\pm$ 0.34\\
\bottomrule\end{tabular}\end{table}

%% file: tex/figs/figS_smalldata_uq.tex
\begin{figure*}[tbp]
\centering
\figinclude{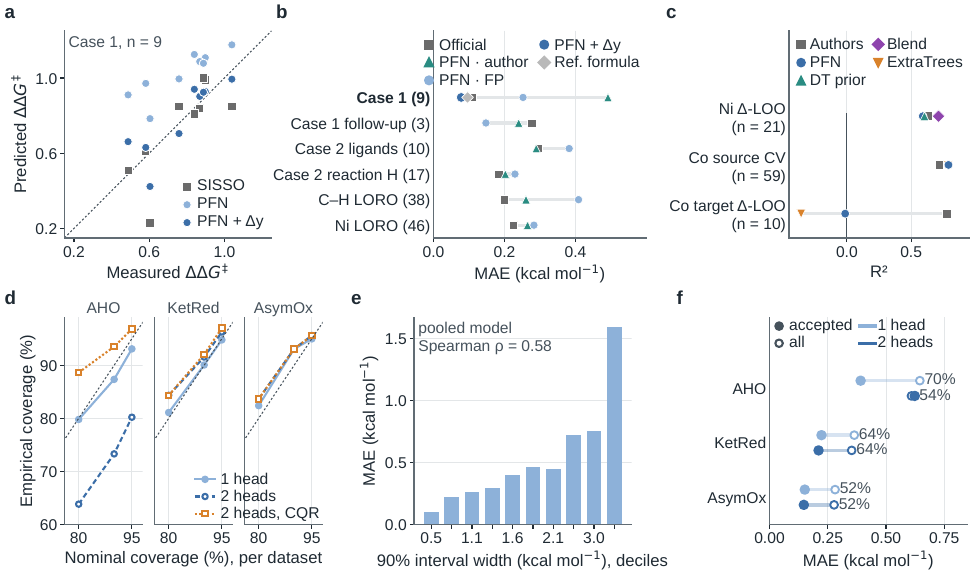}{figures/fig5_smalldata_uq.pdf}
\caption{\textbf{Sparse-data studies and predictive uncertainty.} \textbf{a}, Nature Case~1: predicted versus measured $\Delta\Delta G^\ddagger$ (kcal\,mol$^{-1}$) for the nine official out-of-sample reactions from the authors' SISSO model and the frozen model with the absolute target or the reference-relative response $\Delta y$. \textbf{b}, MAE on the official tasks of Gallarati et al.\ (number of test reactions in parentheses): official model (Case~1, SISSO; other tasks, the published MLR, reconstructed from the released sources for Ni LORO), the frozen model with author descriptors or fingerprints (FP), and for Case~1 the reference-relative variant and the fitting-free additive reference formula. \textbf{c}, $R^2$ in the Pd$\rightarrow$Ni and Co studies \citep{Xu2025NiTransfer,Zhang2023ChiralAcid}: the authors' delta models, ExtraTrees, the frozen model, a decision-tree-prior variant and the fixed 50/50 ensemble. \textbf{d}, Empirical versus nominal coverage per dataset, each dataset with only its own proper-fit reactions as context and its own calibration reactions: native intervals of the single-head model (RxnPFN w/o decomp.), native intervals of the two-head RxnPFN pipeline and the latter after CQR. On KetRed and AsymOx the two-head pipeline reduces to its magnitude head. \textbf{e}, MAE in deciles of the native 90\% interval width for the pooled single-head model (proper-fit context of 9,739 reactions, all 1,501 test reactions). \textbf{f}, Selective prediction per dataset: MAE of all test reactions (hollow) and of those whose native 90\% width is below the calibration median (filled), with the accepted fraction, for the single-head and two-head models.}
\label{fig:smalldata}
\end{figure*}

%% file: tex/transfer_controls.tex
\begin{table}[htbp]
\centering
\caption{\textbf{Does adapting to new measurements improve on a constant bias correction?} All entries are held-out MAEs. The offset is the mean observed minus historical-only prediction on the $K$ support reactions, added to every historical prediction for that test group.}
\label{tab:matched-transfer}
\small
\begin{tabular}{lrrr}
\toprule
& Sadphos Heck & \multicolumn{2}{c}{Co hydrogenation}\\
\cmidrule(lr){3-4}
Context or predictor & $K=4$ & $K=16$ & $K=64$\\
\midrule
Historical reactions only & 18.10 & 0.751 & 0.751\\
Mean of target support labels & 11.79 & 0.743 & 0.726\\
Historical prediction plus offset & 8.32 & 0.734 & 0.733\\
RxnPFN with historical and target reactions & 5.93 & 0.671 & 0.551\\
\bottomrule
\end{tabular}
\par\medskip
\begin{minipage}{0.92\linewidth}\footnotesize
Heck MAE is in source-reported selectivity units. The historical context comprises the other papers. All four predictors use the same queries and all predictors with target labels see the same four support reactions per draw. We average over the ten draws within each paper, then equally over twelve papers. The mean paired MAE reduction relative to the offset is 2.39 units (95\% interval 1.29--3.67, 10,000 resamples of papers). The offset leaves the historical ranking unchanged. The mean within-paper Spearman correlation is 0.334 for historical predictions and 0.408 for the updated context, averaged in the same order. Co MAE is in kcal\,mol$^{-1}$ for $|\Delta\Delta G^\ddagger|$. The historical context comprises 9,358 non-Co hydrogenations. All columns use the archived support and test memberships. Co results weight the five substrate-grouped folds and three support draws per fold equally. RxnPFN with only the Co support reactions has MAEs of 0.716 and 0.627 at 16 and 64 labels, respectively.
\end{minipage}
\end{table}

%% file: tex/newdata_si.tex
\begin{table}[htbp]\centering\footnotesize\setlength{\tabcolsep}{3.2pt}
\caption{\textbf{Co-AHO (532 reactions, 16 papers).} Test $R^2$ and MAE (kcal\,mol$^{-1}$) of $|\Delta\Delta G^\ddagger|$ and AUPRC at 80\% ee, pooled over the five substrate-grouped folds and over the 16 leave-one-publication-out folds. $\Delta R^2$: difference to RxnPFN with 95\% paired cluster-bootstrap interval (grouped folds). ``+ AHO'': the 9,358 non-Co AHO reactions added to the context. Deep models: mean over three seeds (grouped folds) or one seed (publication folds).}
\label{tab:newdata-co}
\begin{tabular}{lrrrrrr}
\toprule
 & \multicolumn{4}{c}{New substrates (5 folds)} & \multicolumn{2}{c}{New paper (16 folds)}\\
\cmidrule(lr){2-5}\cmidrule(lr){6-7}
Model & $R^2$ & MAE & AUPRC & $\Delta R^2$ vs RxnPFN & $R^2$ & AUPRC\\
\midrule
RxnPFN & 0.455 & 0.485 & 0.944 &  & $-$0.272 & 0.850\\
TabPFN 3.5 (FDE) & 0.428 & 0.491 & 0.938 & [$-$0.007, 0.065] & $-$0.140 & 0.846\\
Kumo (RTF) & 0.378 & 0.525 & 0.934 & [0.039, 0.114] & $-$0.200 & 0.845\\
ExtraTrees (RTF) & 0.328 & 0.532 & 0.914 & [0.066, 0.202] & $-$0.099 & 0.842\\
RandomForest (RTF) & 0.338 & 0.537 & 0.920 & [0.057, 0.173] & $-$0.105 & 0.855\\
XGBoost (RTF) & 0.316 & 0.544 & 0.920 & [0.067, 0.211] & $-$0.284 & 0.823\\
ExtraTrees (R-SPOC) & 0.330 & 0.510 & 0.916 & [0.039, 0.197] & $-$0.539 & 0.794\\
RandomForest (R-SPOC) & 0.371 & 0.515 & 0.928 & [0.017, 0.143] & $-$0.171 & 0.822\\
XGBoost (R-SPOC) & 0.297 & 0.548 & 0.919 & [0.080, 0.226] & $-$0.350 & 0.832\\
DRFP + XGBoost & 0.263 & 0.554 & 0.899 & [0.097, 0.280] & $-$0.019 & 0.837\\
Chemprop & 0.170 & 0.593 & 0.881 & [0.161, 0.315] & -- & --\\
rxnfp-BERT (fine-tuned) & 0.124 & 0.623 & 0.885 & [0.158, 0.355] & -- & --\\
CheMeleon-Chemprop & 0.216 & 0.597 & 0.890 & [0.104, 0.309] & -- & --\\
Context mean & $-$0.005 & 0.739 & 0.806 & [0.379, 0.559] & $-$0.033 & 0.795\\
RxnPFN + AHO & 0.481 & 0.470 & 0.957 &  & $-$0.244 & 0.880\\
ExtraTrees (RTF) + AHO & 0.384 & 0.534 & 0.935 & [0.037, 0.161] & 0.156 & 0.897\\
RxnPFN, AHO context only ($K=0$) & $-$0.139 & 0.769 & 0.891 & & & \\
\bottomrule
\end{tabular}
\end{table}

\begin{table}[htbp]\centering\footnotesize\setlength{\tabcolsep}{3.2pt}
\caption{\textbf{Out-of-sample Ir/Rh hydrogenations of Singh and Hern{\'a}ndez-Lobato (246 reactions, 5 papers).} AUPRC and AUROC at 80\% ee, MAE of $|\Delta\Delta G^\ddagger|$; mean $\pm$ s.e. over ten draws of 64 support and 128 query reactions. Published values (main text of ref.~\citealp{Singh_2025}) are for models meta-trained on 11,932 Ir, Rh and Co reactions and adapted on the support set; products of this set were generated (Methods).}
\label{tab:newdata-oos}
\resizebox{\linewidth}{!}{%
\begin{tabular}{lrrr}
\toprule
Model & AUPRC & AUROC & MAE\\
\midrule
RxnPFN, AHO context + 64 support & 0.972 $\pm$ 0.002 & 0.888 & 0.417\\
TabPFN 3.5 (FDE), AHO context + 64 support & 0.972 $\pm$ 0.003 & 0.890 & 0.429\\
ExtraTrees (RTF), AHO context + 64 support & 0.964 $\pm$ 0.003 & 0.871 & 0.476\\
Context mean, AHO context + 64 support & 0.916 $\pm$ 0.005 & 0.762 & 0.719\\
RxnPFN, 64 support only & 0.964 $\pm$ 0.005 & 0.862 & 0.462\\
TabPFN 3.5 (FDE), 64 support only & 0.956 $\pm$ 0.006 & 0.841 & 0.474\\
ExtraTrees (RTF), 64 support only & 0.934 $\pm$ 0.012 & 0.799 & 0.506\\
Context mean, 64 support only & 0.916 $\pm$ 0.005 & 0.762 & 0.692\\
RxnPFN, AHO context only & 0.947 $\pm$ 0.005 & 0.808 & 0.732\\
TabPFN 3.5 (FDE), AHO context only & 0.957 $\pm$ 0.003 & 0.837 & 0.665\\
ExtraTrees (RTF), AHO context only & 0.941 $\pm$ 0.005 & 0.803 & 0.727\\
Context mean, AHO context only & 0.916 $\pm$ 0.005 & 0.762 & 0.720\\
ProtoNet (published) & 0.9341 $\pm$ 0.0053 & -- & --\\
GNN (published) & 0.8764 $\pm$ 0.0090 & -- & --\\
Random forest (published) & 0.8631 $\pm$ 0.0061 & -- & --\\
\bottomrule
\end{tabular}}
\end{table}

\begin{table}[htbp]\centering\footnotesize\setlength{\tabcolsep}{3.2pt}
\caption{\textbf{Buchwald--Hartwig HTE (3,955 reactions).} Test $R^2$ of yield on the ten random 70/30 splits (mean $\pm$ s.d.) and on the additive out-of-sample Tests 1--4; RMSE in percentage points. Published values from Table~2 of ref.~\citealp{Probst_2022} on the same splits. Deep models: one seed on the random splits, mean of three seeds on Tests 1--4; BH-RGNN instead averages three seeds within each random split, with the displayed s.d. across ten split means. n.e., not evaluated.}
\label{tab:newdata-bh}
\resizebox{\linewidth}{!}{%
\begin{tabular}{lrrrrrrr}
\toprule
Model & Random $R^2$ & RMSE & Test 1 & Test 2 & Test 3 & Test 4 & Mean 1--4\\
\midrule
RxnPFN & 0.980 $\pm$ 0.002 & 3.8 & 0.729 & 0.820 & 0.718 & 0.547 & 0.704\\
TabPFN 3.5 (FDE) & 0.975 $\pm$ 0.003 & 4.3 & 0.776 & 0.765 & 0.642 & 0.530 & 0.678\\
Kumo (RTF) & 0.979 $\pm$ 0.002 & 3.9 & 0.652 & 0.759 & 0.671 & 0.471 & 0.638\\
ExtraTrees (RTF) & 0.935 $\pm$ 0.008 & 6.9 & 0.771 & 0.820 & 0.637 & 0.466 & 0.674\\
RandomForest (RTF) & 0.922 $\pm$ 0.009 & 7.6 & 0.764 & 0.780 & 0.588 & 0.483 & 0.654\\
XGBoost (RTF) & 0.940 $\pm$ 0.007 & 6.7 & 0.724 & 0.783 & 0.587 & 0.441 & 0.634\\
ExtraTrees (R-SPOC) & 0.930 $\pm$ 0.008 & 7.2 & 0.562 & 0.842 & 0.606 & 0.415 & 0.606\\
RandomForest (R-SPOC) & 0.932 $\pm$ 0.007 & 7.1 & 0.677 & 0.851 & 0.726 & 0.379 & 0.658\\
XGBoost (R-SPOC) & 0.945 $\pm$ 0.007 & 6.4 & 0.603 & 0.837 & 0.668 & 0.344 & 0.613\\
DRFP + XGBoost & 0.949 $\pm$ 0.003 & 6.1 & 0.816 & 0.828 & 0.703 & 0.496 & 0.711\\
Chemprop & 0.936 $\pm$ 0.004 & 6.9 & 0.751 & 0.745 & 0.647 & 0.451 & 0.649\\
rxnfp-BERT (fine-tuned) & 0.912 $\pm$ 0.010 & 8.1 & 0.808 & 0.818 & 0.709 & 0.449 & 0.696\\
CheMeleon-Chemprop & 0.945 $\pm$ 0.005 & 6.4 & 0.863 & 0.823 & 0.703 & 0.479 & 0.717\\
BH-RGNN, retrained & 0.957 $\pm$ 0.003 & 5.6 & n.e. & n.e. & n.e. & n.e. & n.e.\\
Context mean & $-$0.001 $\pm$ 0.001 & 27.3 & $-$0.020 & $-$0.003 & $-$0.000 & $-$0.013 & $-$0.009\\
\midrule
DFT RF (published) & 0.92 & -- & 0.80 & 0.77 & 0.64 & 0.54 & 0.69\\
Yield-BERT (published) & 0.95 $\pm$ 0.005 & -- & 0.84 & 0.84 & 0.75 & 0.49 & 0.73\\
Yield-BERT (aug.) (published) & 0.97 $\pm$ 0.003 & -- & 0.80 & 0.88 & 0.56 & 0.43 & 0.67\\
DRFP (xgboost) (published) & 0.95 $\pm$ 0.005 & -- & 0.81 & 0.83 & 0.71 & 0.49 & 0.71\\
\bottomrule
\end{tabular}}
\end{table}